# AI-assisted mitotic counting improves reproducibility and efficiency across multiple tumour types

Simon Graham[*a], Mostafa Jahanifar[a], Quoc Dang Vu[a], Vygantė Maskoliūnaitė[b], Donatas Petroška[b], Rūta Barbora Valkiūnienė[b], Ayat Gamal Lashen[c], Jen Hong Ong[d], Amede Ogechi Nnorom[d], Sinclair Couper[d], Natasha Kardasz[e], Reshma Agrawal[f], Brinder Singh Chohan[g], José Luis Solorzano Rendón[h], Shonali Natu[e], Arvydas Laurinavičius[b], Nasir Rajpoot[a], David Snead[a]

[a] Histofy Ltd, Coventry, United Kingdom
[b] Vilnius University Faculty of Medicine and National Center of Pathology, Affiliate of Vilnius University Hospital Santaros Klinikos, Vilnius, Lithuania
[c] Nottingham University Hospitals NHS Trust, Nottingham, United Kingdom
[d] The Newcastle Upon Tyne Hospitals NHS Foundation Trust, Newcastle, United Kingdom
[e] University Hospitals Tees, Stockton-on-Tees, United Kingdom
[f] University Hospitals Coventry and Warwickshire NHS Trust, Coventry, United Kingdom
[g] University Hospitals of Derby and Burton NHS Foundation Trust, Derby, United Kingdom
[h] Fundación MD Anderson Cancer Center España, Madrid, Spain

[*] Corresponding author: s.graham@histofy.ai

## Abstract

Mitotic counting is an important component of tumour grading, diagnosis and prognostic assessment across several tumour types, but manual assessment is time-consuming and subject to inter-pathologist variability. To help address these challenges, we developed MitPro, an artificial intelligence (AI) tool designed to improve consistency and efficiency by directing pathologists towards regions with the highest predicted mitotic activity and highlighting candidate mitotic figures for review, while retaining pathologist control over region selection and the final count. We evaluated its effect on the reproducibility and efficiency of mitotic counting in a retrospective, non-interventional, multi-centre paired reader study comprising 385 whole-slide images from three centres in three countries and seven tumour types using three different scanners. Thirteen pathologists participated, with each slide assessed independently by three pathologists without AI assistance and again with AI assistance after a minimum two-week washout period. Across all 385 slides, AI-assisted counting increased the intraclass correlation coefficient from 0.589 to 0.949, a difference of 0.360 with a 95% confidence interval of [0.260, 0.449]. Mean pathologist-level median assessment time decreased from 286.4 to 127.8 seconds, corresponding to an average saving of 151.8 seconds, or approximately 55%, per assessment. Improvements in agreement and efficiency were also observed in supporting analyses using HALO AP and Sectra image management systems and in two additional tumour types outside the main study population. AI-assisted assessment was associated with a limited shift towards higher mitotic counts and scores, consistent with identification of more active mitotic hotspots and fewer missed mitotic figures. The frequency of score change between unassisted and AI-assisted assessment was comparable with inter-pathologist variation during routine counting. These findings support the use of MitPro as an assistive tool for more consistent and efficient mitotic assessment in routine practice.

# Introduction

Mitotic figures are a morphological marker of cell division, and their quantification within a defined tumour area provides a practical measure of proliferative activity. Higher mitotic activity is associated with more aggressive tumour behaviour and poorer outcomes across a range of tumour types[1-11]. Mitotic counting is, therefore, an established component of several tumour grading systems, including those used for breast carcinoma[12], meningioma[13], sarcoma[14] and neuroendocrine tumours[15].

Given that malignant tumours are generally heterogeneous, mitotic activity can vary across different regions of the same tumour. Pathologists therefore identify areas with high mitotic activity, referred to as hotspots, and perform counts within these areas to capture the highest levels of mitotic rate. Because the area represented by a high-power field varies between microscopes, reporting mitotic activity as mitoses per mm² is recommended, particularly with the increasing digitalisation of routine slide assessment, to improve comparability between centres and reduce variation caused by differences in microscope field size[16].

Despite efforts to improve standardisation, mitotic counting remains subject to inter-pathologist variation[17-23]. Two likely sources of this variation are the selection of hotspots for assessment and the recognition of individual mitotic figures within them. Mitotic figures may be sparse and easily overlooked, while their identification can be subjective because malignant mitoses may be atypical and several cellular features can mimic mitotic figures, including apoptotic bodies, pyknotic nuclei and crushed or degenerate cells.

The adoption of digital pathology has streamlined the mitotic counting workflow[24-26] and laid the foundation for computer-assisted mitotic assessment. One advantage of digital pathology is that fixed-area regions of interest can be defined directly on the computer monitor, supporting more consistent counting over a known area in mm²[16].

Artificial Intelligence (AI) offers an opportunity to improve the consistency of mitotic counting between pathologists. Current approaches to automated mitotic figure recognition are largely based on deep learning, with recent advances driven by improvements in neural architectures and the availability of large annotated datasets for model development[27-35]. Detecting mitotic figures across a whole-slide image[36] facilitates objective identification of hotspot regions with the highest mitotic activity. This can help standardise hotspot selection by directing the pathologist towards regions containing the greatest number of detected mitotic figures. Within the selected counting regions, the detected mitotic figures can also be displayed to the pathologist, helping with their identification, review and verification during counting.

Several studies have demonstrated the potential clinical benefit of AI-assisted mitotic counting. Pantanowitz *et al.*[37] showed improved accuracy and reduced assessment time in invasive breast carcinoma, although the study used selected high-power fields rather than a whole-slide workflow. In a study of canine cutaneous mast cell tumours, Bertram *et al.*[38] found that AI assistance improved interobserver agreement by supporting both hotspot selection and recognition of individual mitotic figures. Balkenhol *et al.*[39] demonstrated improved agreement in mitotic counts in invasive breast carcinoma when pathologists assessed AI-selected hotspots, highlighting hotspot selection as an important source of variability. More recent studies by Tessier *et al.*[40] and Simmat *et al.*[41] have reported reduced assessment time and improved reproducibility with AI-assisted mitotic counting and scoring and have supported regulatory approval of commercial tools.

There remains limited clinical evidence for AI-assisted mitotic counting across the broader range of human tumour types in which mitotic assessment is routinely performed. Further validation is, therefore, needed to determine whether improvements in reproducibility and efficiency are maintained across different tumour types and clinical settings. In this study, we address this gap through a large multi-centre evaluation of AI-assisted mitotic counting across seven human tumour types.

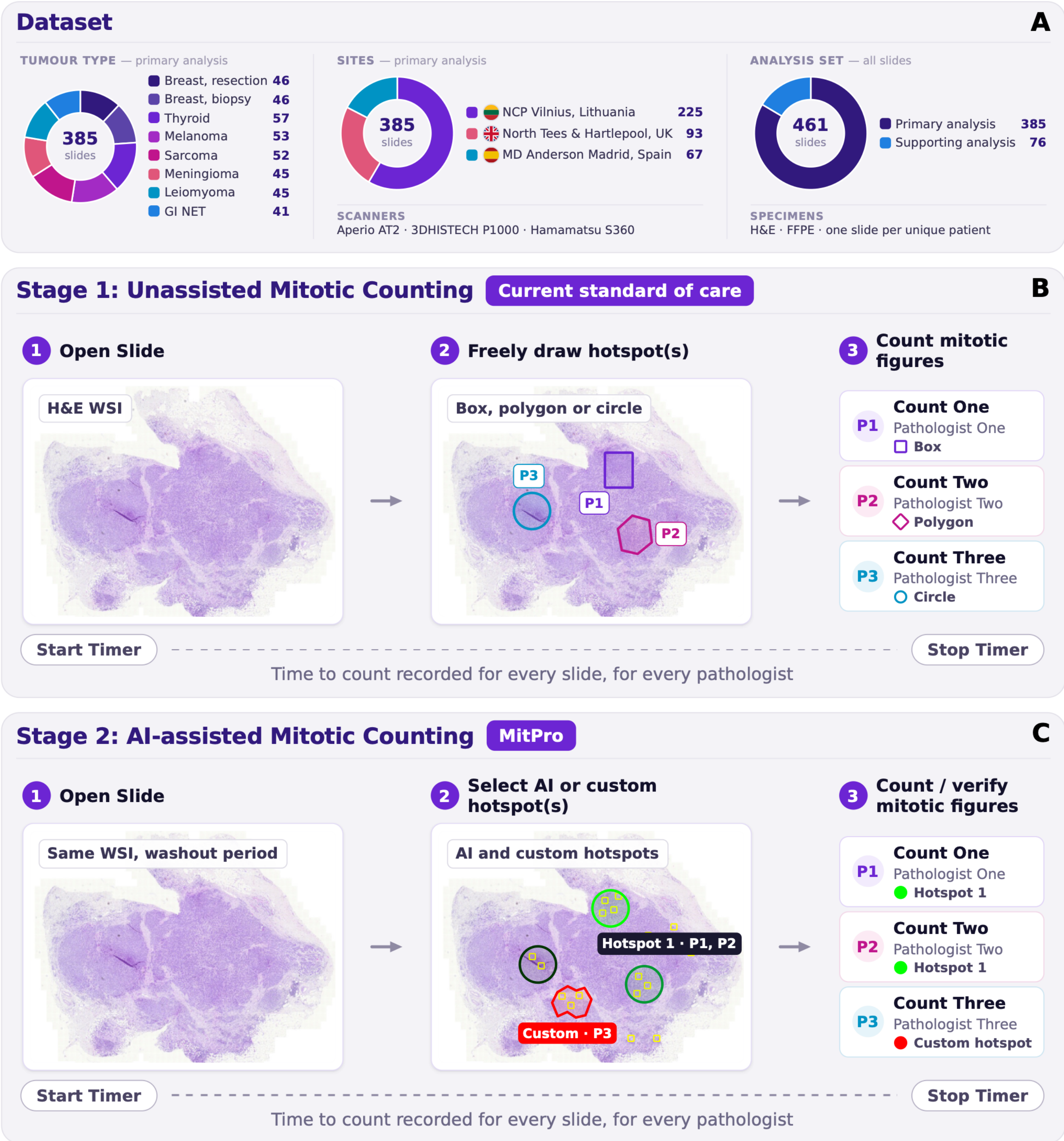

***Figure 1:** Overview of the clinical validation study. A) Datasets from three sites, 461 whole-slide images. B) Stage 1: unassisted mitotic counting, reflecting current standard of care. C) Stage 2: AI-assisted counting with MitPro.*

# Materials and methods

## Study design

A retrospective, non-interventional, multi-centre, multi-reader, two-stage paired design was implemented to evaluate the effect of AI assistance on the consistency and efficiency of mitotic counting under conditions representative of routine diagnostic workflows.

The study was conducted using the HALO AP (Indica Labs, 8700 Education Pl NW, Bldg. B, Albuquerque, NM 87114, United States) image management system (IMS) and comprised two stages. In Stage 1, pathologists performed mitotic counting according to standard clinical practice without AI assistance. In Stage 2, MitPro provided detected mitotic figures, proposed hotspot regions and associated mitotic counts. Each WSI was reviewed independently by three randomly selected pathologists, with the same three readers assessing that slide in both stages to preserve the paired design. A minimum washout period of two weeks separated the stages to minimise recall bias and support independent reassessment.

For each stage, interobserver agreement in mitotic counts and the time required to complete assessment were recorded. Timing began when the slide was opened and ended when the final mitotic count was confirmed, capturing slide review, region selection, assessment of mitotic figures and confirmation of the final count.

Supporting analyses (Supplementary Section S2) evaluated MitPro across two IMS environments and in two additional tumour types outside its intended use. A subset of slides was assessed in both HALO AP and Sectra (Sectra AB, Teknikringen 20, SE-583 30 Linköping, Sweden), with five pathologists reviewing each slide in each IMS. There was no reader crossover between systems, so no pathologist assessed the same slide in both environments.

The two study stages are summarised in Figure 1.

## Dataset

We collected retrospective WSIs from 385 unique patients treated at three independent sites. The sites and cases were distinct from those used for training and validation of the detection model. The participating centres were the National Centre of Pathology, Vilnius, Lithuania; North Tees and Hartlepool NHS Foundation Trust, United Kingdom; and MD Anderson Cancer Center Madrid, Spain. Slides came from three laboratories in three countries, each using its own staining protocol and scanner. WSIs were acquired using the Aperio AT2 (Leica Biosystems, Heidelberger Straße 17–19, 69226 Nussloch, Germany), Hamamatsu NanoZoomer S360 (Hamamatsu Photonics, 325-6, Sunayama-cho, Chuo-ku, Hamamatsu City, Shizuoka Pref., 430-8587, Japan), and 3DHISTECH P1000 (3DHISTECH, H-1141 Budapest, Öv u. 3., Hungary) scanners, with scan resolutions ranging from 0.23 to 0.50 µm/pixel.

All slides consisted of formalin-fixed, paraffin-embedded (FFPE) human tissue stained with Haematoxylin and Eosin (H&E) and prepared according to each site's standard histopathology protocols. Cases were selected to represent the seven tumour types for which mitotic counting forms part of routine clinical assessment. These comprised breast carcinoma, cutaneous melanoma, thyroid tumours, meningioma, gastrointestinal neuroendocrine tumours (GI NETs), sarcoma and leiomyoma. Both resection and biopsy specimens were included for breast carcinoma. For all other tumour types, only surgical resection specimens were included.

A minimum of 40 slides per tumour type was specified based on the power analysis detailed in the statistical analysis plan. Only tumour types within the scope of the primary analysis were considered for inclusion. In some tumour groups, additional samples were recruited to ensure adequate representation of histological subtypes. The distribution of the 385 samples included in the primary analysis by tumour type, subtype and clinical site is summarised in Table 1.

All images were fully de-identified before transfer to remove patient-identifiable information from the image and associated metadata. Before inclusion, all WSIs underwent quality review for staining, focus, and artefacts that could compromise interpretation. Further information on the study dataset is provided in Supplementary Section S1.1.

***Table 1:*** *Summary of tumour type distribution among the 385 slides included in the primary analysis of the clinical validation study. Benign tumours are indicated by an asterisk (*).*

| | Vilnius | North Tees | MDA Madrid | Total |
|---|---|---|---|---|
| **Breast Resection** | | | | **46** |
| Lobular carcinoma | 4 | 0 | 0 | 4 |
| Ductal carcinoma | 16 | 20 | 6 | 42 |
| **Breast Biopsy** | | | | **46** |
| Lobular carcinoma | 0 | 0 | 9 | 9 |
| Ductal carcinoma | 12 | 15 | 10 | 37 |
| **Meningioma** | | | | **45** |
| Meningothelial | 20 | 5 | 1 | 26 |
| Fibrous | 2 | 1 | 0 | 3 |
| Transitional | 5 | 3 | 0 | 8 |
| Atypical | 4 | 1 | 2 | 7 |
| Anaplastic | 0 | 0 | 1 | 1 |
| **GI NET** | 35 | 6 | 0 | **41** |
| **Thyroid** | | | | **57** |
| Medullary carcinoma | 7 | 1 | 0 | 8 |
| Oncocytic carcinoma | 0 | 3 | 0 | 3 |
| Papillary carcinoma | 22 | 4 | 6 | 32 |
| Follicular carcinoma | 1 | 5 | 1 | 7 |
| Follicular adenoma* | 0 | 5 | 2 | 7 |
| **Melanoma** | | | | **53** |
| Superficial Spreading melanoma | 11 | 3 | 6 | 20 |
| Nodular melanoma | 3 | 0 | 2 | 5 |
| Lentigo maligna melanoma | 5 | 0 | 0 | 5 |
| Acral lentiginous melanoma | 5 | 0 | 0 | 5 |
| Malignant, NOS | 13 | 1 | 0 | 14 |
| Dysplastic naevi* | 0 | 4 | 0 | 4 |
| **Leiomyoma** | 25 | 8 | 12 | **45** |
| **Sarcoma** | | | | **52** |
| Leiomyosarcoma | 1 | 2 | 3 | 6 |
| Liposarcoma | 3 | 0 | 3 | 6 |
| Fibrosarcoma & fibroblastic/myofibroblastic tumours | 6 | 2 | 0 | 8 |
| Rhabdomyosarcoma | 6 | 0 | 0 | 6 |
| Ewing sarcoma family tumours | 3 | 0 | 0 | 3 |
| Synovial sarcoma | 1 | 0 | 2 | 3 |
| Epithelioid sarcoma | 4 | 1 | 0 | 5 |
| Clear cell sarcoma | 3 | 1 | 0 | 4 |
| Malignant peripheral nerve sheath tumour | 3 | 0 | 1 | 4 |
| Other | 5 | 2 | 0 | 7 |

## Mitotic figure detection

Automated mitotic figure detection was performed using MitPro (Histofy Ltd., The Venture Centre, Sir Williams Lyons Road, Coventry, CV4 7EZ, United Kingdom) at a resolution of 0.25 µm per pixel, using a U-Net[42] based deep learning model inspired by the MDFS framework[31]. The model was trained on 28,222 annotated mitotic figures drawn from a diverse, multi-centre dataset spanning multiple cancer types, countries, institutions, and WSI scanners. The training set combined publicly available datasets[29, 43] with additional regions manually annotated from WSIs collected internationally, thereby increasing both the size and heterogeneity of the training data. We validated the performance of the model on a dataset independent of that used for training, consisting of 463 regions, each measuring 1 mm², taken from 215 WSIs across three centres. Each region was annotated by multiple pathologists, resulting in a dataset of 1,827 consensus mitotic figures. Across all tumour types and centres combined, the model achieved an F1 score of 0.803, with a precision of 0.728 and a recall of 0.894. Details of the analytical validation dataset and corresponding results are provided in Supplementary Section S3.

## Pathologist onboarding

A total of 13 pathologists participated in the primary analysis. Four additional pathologists were recruited for the supporting analysis, giving a total of 17 pathologists across the full study. Participants were recruited from multiple countries to provide representation across different diagnostic practices and training backgrounds. Before participation, all pathologists completed onboarding and training covering the study protocol, objectives, workflow and the relevant functionality of MitPro within the HALO AP and Sectra IMS platforms, including practising the workflow using a predefined set of training slides. This allowed them to become familiar with the tools, timing procedures and data recording methods before reporting study counts.

## Unassisted mitotic counting

When reporting the mitotic count according to standard of care, pathologists were free to follow their usual clinical practice. Counting was performed manually with no AI-generated information, overlays or hotspot suggestions available.

Mitotic counts were reported over a 2 mm² area for breast carcinoma, thyroid tumours, gastrointestinal neuroendocrine tumours and sarcoma, and over a 1 mm² area for melanoma, meningioma and leiomyoma, in line with the relevant tumour-specific guidelines[12-15, 44-46].

Pathologists also recorded the time in seconds required to complete each assessment by stopwatch, measured from opening the slide in the viewer to the final mitotic count decision. For each image, the mitotic count, time and assessment date were recorded.

## AI-assisted mitotic counting

In Stage 2, pathologists repeated the assessment with results provided by MitPro. When a slide was opened in the IMS viewer, MitPro displayed visual overlays showing detected mitotic figures and proposed hotspot regions corresponding to areas with the highest predicted mitotic activity. The hotspot area matched the unit area required for reporting for each tumour type, and each hotspot was accompanied by an AI-generated predicted mitotic count.

Pathologists could define a custom hotspot if they considered another region more suitable for assessment. For example, this could be done if the hotspot proposed by MitPro included white space or non-tumour tissue. MitPro then calculated the number of detected mitotic figures within the region selected by the pathologist.

Pathologists were able to correct any false positive or false negative detections made by MitPro and confirmed the final count based on their professional judgement. The proposed hotspots, therefore, supported region selection and counting but did not determine the final assessment.

Counts and assessment times were recorded using the same methodology as in Stage 1. In addition, pathologists recorded which AI-proposed hotspot was used or whether a custom region had been defined. For each assessment, the mitotic count, assessment time, date of assessment and hotspot selection method were documented.

# Statistical analysis plan

The primary statistical endpoint was agreement in mitotic counts, measured using the one-way random-effects intraclass correlation coefficient ICC(1,1)[47]. This statistic was selected because it estimates agreement between individual pathologists, with the pathologists treated as a random sample from a broader population of readers and allows different sets of pathologists to assess different images. ICC was calculated separately for Stage 1 and Stage 2 within each tumour group, and the effect of AI assistance was assessed using the difference between the two estimates. Confidence intervals for each ICC and their difference were estimated using a bias-corrected and accelerated paired cluster bootstrap with 20,000 samples. Slides were resampled with replacement using the same indices for both stages to preserve pairing.

Power calculations were performed using simulations based on the primary endpoint, the difference in ICC(1,1) between Stage 1 and Stage 2. Simulations assumed three pathologists per slide and incorporated the paired study design. A range of Stage 1 ICC values from 0.65 to 0.75 and Stage 2 ICC values from 0.80 to 0.95 was considered. Based on these simulations, a target of approximately 40 cases per tumour group was selected. This provided at least 80% power under scenarios in which the Stage 2 ICC reached 0.90 across the range of Stage 1 ICC values considered.

Pairwise Pearson correlation was used as a supporting measure of consistency between individual pathologists. Correlations were calculated separately for Stage 1 and Stage 2 for each reader pair using slides assessed by both readers. The difference between stages was used to describe the change associated with AI assistance. Confidence intervals for the correlations and their difference were estimated using the same paired bootstrap approach. Results are reported in Section S1.5 of the Supplementary Material.

Agreement in mitotic scores using in tumour grading was assessed using Fleiss' kappa, calculated separately for Stage 1 and Stage 2. Confidence intervals for each estimate and their difference were estimated using the same paired bootstrap approach. Mitotic scores were assigned using tumour-specific thresholds (Table S1), as used in clinical practice to support tumour grading and diagnosis[12-15, 44-46, 48, 49]. The mitotic score represents one component of the wider pathological assessment[12, 15, 50] and does not necessarily determine the final grade or diagnosis.

Assessment time was compared between Stage 1 and Stage 2 using paired observations from the same pathologist and slide. For each stage, the median time was calculated for each pathologist across the relevant slides, and these medians were then averaged across pathologists to obtain the stage-level summary. Time saved was calculated from the paired difference between Stage 1 and Stage 2 assessment times for each assessment, rather than from the difference between the stage-level summary values and was reported with a 95% confidence interval.

Beyond assessing whether AI assistance improved agreement between pathologists, we also examined whether it affected the final mitotic counts and derived scores by comparing matched Stage 1 and Stage 2 results for evidence of a systematic shift. Mean counts and standard deviations were calculated for each stage, and the change in mean count was considered relative to the Stage 1 standard deviation. Changes in tumour-specific mitotic scores were assessed using paired Stage 1 and Stage 2 scores and classified as unchanged, increased or decreased. To place these changes in the context of routine inter-pathologist variability, the proportion of assessments with unchanged scores between stages was compared with the proportion of pairwise Stage 1 assessments in which pathologists assigned the same score.

# Results

## Interobserver agreement in mitotic count

The ICCs for Stage 1 and Stage 2, together with the between-stage differences, are presented in Table 2 and Figure 2.

Across all 385 slides, the combined ICC increased from 0.589 in Stage 1, corresponding to moderate agreement (ICC 0.50-0.75), to 0.949 in Stage 2, corresponding to excellent agreement (ICC >0.90). The between-stage difference was 0.360, with a 95% confidence interval of 0.260 to 0.449. As the full confidence interval was above zero, this represented a significant improvement in inter-pathologist agreement during AI-assisted assessment. The increase from moderate to excellent agreement also indicates a clear reduction in interobserver variation in reported mitotic counts.

Across individual tumour types, Stage 1 ICC values ranged from 0.001 for thyroid tumours to 0.753 for breast biopsy specimens, whereas Stage 2 ICC values ranged from 0.405 for thyroid tumours to 0.973 for sarcoma. Agreement was higher with AI assistance for every tumour type, and the interval for the between-stage difference remained above zero in each case, indicating statistically significant improvements across all tumour types assessed.

***Table 2:*** *Inter-pathologist agreement (ICC) during Stage 1 conventional mitotic counting and Stage 2 AI-assisted mitotic counting, with the difference between stages (Δ ICC) and 95% confidence intervals, presented for the combined dataset and by tumour type, participating centre, sex, age group and scanner.*

| Group | Slides | ICC Stage 1 | ICC Stage 2 | Δ ICC |
|---|---|---|---|---|
| Combined | 385 | 0.589 [0.498, 0.684] | 0.949 [0.924, 0.970] | 0.360 [0.260, 0.449] |
| Vilnius | 225 | 0.533 [0.413, 0.636] | 0.958 [0.929, 0.981] | 0.426 [0.311, 0.534] |
| North Tees | 93 | 0.751 [0.609, 0.884] | 0.920 [0.814, 0.954] | 0.169 [0.041, 0.307] |
| MD Anderson Madrid | 67 | 0.542 [0.271, 0.754] | 0.944 [0.851, 0.981] | 0.402 [0.212, 0.595] |
| Breast resection | 46 | 0.548 [0.391, 0.671] | 0.933 [0.779, 0.987] | 0.384 [0.209, 0.523] |
| Breast biopsy | 46 | 0.753 [0.469, 0.919] | 0.912 [0.776, 0.958] | 0.159 [0.014, 0.337] |
| Melanoma | 53 | 0.444 [0.247, 0.601] | 0.864 [0.708, 0.926] | 0.420 [0.241, 0.618] |
| Thyroid | 57 | 0.001 [-0.130, 0.245] | 0.405 [0.172, 0.614] | 0.404 [0.178, 0.605] |
| GI NET | 41 | 0.503 [-0.099, 0.889] | 0.874 [0.458, 0.933] | 0.371 [0.043, 0.737] |
| Meningioma | 45 | 0.617 [0.159, 0.895] | 0.939 [0.746, 0.967] | 0.323 [0.075, 0.608] |
| Leiomyoma | 45 | 0.177 [0.038, 0.517] | 0.923 [0.587, 0.971] | 0.746 [0.401, 0.904] |
| Sarcoma | 52 | 0.547 [0.427, 0.667] | 0.973 [0.937, 0.992] | 0.426 [0.286, 0.545] |
| Female | 295 | 0.587 [0.456, 0.722] | 0.936 [0.893, 0.964] | 0.349 [0.213, 0.476] |
| Male | 90 | 0.582 [0.455, 0.722] | 0.961 [0.930, 0.987] | 0.380 [0.215, 0.505] |
| <40 | 52 | 0.715 [0.457, 0.928] | 0.927 [0.802, 0.965] | 0.211 [-0.001, 0.464] |
| 40-69 | 231 | 0.532 [0.401, 0.626] | 0.949 [0.905, 0.978] | 0.417 [0.325, 0.534] |
| 70+ | 102 | 0.643 [0.481, 0.822] | 0.958 [0.938, 0.985] | 0.316 [0.123, 0.494] |
| Aperio AT2 | 225 | 0.533 [0.411, 0.637] | 0.958 [0.930, 0.981] | 0.426 [0.311, 0.537] |
| Hamamatsu NanoZoomer S360 | 140 | 0.687 [0.545, 0.845] | 0.924 [0.856, 0.956] | 0.238 [0.086, 0.368] |
| 3DHistech P1000 | 20 | 0.723 [-0.110, 0.811] | 0.980 [0.838, 0.992] | 0.258 [0.176, 0.980] |

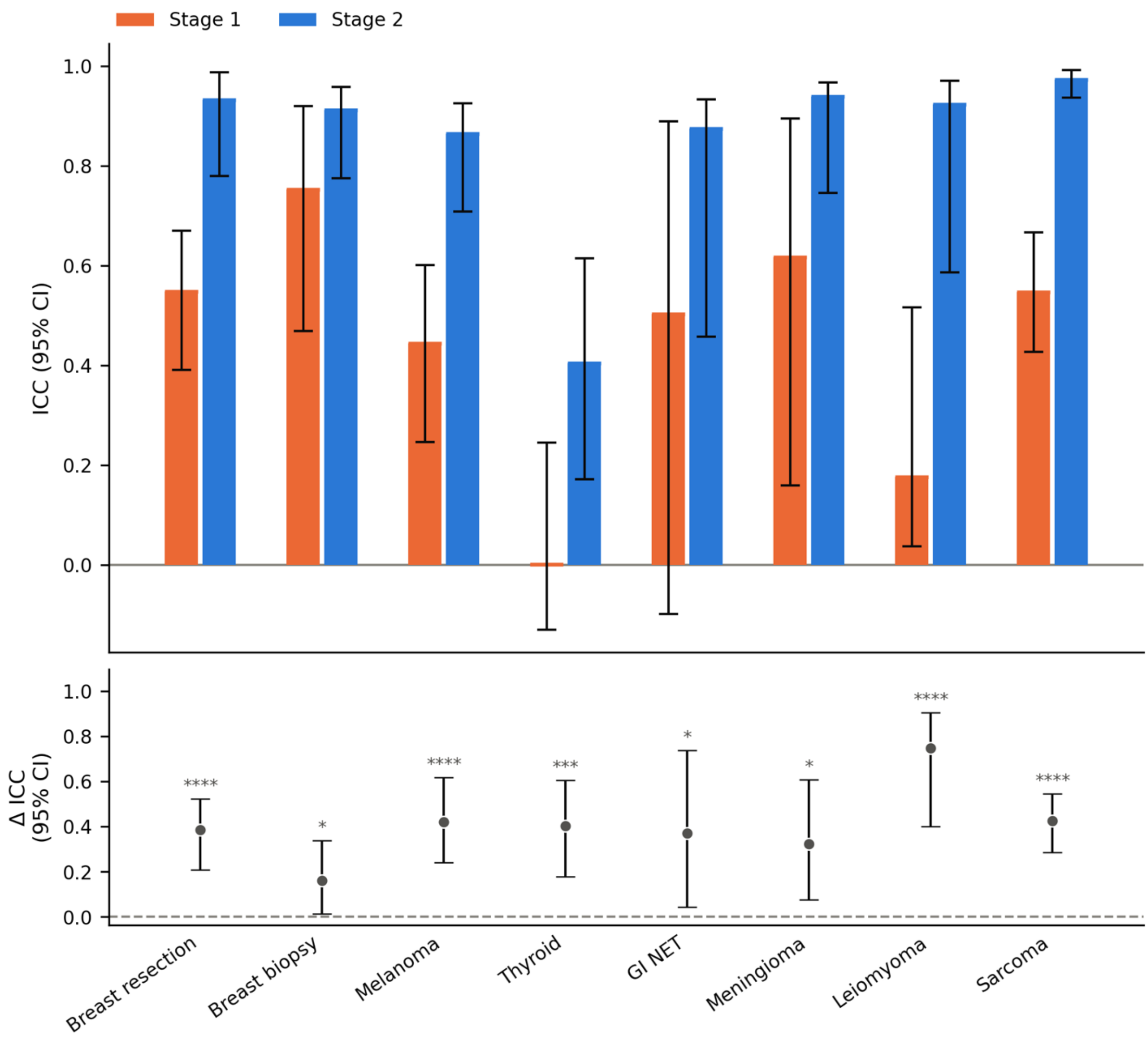

***Figure 2:*** *Inter-pathologist agreement (ICC) by tumour type during Stage 1 conventional mitotic counting and Stage 2 AI-assisted mitotic counting. The upper panel presents the Stage 1 and Stage 2 ICC estimates with 95% confidence intervals. The lower panel presents the difference between stages (ΔICC = Stage 2 - Stage 1) with 95% confidence intervals. Statistical significance of the difference between Stage 1 and Stage 2 is indicated by asterisks: $p < 0.05$ = *, $p < 0.01$ = **, $p < 0.001$ = ***, and $p < 0.0001$ = ****.*

Interobserver agreement improved across all three participating centres, as shown in Table 2. Despite slight differences in slide numbers, sample preparation, scanners and pathologists, Stage 2 ICC exceeded 0.910 at every centre. The 95% confidence interval for the difference in ICC between stages was also above zero for all centres. ICC increased from 0.533 to 0.958 at Vilnius, from 0.751 to 0.920 at North Tees, and from 0.542 to 0.944 at MD Anderson Madrid. The smaller increase at North Tees reflected the higher level of agreement already present in Stage 1. Visual results for each centre are presented in Figure S3.

Inter-pathologist agreement was similar for female and male patients in Stage 1, with ICC values of 0.587 and 0.582, respectively. Agreement increased in Stage 2 for both groups, reaching 0.936 for female patients and 0.961 for male patients.

When assessing results by age, slides from all tumour types were combined and grouped into three categories: under 40 years, 40-69 years, and 70 years or older. Inter-pathologist agreement increased with AI assistance in all age groups, with ICC rising from 0.715 to 0.927 in patients under 40 years, from 0.532 to 0.949 in those aged 40-69 years, and from 0.643 to 0.958 in those aged 70 years or older. The largest improvement was observed in the 40-69-year group. For patients under 40 years, the 95% confidence interval for the difference extended just below zero, likely reflecting the smaller sample size of 52 slides and the relatively high Stage 1 ICC.

Finally, all three scanners showed higher ICCs in Stage 2, with the 95% confidence intervals for the differences remaining above zero. However, the study was not designed to compare scanner performance directly, and differences should not be interpreted as showing better performance with one scanner than another. The scanner groups differed in sample size, participating centre, tumour distribution and reader composition, which may have influenced the observed differences between groups. For slides scanned using the Aperio AT2, ICC increased from 0.533 in Stage 1 to 0.958 in Stage 2, while increases from 0.687 to 0.924 and from 0.723 to 0.980 were observed for the Hamamatsu NanoZoomer S360 and 3DHistech P1000, respectively.

Detailed analyses by sex, age group and scanner are presented in Supplementary Section S1.2, with Bonferroni-adjusted results provided in Supplementary Section S1.3. Additional analyses by IMS and for two further tumour types are reported in Supplementary Sections S2.1.2 and S2.2.2, respectively.

## Interobserver agreement on mitotic scores

Fleiss' kappa increased from Stage 1 to Stage 2 across all tumour groups (see Figure 3), indicating that AI-assisted counting led to greater agreement among pathologists in assigning mitotic scores. These findings were consistent with the corresponding analysis of inter-pathologist agreement in mitotic counts. Statistically significant improvements in agreement were observed for breast resection, breast biopsy, thyroid, gastrointestinal neuroendocrine tumours and sarcoma. Improvements increased for meningioma and leiomyoma, although these did not meet significance at the 95% confidence interval level. Melanoma was excluded from this analysis, because a categorical mitotic score is no longer routinely assigned. Thyroid adenomas were also excluded, because mitotic scoring is applied to malignant thyroid tumours and is not applicable to benign adenomas.

The largest improvement was observed for thyroid tumours, where Fleiss' kappa increased from 0.084 during routine counting to 0.590 with AI assistance. Sarcoma also showed a marked increase, from 0.407 to 0.892, representing the highest Stage 2 estimate among the tumour groups assessed. For gastrointestinal neuroendocrine tumours, Fleiss' kappa increased from 0.088 to 0.496. Agreement also improved in both breast specimen groups, from 0.363 to 0.745 for breast resections and from 0.323 to 0.691 for breast biopsies. For meningioma and leiomyoma, Fleiss' kappa increased from 0.282 to 0.663 and from 0.300 to 0.423, respectively, although the confidence intervals for the differences included zero.

The results are summarised in Figure 3. The upper panel shows Fleiss' kappa for Stage 1 and Stage 2 for each tumour group, while the lower panel shows the corresponding difference in Fleiss' kappa between stages. The numerical results, including confidence intervals, are provided in Supplementary Section S1.4.

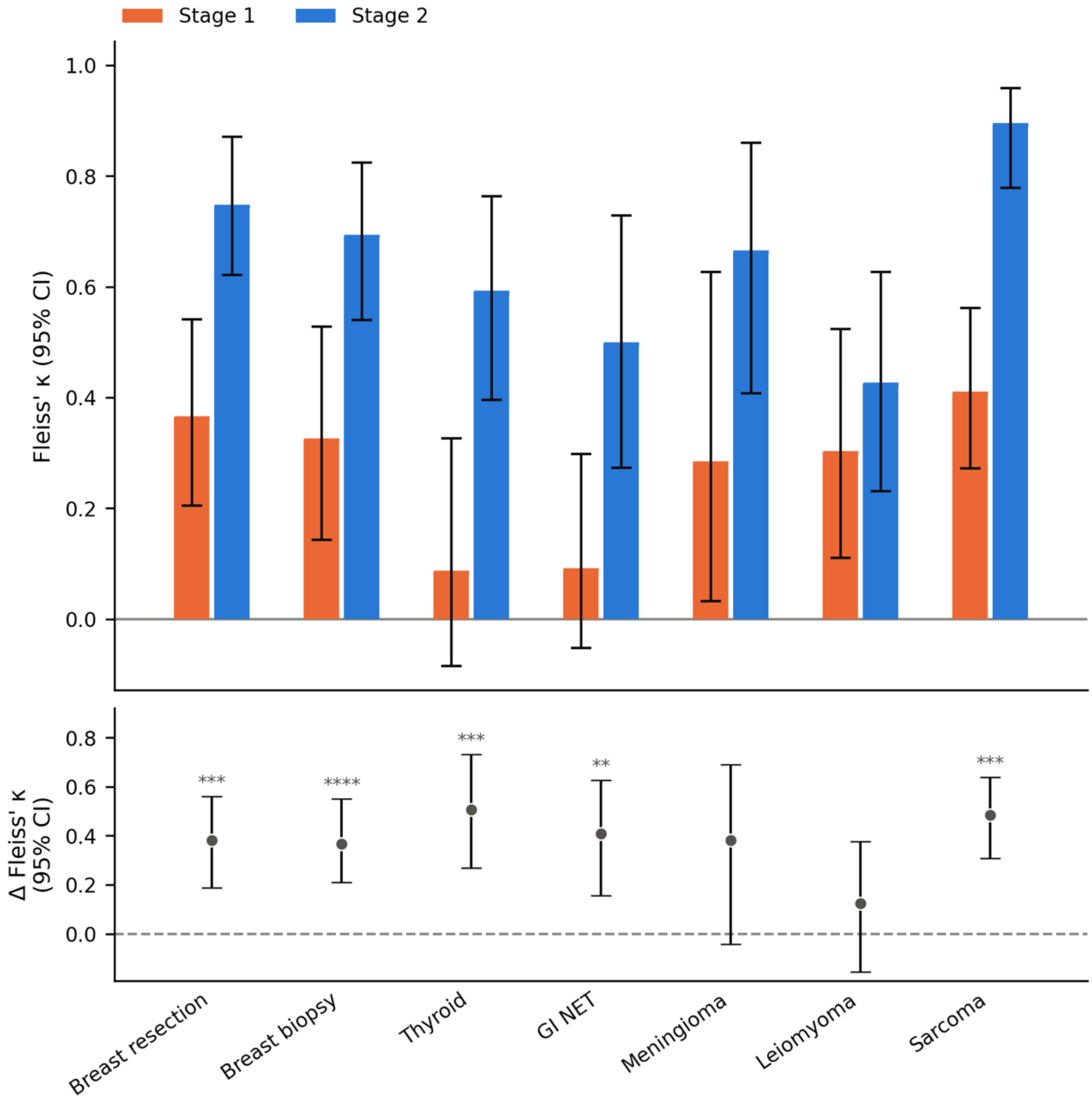

***Figure 3:*** *Fleiss' kappa by tumour type. The upper panel shows Fleiss' kappa estimates for Stage 1 and Stage 2 for each tumour type, with whiskers representing 95% confidence intervals. The lower panel shows the corresponding between-stage difference (Δκ = Stage 2 - Stage 1) with 95% confidence intervals. Statistical significance of the difference between Stage 1 and Stage 2 is indicated by asterisks: p < 0.05 = *, p < 0.01 = **, p < 0.001 = ***, and p < 0.0001 = ****.*

## Mitotic count assessment time

Across all slides, the average of the pathologist-level median assessment times decreased from 286.4 seconds in Stage 1 to 127.8 seconds in Stage 2. This corresponded to an estimated saving of 151.8 seconds (2.5 minutes) per assessment, representing an approximate 55% reduction in assessment time with AI-assisted counting. Across individual pathologists, the average time saved ranged from 30 to 380 seconds per assessment.

Figure 4 shows the Stage 1 and Stage 2 median assessment times for each pathologist. The combined value represents the median across all assessments from all pathologists, while the mean represents the average of the individual pathologist-level median assessment times. We also show the combined results, which groups all results together rather than taking the average between the pathologists.

Reductions were observed across every tumour group. Estimated time savings ranged from 137.4 seconds for melanoma, to 214.6 seconds for sarcoma.

Assessment time was lower in Stage 2 across all three participating centres. At Vilnius, the average of the pathologist-level median times decreased from 315.9 to 136.3 seconds, corresponding to an estimated time saving of 169.5 seconds, or a 57% reduction. At North Tees, it decreased from 289.8 to 146.6 seconds, with 143.6 seconds saved, representing a 49% reduction. At MD Anderson Madrid, the corresponding decrease was from 193.3 to 128.0 seconds, with an estimated 67.7 seconds saved and a 34% reduction. The amount of time saved varied between centres, which may reflect differences in reader composition, tumour distribution and baseline workflow characteristics, and should not be interpreted as a direct comparison of centre performance.

For female patients, assessment time decreased from 227.0 to 98.0 seconds, giving an estimated saving of 111.0 seconds and a 57% reduction. For male patients, it decreased from 238.0 to 105.0 seconds, with 117.0 seconds saved, corresponding to a 56% reduction.

Across age groups, Stage 2 times were consistently lower. For patients under 40 years, assessment time decreased from 240.0 to 83.5 seconds, giving an estimated saving of 126.0 seconds and reducing assessment time by approximately 65%. For patients aged 40 to 69 years, it decreased from 225.0 to 103.0 seconds, with 111.0 seconds saved, equivalent to a 54% reduction. For patients aged 70 years or older, the decrease from 227.0 to 100.0 seconds corresponded to an estimated saving of 113.0 seconds and a 56% reduction.

A similar pattern was observed across scanners. For Aperio AT2, assessment time decreased from 230.0 to 78.0 seconds, giving an estimated time saving of 127.5 seconds and a reduction of approximately 66%. For Hamamatsu NanoZoomer S360, it decreased from 215.5 to 142.0 seconds, with 91.5 seconds saved and assessment time reduced by approximately 34%. For 3DHistech P1000, the decrease from 300.0 to 150.5 seconds corresponded to an estimated time saving of 137.0 seconds, equivalent to a 50% reduction.

The amount of time saved varied between readers and tumour groups, which may reflect differences in baseline assessment speed and case complexity. The largest individual reductions exceeded six minutes per assessment, suggesting that MitPro may provide particularly large efficiency gains for more time-consuming assessments. From a workflow perspective, an average saving of approximately 2.5 minutes per assessment could translate into a meaningful reduction in cumulative reporting time when many slides are assessed.

Detailed analyses by individual pathologist, tumour type, age group, sex and scanner are presented in Supplementary Section S1.6. Additional analyses by IMS and for two further tumour types are reported in Supplementary Sections S2.1.3 and S2.2.3, respectively.

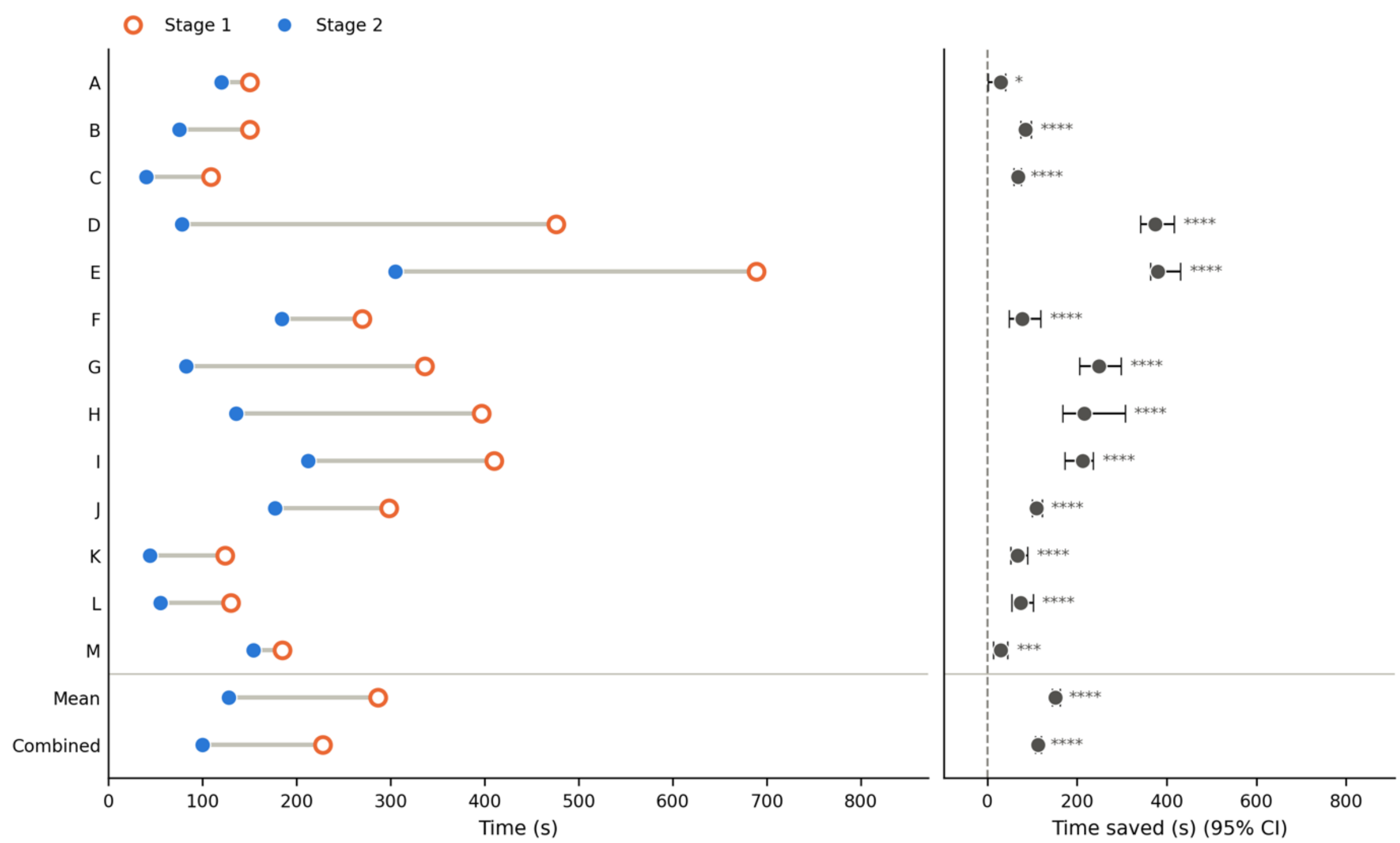

***Figure 4:*** *Median assessment time in Stage 1 and Stage 2 by pathologist. "Combined" represents the median assessment time across all assessments, while "Mean" represents the mean of the pathologist-level median assessment times. Time saved is calculated as Stage 1 minus Stage 2. Statistical significance of the between-stage difference is indicated by asterisks: p < 0.05 = *, p < 0.01 = **, p < 0.001 = ***, and p < 0.0001 = ****.*

## AI-assisted hotspots

Pathologist used an AI-assisted hotspot in 87.2% of assessments, while a manual region was used in 12.8%. Use of proposed hotspots remained equally high across tumour groups, sex, age and scanner. All pathologists used a custom hotspot in at least one case, and custom hotspots were used by at least some pathologists across all tumour types. Hotspot 1 was selected in 68.3% of assessments, Hotspot 2 in 13.2%, and Hotspot 3 in 5.7%, while a custom hotspot was selected in 12.8% of assessments. A visual summary by individual pathologist and tumour type is provided in Figure 5.

The use of manual regions in a minority of assessments highlights the importance of retaining pathologist control over hotspot selection when the AI-proposed regions are considered less appropriate for mitotic counting.

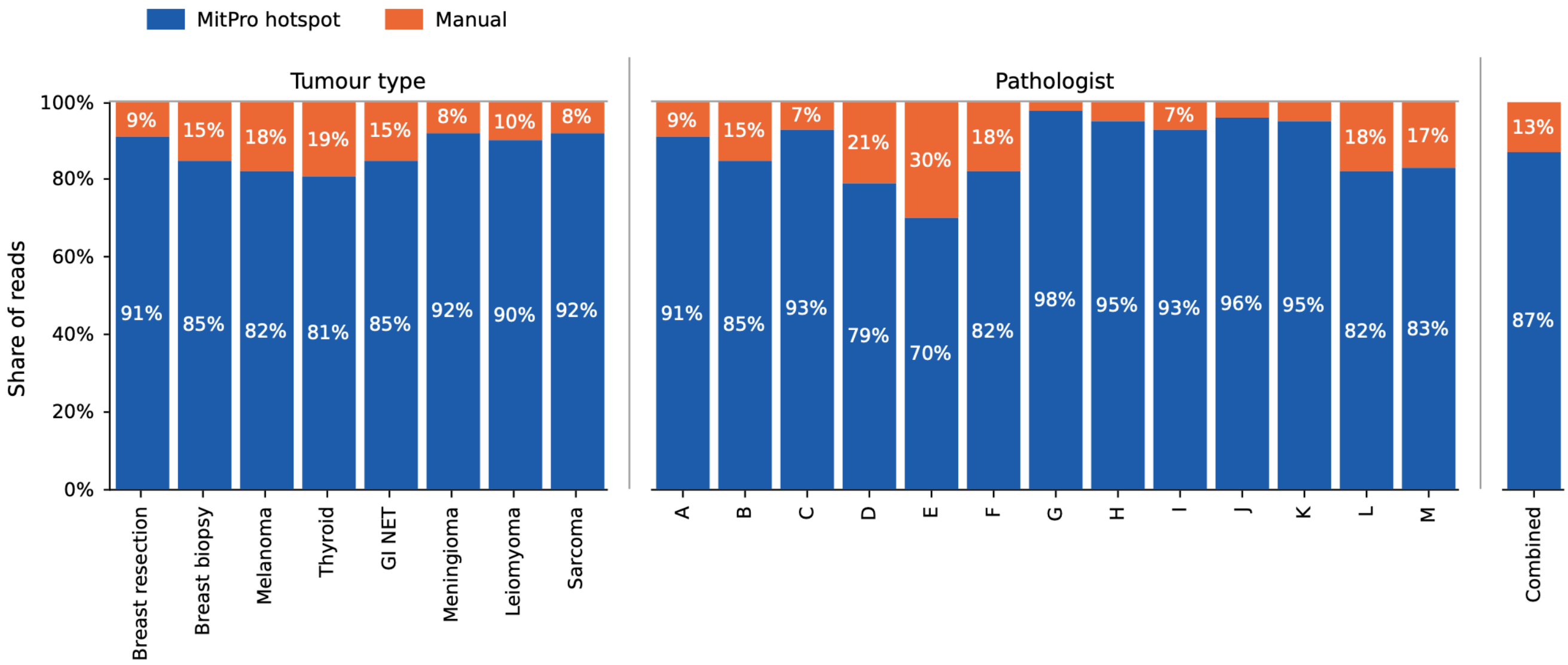

***Figure 5:*** *Hotspot selection by pathologist and tumour type. Bars show the proportion of assessments in which the pathologist selected a hotspot proposed by MitPro or manually defined an alternative hotspot.*

## Impact on mitotic count and score

Across all tumour groups, Stage 2 mitotic counts showed a slight shift towards higher values. For all eight tumour groups, the Stage 2 mean remained within one standard deviation of the corresponding Stage 1 mean. The largest shifts were generally seen in tumour groups with higher mitotic activity. A visual summary of the Stage 1 and Stage 2 count distributions is provided in Figure 6A.

We also compared the overall distribution of mitotic scores between Stage 1 and Stage 2. As shown in Figure 6B, the Stage 2 distribution showed a slight shift towards higher scores, consistent with the corresponding shift towards higher mitotic counts observed during AI-assisted assessment.

Across the combined dataset, scores remained unchanged in 64.9% of assessments, were higher in 29.1% and lower in 5.9%. The same general pattern was observed across tumour groups, sex, age and scanner, with scores often remaining unchanged and increases occurring more frequently than decreases. The corresponding distributions are shown in Figure S9 and Figure S10.

We also compared scores from both stages with the Stage 1 consensus mitotic score for each case, defined as the majority score assigned by the three pathologists during Stage 1. This provided a reference point for the scores assigned during routine assessment. The proportion of individual scores matching the Stage 1 consensus decreased from 82% in Stage 1 to 65% in Stage 2. This was largely due to an increase in scores above the Stage 1 consensus, from 12% to 33%, together with a reduction in scores below the consensus, from 6% to 2%. This pattern is shown in Figure 6C. Despite this shift, agreement between pathologists increased in Stage 2. Across all tumour types, the proportion of individual scores matching the consensus for the respective stage increased from 82% in Stage 1 to 92% in Stage 2. The percentage of slides for which all three pathologists assigned the same score also increased in Stage 2, from 50% to 75%. These findings indicate that the shift towards higher mitotic scores from Stage 1 to Stage 2 was accompanied by greater consistency between pathologists, rather than increased variation in scoring.

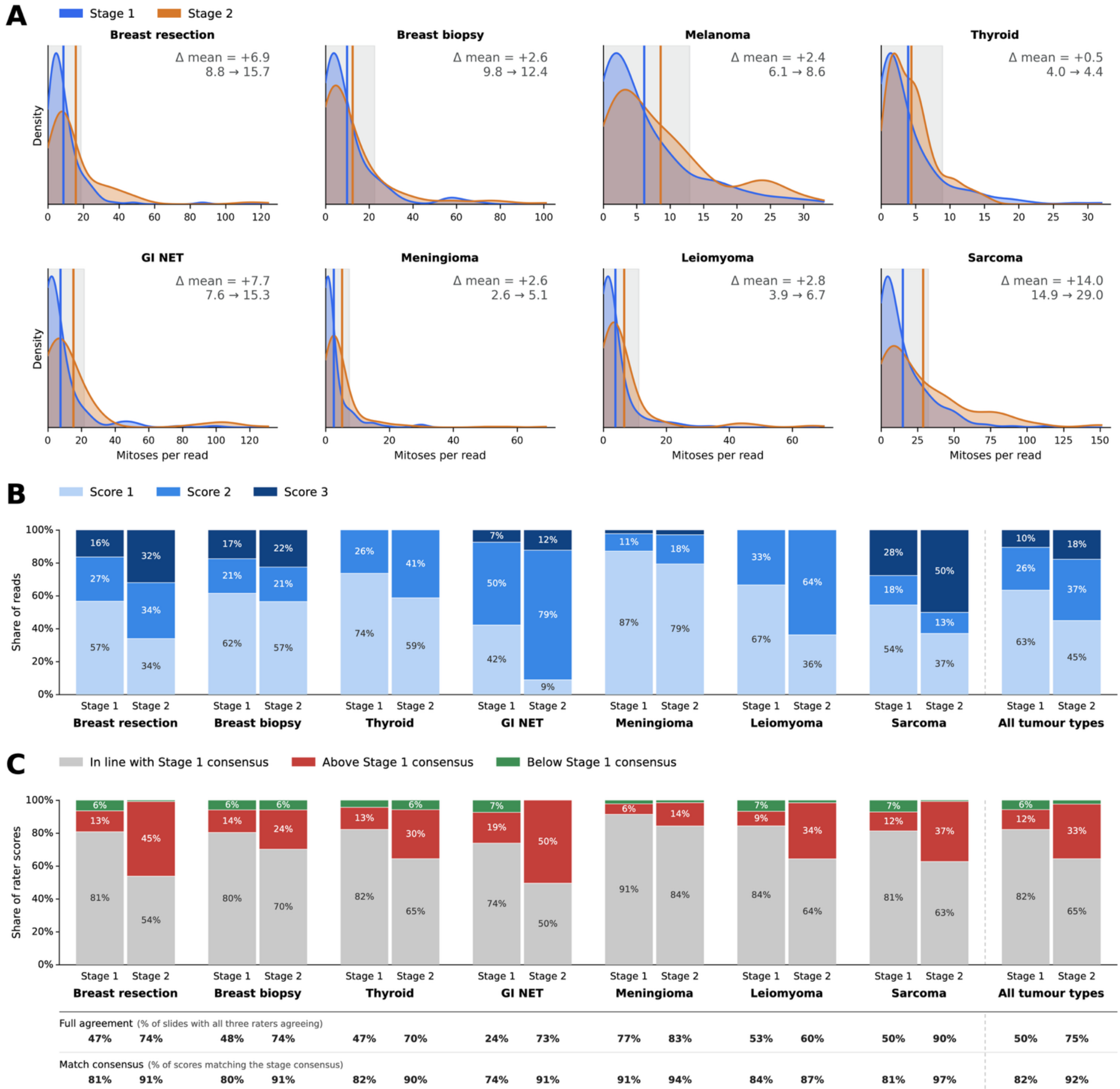

***Figure 6:*** *Distribution of mitotic counts and scores in Stage 1 and Stage 2 by tumour type. A) Distribution of mitotic counts per case (x-axis) and density (y-axis). The distributions show a modest shift between stages, with the magnitude varying by tumour type. Across all eight tumour groups, the Stage 2 mean remained within one standard deviation of the Stage 1 mean. B) Distribution of mitotic scores in Stage 1 and Stage 2 by tumour type. C) Proportion of mitotic scores in Stage 1 and Stage 2 that matched, exceeded, or fell below the Stage 1 consensus for each slide. The Stage 1 consensus was defined as the majority score assigned by the three Stage 1 pathologists for that slide. Below the bar plot, the percentage of slides with complete agreement among all three pathologists and the percentage of individual scores matching the within-stage consensus are shown for each stage.*

To place these changes in context, we also compared mitotic scores with score variation between pathologists during Stage 1. Across all tumour groups, 65.7% of Stage 1 pairwise comparisons produced the same score, compared with 64.9% of assessments remaining unchanged between Stage 1 and Stage 2. This indicates that the frequency of score change between stages was comparable with the variation already present between pathologists during routine assessment. For most tumour groups, the proportion of unchanged scores from Stage 1 to Stage 2 was at least as high as the agreement between pathologists during Stage 1. Breast resection and leiomyoma showed a greater proportion of score changes between stages, and these changes were predominantly towards higher Stage 2 scores. This was consistent with the broader shift towards higher mitotic counts observed during AI-assisted assessment, rather than reflecting greater inconsistency in score assignment. A visual comparison of Stage 1 inter-pathologist score agreement and changes in scores between Stage 1 and Stage 2 is provided in Figure S11.

# Discussion

Our study shows that AI assistance, using MitPro, delivers improved agreement between pathologists and reduction in the time taken to perform mitotic counts across all eight tumour groups. Agreement also improved for the categorical mitotic scores used in grading. Reductions in assessment time were seen across all pathologists as well as all tumour types. These benefits were observed consistently across centres, demographic groups and scanners. Pairwise correlations between pathologists showed the same pattern and are reported in the Supplementary Section S1.5.

AI assistance with MitPro addresses both main potential sources of variability in mitotic counting by supporting hotspot selection and recognition of individual mitotic figures within the selected region. By supporting both steps, MitPro helps pathologists reach more consistent assessments while retaining pathologist control over the final count.

The hotspot results also support this interpretation. Pathologists selected one of the proposed hotspots in most AI-assisted assessments, indicating that the suggested regions were usually considered suitable. However, manual regions were still used in a minority of cases, underlining the importance of allowing the pathologist to override the AI when a proposed hotspot was not appropriate.

AI assistance was associated with a shift towards higher mitotic counts in some tumour groups, which is consistent with improved identification of more mitotically active regions and candidate mitotic figures. A similar pattern was observed by Simmat *et al.*[41], where AI assistance was associated with a shift towards higher mitotic scores, alongside increased use of AI-proposed hotspots and greater agreement in counting regions. Such a change is to be anticipated if interobserver variation is due to selection of regions that do not capture mitotic hotspots or failure to record all mitoses within a selected region or indeed a combination of both errors. In this study most mitotic scores remained unchanged between Stage 1 and Stage 2, and when scores changed, they were more often higher than lower. When compared with the Stage 1 consensus, scores were higher in Stage 2, however this upward shift was accompanied by greater agreement between pathologists both in the number of cases with complete agreement in mitotic scoring and in the number of cases where scores matched the consensus. The changes in mitotic score were not significant, but the findings suggest that the upward shift in scores is due to improved detection of mitotic figures and reduced variation in scoring. Score changes between stages occurred at a frequency comparable to the differences already observed between pathologists during routine assessment, indicating that AI assistance did not introduce additional variability beyond that already present in mitotic counting.

The study was designed to assess reproducibility and efficiency rather than accuracy, and no independent ground truth was established for the clinical study slides. The observed shift towards higher counts should therefore not be interpreted as evidence that AI-assisted counts were more accurate. Previous studies that included an independent reference standard provide complementary evidence on accuracy. Tessier *et al.*[40] found that mitotic score accuracy was not reduced with AI-assisted counting, while Simmat *et al.*[41] reported improved accuracy against a consensus reference established by expert pathologists.

A key feature of the proposed workflow is that it provides decision support rather than an autonomous result. Pathologists remain responsible for selecting the counting region, reviewing detections, correcting errors and confirming the final mitotic count. This is particularly important when interpreting changes in count or score, as the final assessment continues to depend on pathologist review and judgement. Automation bias has been reported in AI-assisted mammography and computational pathology[51, 52], where increasing familiarity with decision-support tools may lead readers to rely more on their outputs and become less likely to challenge incorrect suggestions. It is therefore important that this risk is explicitly recognised and that laboratories adopting this technology implement quality-control practices designed to detect and monitor such effects over time. This could include the regular use of test slides containing diagnostically challenging mitotic mimics and for which the true mitotic count is already known. Such measures are likely to become increasingly important as the technology is adopted more widely and becomes embedded in routine practice[53].

Improvements in agreement, correlation and assessment time were reproduced in both HALO AP and Sectra, despite differences in interface and verification workflow.

The study design included several features that strengthen the reliability and generalisability of the findings. The dataset originated from three independent clinical centres from three different countries, covered seven tumour types, three scanner models and variation in patient and specimen characteristics. The paired reader design allowed routine and AI-assisted assessment to be compared using the same readers for each slide, reducing the influence of differences in reader composition between stages. Detailed analyses demonstrated findings translate across to a second image management system, providing further evidence that the observed benefits were consistent beyond the primary study setting. Similar results across 76 glioblastoma and non-small cell lung cancer slides suggest that the observed benefits of MitPro were not limited to the tumour types included in the primary study (Supplementary Section S2.2).

Several limitations should be considered. In the primary analysis, each slide was assessed by three pathologists, which may not capture the full range of inter-pathologist variability. The study was powered at the level of the main tumour groups rather than individual histological subtypes. Findings for less common subtypes should therefore be interpreted as supportive evidence that the technology is likely to generalise to these rarer tumour types, rather than as definitive evidence of performance within each subtype. Although the dataset included variation in clinical site, staining, tissue preparation and scanner source, it cannot represent every combination of specimen characteristics, laboratory processes and technical configurations encountered in practice.

Against this background, the findings were consistent across the primary and supporting analyses. AI assistance improved inter-pathologist agreement, increased agreement in mitotic scores and reduced assessment time across a broad range of tumour groups and clinical settings. The shift towards higher mitotic counts was consistent with the way MitPro supports hotspot selection and mitotic figure detection, while the frequency of resulting score changes remained broadly comparable with the variation already observed between pathologists during routine assessment. These findings support MitPro as a clinically useful decision-support tool for mitotic assessment.

# Acknowledgements

This work was supported by Innovate UK (10040491) and Histofy Ltd. The authors acknowledge the DAISY project at Fundación MD Anderson Cancer Center España for supporting the contribution of MD Anderson Cancer Center Madrid to this multicentre validation study.

# Author Contributions

SG led the development of MitPro, conducted the study and drafted the manuscript. MJ performed the statistical analysis. QDV led the integration of MitPro within the image management systems. VM, DP, RBV, AGL, JHO, AON, SC, RA and BSC participated as pathologists in the study. JLSR, SN and AL were principal investigators at the participating study sites. NR and DS provided oversight of both the study and the development of MitPro. All authors reviewed and approved the final manuscript.

# Declaration of competing interests

NR and DS are directors and shareholders of Histofy Ltd. SG is a shareholder of Histofy Ltd.

# Ethical Approval

Ethical approval for the study was granted by the London - Fulham Research Ethics Committee on 17 February 2025, with a favourable opinion issued on 20 February 2025. The study was approved under REC reference 25/LO/0158, protocol number MitPro 2024 and IRAS project ID 342246.

# References

1. Aaltomaa S, Lipponen P, Eskelinen M, et al. Mitotic indexes as prognostic predictors in female breast cancer. *Journal of cancer research and clinical oncology.* 1992. https://doi.org/https://doi.org/10.1007/BF01192316.
2. Andre F, Khalil A, Slimane K, et al. Mitotic index and benefit of adjuvant anthracycline-based chemotherapy in patients with early breast cancer. *The Journal of Clinical Oncology.* 2005. https://doi.org/https://doi.org/10.1200/JCO.2005.08.046.
3. Chen T, Ye L-Y, Feng X-Y, et al. Performance of risk stratification systems for gastrointestinal stromal tumors: A multicenter study. *World Journal of Gastroenterology.* 2019;25(10):1238.
4. Duhig EE, Clarke BE, Godbolt D, et al. The application of mitotic activity and necrosis to lung adenocarcinoma and correlation with the IASLC/ATS/ ERS multidisciplinary classification of adenocarcinoma in stage 1 disease. *Journal of Thoracic Oncology.* 2012. https://doi.org/10.1097/JTO.0b013e318268ff90.
5. Fidanzi C, Bevilacqua M, D'Erme AM, et al. Mitotic rate as a predictive factor for positive sentinel lymph nodes in pT1 and pT2 melanomas. *Journal of Plastic, Reconstructive and Aesthetic Surgery.* 2023. https://doi.org/10.1016/j.bjps.2023.04.085.
6. Hardell E, Josefson S, Ghaderi M, et al. Validation of a Mitotic Index Cutoff as a Prognostic Marker in Undifferentiated Uterine Sarcomas. *The American journal of surgical pathology.* 2017. https://doi.org/10.1097/PAS.0000000000000894.
7. Kashani-Sabet M, Miller III JR, Lo S, et al. Reappraisal of the prognostic significance of mitotic rate supports its reincorporation into the melanoma staging system. *Cancer.* 2020;126(21):4717–4725.
8. Kronqvist PKTCY. Quantitative thresholds for mitotic counts in histologic grading: Confirmation in nonfrozen samples of invasive ductal breast cancer. *Annals of Diagnostic Pathology.* 2000. https://doi.org/10.1016/S1092-9134%2800%2990013-3.
9. Lino-Silva LS, Segales-Rojas P, Aguilar-Cruz E, Salcedo-Hernandez RA, Zepeda-Najar C. Gastrointestinal Stromal Tumors Risk of Recurrence Stratification by Tumor Volume is a Best Predictor Compared with Risk Based on Mitosis and Tumor Size. *Journal of Gastrointestinal Cancer.* 2019. https://doi.org/10.1007/s12029-018-0115-2.
10. Pennanen M, Heiskanen I, Sane T, et al. Helsinki score - A novel model for prediction of metastases in adrenocortical carcinomas. *Human Pathology.* 2015. https://doi.org/10.1016/j.humpath.2014.11.015.
11. Pinero-Madrona A, Ruiz-Merino G, Cerezuela Fuentes P, Martinez-Barba E, Rodriguez-Lopez JN, Cabezas-Herrera J. Mitotic rate as an important prognostic factor in cutaneous malignant melanoma. *Clinical and Translational Oncology.* 2019. https://doi.org/10.1007/s12094-019-02064-4.
12. *WHO Classification of Tumours Editorial Board. Breast tumours [Internet]. Lyon (France): International Agency for Research on Cancer. (WHO classification of tumours series, 5th ed.; vol. 2). Available from: https://tumourclassification.iarc.who.int/chapters/32.2019*.
13. *WHO Classification of Tumours Editorial Board. Central nervous system tumours [Internet]. Lyon (France): International Agency for Research on Cancer. (WHO classification of tumours series, 5th ed.; vol. 6). Available from: https://tumourclassification.iarc.who.int/chapters/45.2021*.
14. *WHO Classification of Tumours Editorial Board. Soft tissue and bone tumours [Internet]. Lyon (France): International Agency for Research on Cancer. (WHO classification of tumours series, 5th ed.; vol. 3). Available from: https://tumourclassification.iarc.who.int/chapters/33.2020*.
15. *WHO Classification of Tumours Editorial Board. Endocrine and neuroendocrine tumours [Internet]. Lyon (France): International Agency for Research on Cancer. (WHO classification of tumours series, 5th ed.; vol. 10). Available from: https://tumourclassification.iarc.who.int/chapters/53.2025*.
16. Cree IA, Tan PH, Travis WD, et al. Counting mitoses: SI(ze) matters! *Mod Pathol.* 2021. https://doi.org/10.1038/s41379-021-00825-7.
17. Ibrahim A, Lashen A, Toss M, Mihai R, Rakha E. Assessment of mitotic activity in breast cancer: revisited in the digital pathology era. *J Clin Pathol.* 2022. https://doi.org/10.1136/jclinpath-2021-207742.
18. Meyer JS, Cosatto E, Graf HP. Mitotic index of invasive breast carcinoma: achieving clinically meaningful precision and evaluating tertial cutoffs. *Archives of pathology & laboratory medicine.* 2009. https://doi.org/https://doi.org/10.5858/133.11.1826.

19. Silverberg, S G. Reproducibility of the mitosis count in the histologic diagnosis of smooth muscle tumors of the uterus. *Human Pathology.* 1976. https://doi.org/https://doi.org/10.1016/S0046-8177(76)80058-5.
20. Soliman A, Li Z, Parwani AV. Artificial intelligence's impact on breast cancer pathology: a literature review. *Diagn Pathol.* 2024. https://doi.org/10.1186/s13000-024-01453-w.
21. van Diest PJ, Baak JP, Matze-Cok P, et al. Reproducibility of mitosis counting in 2,469 breast cancer specimens: results from the Multicenter Morphometric Mammary Carcinoma Project. *Human pathology.* 1992;23(6):603–607.
22. van Dooijeweert C, van Diest PJ, Ellis IO. Grading of invasive breast carcinoma: the way forward. *Virchows Arch.* 2022. https://doi.org/10.1007/s00428-021-03141-2.
23. Yigit N, Gunal A, Kucukodaci Z, Karslioglu Y, Onguru O, Ozcan A. Are we counting mitoses correctly? *Annals of Diagnostic Pathology.* 2013;17(6):536–539.
24. Lashen A, Ibrahim A, Katayama A, et al. Visual assessment of mitotic figures in breast cancer: a comparative study between light microscopy and whole slide images. *Histopathology.* 2021. https://doi.org/10.1111/his.14543.
25. Tabata K, Uraoka N, Benhamida J, et al. Validation of mitotic cell quantification via microscopy and multiple whole-slide scanners. *Diagn Pathol.* 2019. https://doi.org/10.1186/s13000-019-0839-8.
26. Wei BR, Halsey CH, Hoover SB, et al. Agreement in Histological Assessment of Mitotic Activity Between Microscopy and Digital Whole Slide Images Informs Conversion for Clinical Diagnosis. *Acad Pathol.* 2019. https://doi.org/10.1177/2374289519859841.
27. Akram SU, Qaiser T, Graham S, Kannala J, Heikkilä J, Rajpoot N. Leveraging unlabeled whole-slide-images for mitosis detection. *International Workshop on Ophthalmic Medical Image Analysis.* 2018. https://doi.org/https://doi.org/10.1007/978-3-030-00949-6_9.
28. Aubreville M, Stathonikos N, Bertram CA, et al. Mitosis domain generalization in histopathology images - The MIDOG challenge. *Med Image Anal.* 2023. https://doi.org/10.1016/j.media.2022.102699.
29. Aubreville M, Wilm F, Stathonikos N, et al. A comprehensive multi-domain dataset for mitotic figure detection. *Scientific data.* 2023;10(1):484.
30. Chen H, Dou Q, Wang X, Qin J, Heng PA. Mitosis detection in breast cancer histology images via deep cascaded networks. *Proceedings of the AAAI conference on artificial intelligence.* 2016. https://doi.org/https://doi.org/10.1609/aaai.v30i1.10140.
31. Jahanifar M, Shephard A, Zamanitajeddin N, et al. Mitosis detection, fast and slow: robust and efficient detection of mitotic figures. *Medical Image Analysis.* 2024;94:103132.
32. Kanadath A, J AAJ, Urolagin S. HVUNet: A hybrid vision transformer-based UNet for accurate detection and localization in histopathology images. *Comput Biol Med.* 2025. https://doi.org/10.1016/j.compbiomed.2025.110680.
33. Li C, Wang X, Liu W, Latecki LJ, Wang B, Huang J. Weakly supervised mitosis detection in breast histopathology images using concentric loss. *Medical Image Analysis.* 2019. https://doi.org/10.1016/j.media.2019.01.013.
34. Sebai M, Wang X, Wang T. MaskMitosis: a deep learning framework for fully supervised, weakly supervised, and unsupervised mitosis detection in histopathology images. *Medical and Biological Engineering and Computing.* 2020. https://doi.org/10.1007/s11517-020-02175-z.
35. Veta M, Heng YJ, Stathonikos N, et al. Predicting breast tumor proliferation from whole-slide images: the TUPAC16 challenge. *Medical image analysis.* 2019;54:111–121.
36. Jagdewsing DR, Murtaza G, Jagdewsing SA, et al. Evaluation of the Clinicopathological Features Associated With Malignancy of Phyllodes Tumor of the Breast. *Cureus.* 2024. https://doi.org/10.7759/cureus.76221.
37. Pantanowitz L, Hartman D, Qi Y, et al. Accuracy and efficiency of an artificial intelligence tool when counting breast mitoses. *Diagn Pathol.* 2020. https://doi.org/10.1186/s13000-020-00995-z.
38. Bertram CA, Aubreville M, Donovan TA, et al. Computer-assisted mitotic count using a deep learning–based algorithm improves interobserver reproducibility and accuracy. *Veterinary pathology.* 2022. https://doi.org/https://doi.org/10.1177/03009858211067478.
39. Balkenhol MC, Tellez D, Vreuls W, et al. Deep learning assisted mitotic counting for breast cancer. *Laboratory investigation.* 2019;99(11):1596–1606.
40. Tessier L, González-Gonzalo C, Tellez D, Bulten W, Balkenhol M. AI-assisted mitosis counting in breast cancer. A large-scale validation study. 2024.
41. Simmat C, Guichard L, Sockeel S, et al. Evaluating AI-Based Mitosis Detection for Breast Carcinoma in Digital Pathology: A Clinical Study on Routine Practice Integration. *Diagnostics (Basel).* 2025. https://doi.org/10.3390/diagnostics15091127.

42. Ronneberger O, Fischer P, Brox T. U-net: Convolutional networks for biomedical image segmentation. *International Conference on Medical image computing and computer-assisted intervention.* Springer; 2015:234–241.
43. Aubreville M, Bertram CA, Donovan TA, Marzahl C, Maier A, Klopfleisch R. A completely annotated whole slide image dataset of canine breast cancer to aid human breast cancer research. *Sci Data.* 2020. https://doi.org/10.1038/s41597-020-00756-z.
44. *WHO Classification of Tumours Editorial Board. Digestive system tumours [Internet]. Lyon (France): International Agency for Research on Cancer; 2019. (WHO classification of tumours series, 5th ed.; vol. 1). Available from: https://tumourclassification.iarc.who.int/chapters/31.*2019.
45. *WHO Classification of Tumours Editorial Board. Female genital tumours [Internet]. Lyon (France): International Agency for Research on Cancer. (WHO classification of tumours series, 5th ed.; vol. 4). Available from: https://tumourclassification.iarc.who.int/chapters/34.*2020.
46. *WHO Classification of Tumours Editorial Board. Thoracic tumours [Internet]. Lyon (France): International Agency for Research on Cancer. (WHO classification of tumours series, 5th ed.; vol. 5). Available from: https://tumourclassification.iarc.who.int/chapters/35.*2021.
47. Koo TK, Li MY. A guideline of selecting and reporting intraclass correlation coefficients for reliability research. *Journal of chiropractic medicine.* 2016;15(2):155–163.
48. Fisher C, Thway K. Royal College of Pathologists. Dataset for histopathological reporting of soft tissue sarcomas. May 2022.
49. Trojani M, Contesso G, Coindre J, et al. Soft-tissue sarcomas of adults; study of pathological prognostic variables and definition of a histopathological grading system. *International journal of cancer.* 1984;33(1):37–42.
50. Elston CW, Ellis IO. Pathological prognostic factors in breast cancer. I. The value of histological grade in breast cancer: experience from a large study with long-term follow-up. *Histopathology.* 1991;19(5):403–410.
51. Dratsch T, Chen X, Rezazade Mehrizi M, et al. Automation bias in mammography: the impact of artificial intelligence BI-RADS suggestions on reader performance. *Radiology.* 2023;307(4):e222176.
52. Rosbach E, Ammeling J, Ganz J, et al. Stuck on suggestions: automation bias, the anchoring effect, and the factors that shape them in computational pathology. *arXiv preprint arXiv:260311821.* 2026.
53. Parasuraman R, Manzey DH. Complacency and bias in human use of automation: An attentional integration. *Human factors.* 2010;52(3):381–410.

# Supplementary Material

## S1 Primary Analysis

### S1.1 Dataset

The eight study groups were represented by similar sample numbers, with breast biopsies and resections analysed separately. The distribution is shown in Figure S1.

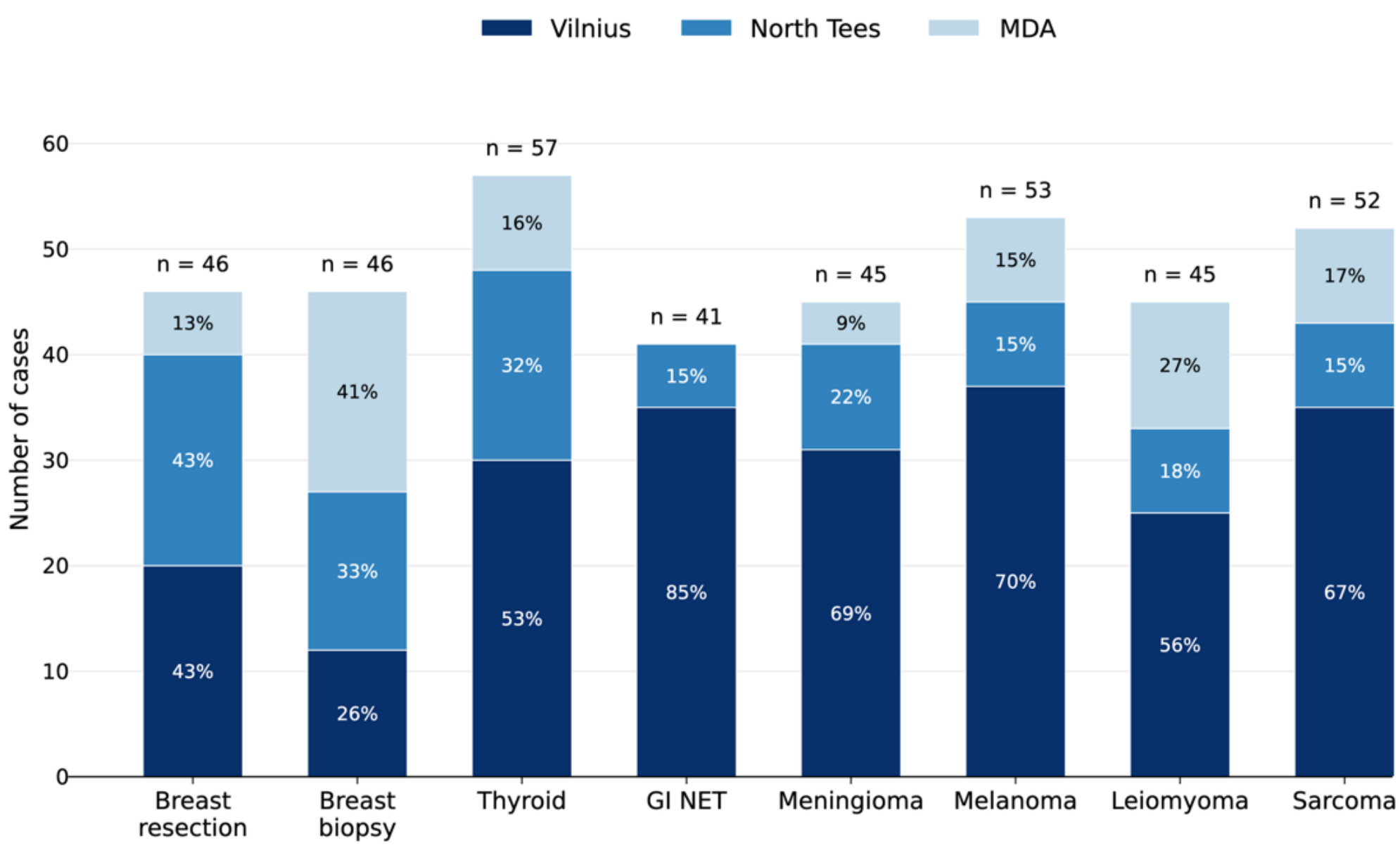

***Figure S1:*** *Composition of the clinical validation dataset by tumour category and contributing site. Stacked bars show the number of cases contributed by each site for each of the eight tumour groups included in the validation study. Percentages indicate each site's share of the category total; total case numbers per category (n) are shown above each bar. Across all categories, 385 cases were included.*

Samples were obtained from three participating sites. Vilnius contributed to every study group and provided the largest proportion of the dataset. North Tees also contributed across all groups, while MD Anderson Madrid contributed to most groups but did not provide gastrointestinal neuroendocrine tumour samples.

Subtypes with insufficient sample numbers or limited relevance to the study were excluded. These included papillary and metaplastic breast carcinoma and rhabdoid and angiomatous meningioma. Non-breast biopsy specimens were also excluded. Biopsy specimens were included only for breast carcinoma, while the other tumour groups were represented by surgical resection specimens.

The sample-size calculation established the minimum number of samples required for the primary analysis within each major tumour group. Dataset adequacy was also assessed by considering representation of supported histological subtypes, mitotic activity and patient demographics.

For breast carcinoma, both biopsy and resection specimens were included because grading is routinely performed on both. The cohort was predominantly infiltrating ductal carcinoma, with a smaller proportion of invasive lobular carcinoma. The meningioma cohort included mainly meningothelial, transitional and fibrous subtypes, with smaller numbers of atypical and anaplastic tumours. Gastrointestinal neuroendocrine tumours included cases across the range of mitotic activity relevant to grading. For thyroid tumours, papillary carcinoma was the most common subtype, with additional follicular, oncocytic and medullary tumours. Leiomyoma represented a comparatively homogeneous tumour group. The melanoma cohort included superficial spreading, nodular, lentigo maligna and acral lentiginous melanoma, together with malignant melanoma not otherwise specified. Melanoma was assessed using continuous mitotic counts rather than categorical mitotic scores. Sarcoma included a broad range of supported tumour types, including leiomyosarcoma, liposarcoma, fibrosarcoma and fibroblastic or myofibroblastic tumours, rhabdomyosarcoma, Ewing sarcoma-family tumours, synovial sarcoma, epithelioid sarcoma, clear cell sarcoma and malignant peripheral nerve sheath tumour.

Across tumour groups, all applicable mitotic score categories were represented, with Stage 1 score distributions shown in Figure S2. The cohort included a broad age range, with a higher proportion of female patients reflecting the inclusion of breast carcinoma and leiomyoma samples. The study was not intended to represent every recognised subtype, but to provide appropriate coverage of the supported tumour types and subtypes included in the clinical evaluation.

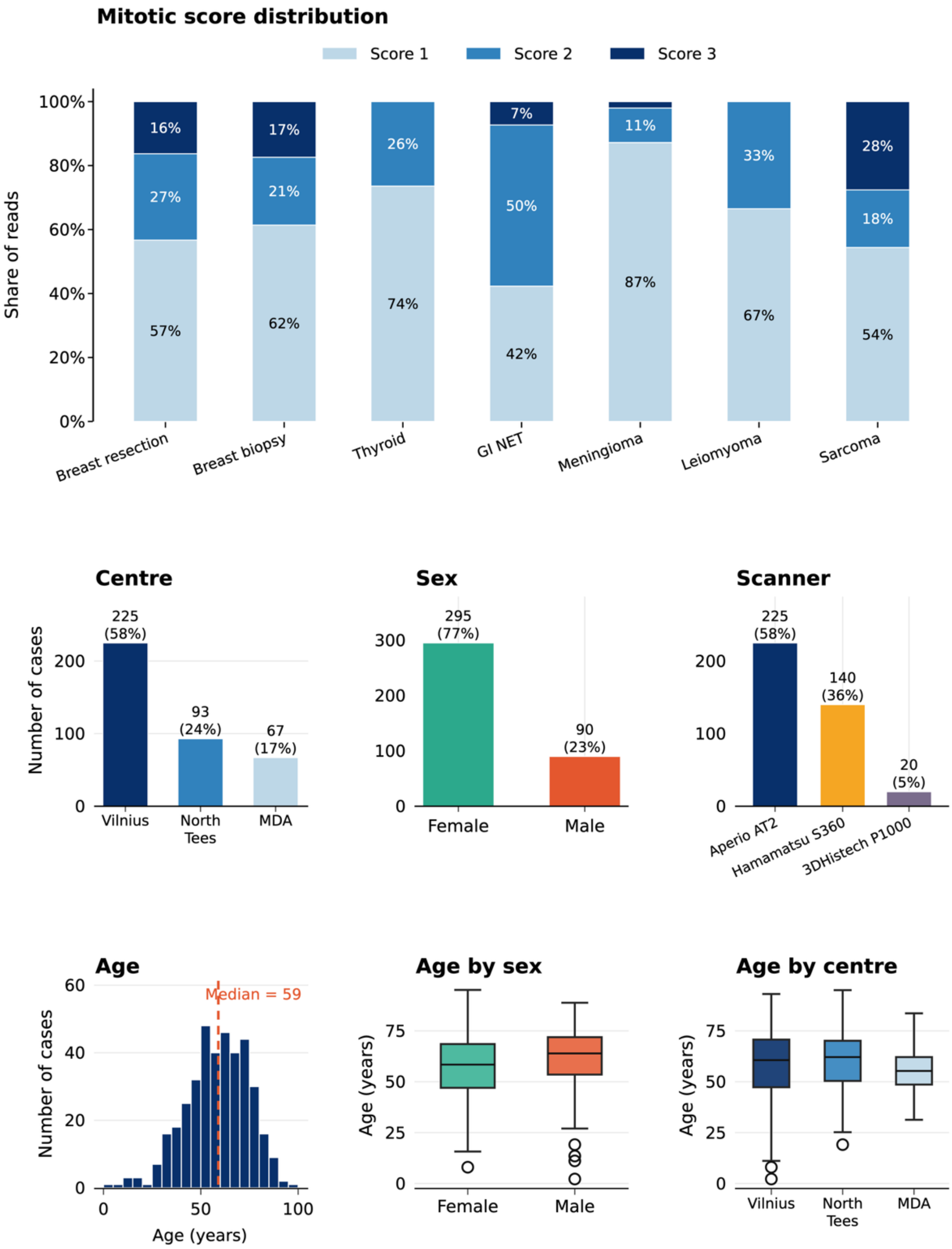

***Figure S2:*** *Clinical characteristics of the clinical validation cohort. Distribution of mitotic activity score, centre, sex, scanner, and age. Mitotic activity score refers to a grading component in several tumour types and to a diagnostic threshold in others, such as mitotically active leiomyoma. Score 3 is not defined for thyroid or leiomyoma. Melanoma is excluded, as no mitotic activity score is assigned for this tumour type.*

## S1.2 Interobserver agreement in mitotic count

In Figure S3 and Figure S4, we present the ICC results from the primary analysis by participating centre, sex, age group and scanner. ICC increased with AI assistance across all subgroups assessed. At the three participating centres, Stage 2 ICC ranged from 0.920 to 0.958, compared with 0.533 to 0.751 in Stage 1. Results were also similar between female and male patients, with Stage 2 ICC values of 0.936 and 0.961, respectively. Across age groups, Stage 2 ICC exceeded 0.920 in all three categories, compared with Stage 1 values ranging from 0.532 to 0.715. A similar pattern was observed across scanner groups, where Stage 2 ICC ranged from 0.924 to 0.980, compared with 0.533 to 0.723 in Stage 1.

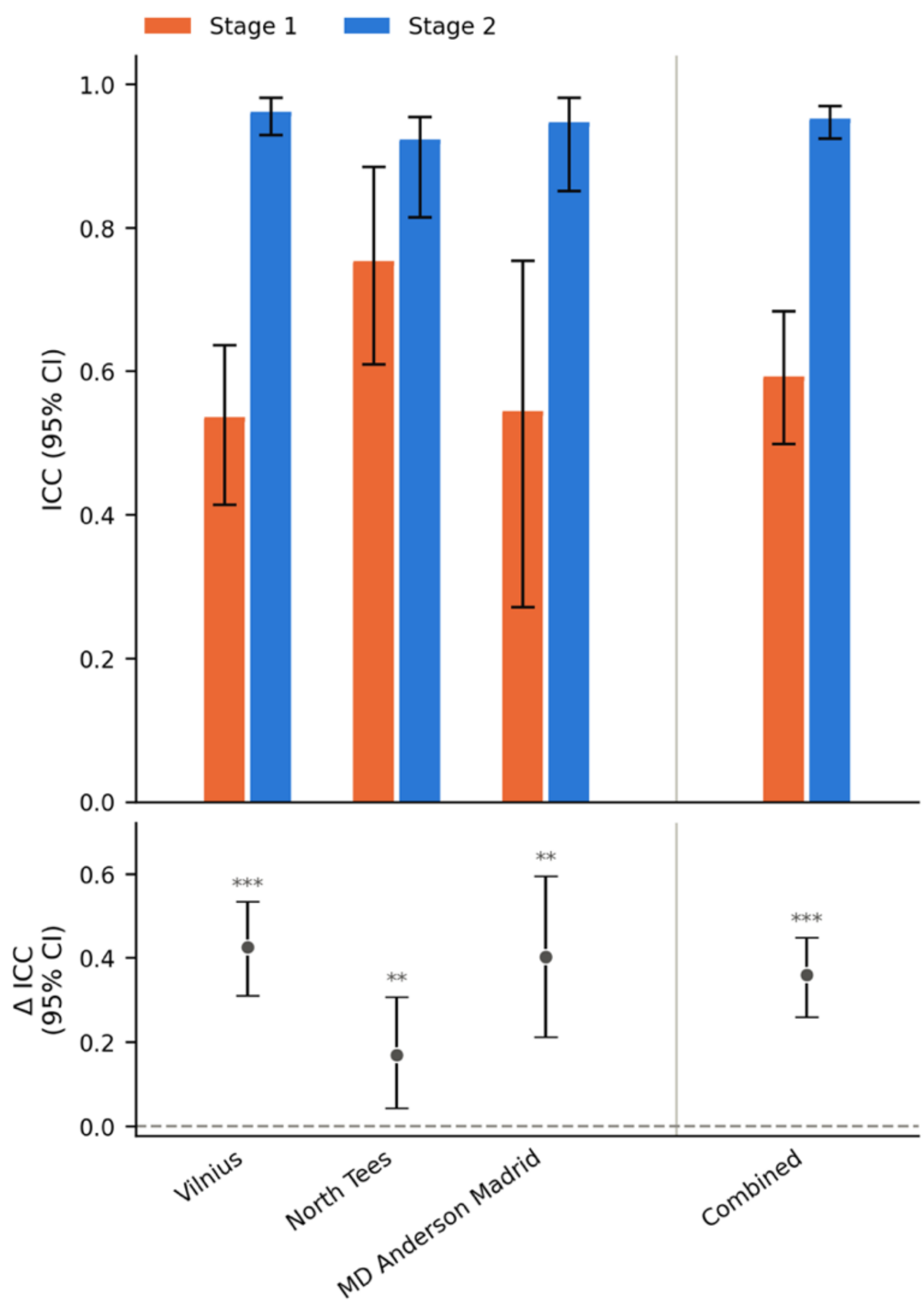

***Figure S3:*** *Inter-pathologist agreement (ICC) by participating centre during Stage 1 conventional mitotic counting and Stage 2 AI-assisted mitotic counting. The upper panel shows Stage 1 and Stage 2 ICC estimates with 95% confidence intervals, while the lower panel shows the between-stage difference (ΔICC = Stage 2 - Stage 1) with 95% confidence intervals. Statistical significance of the difference between Stage 1 and Stage 2 is indicated by asterisks: p < 0.05 = *, p < 0.01 = **, p < 0.001 = ***, and p < 0.0001 = ****.*

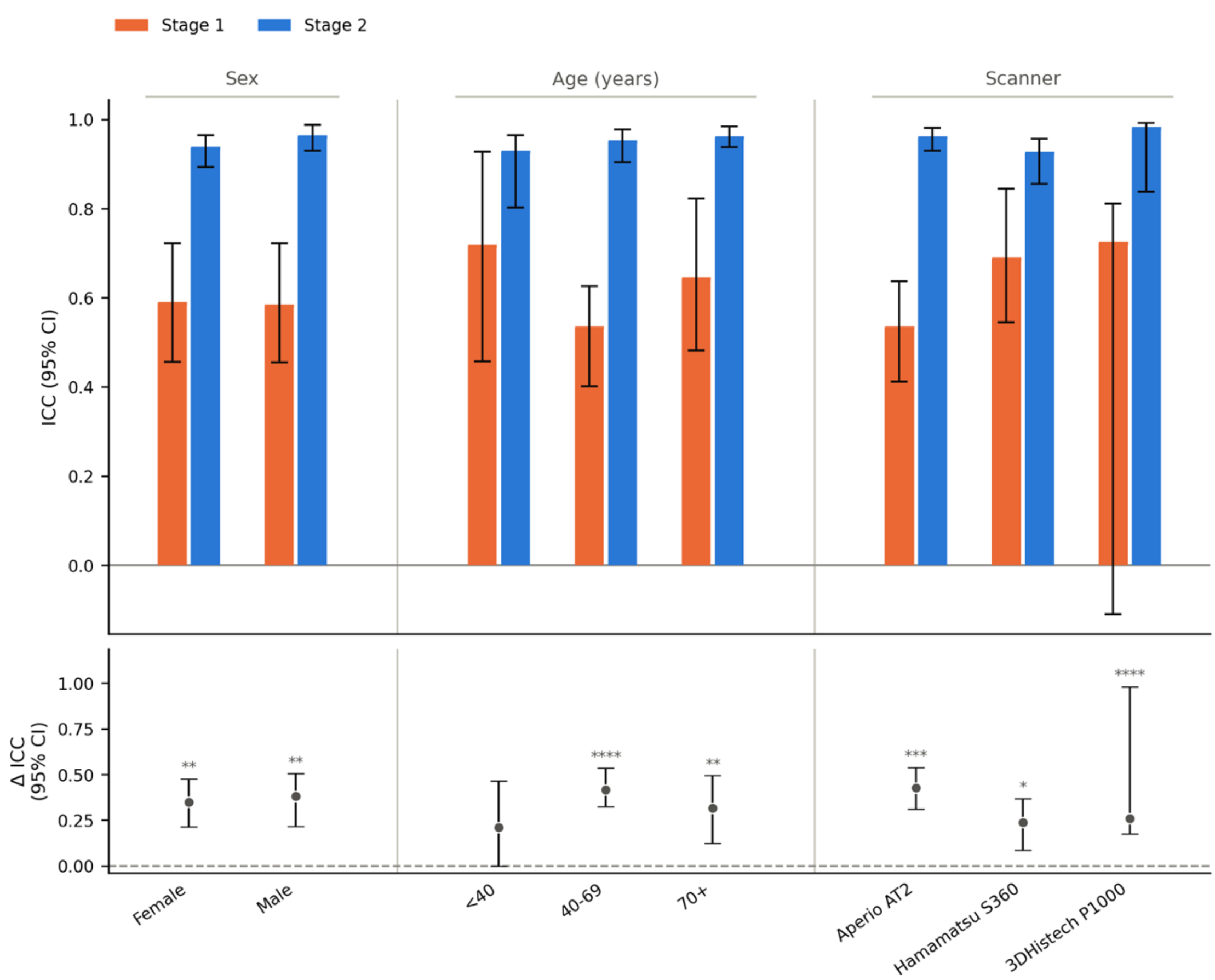

***Figure S4:*** *Inter-pathologist agreement (ICC) by sex, age group, and scanner type during Stage 1 conventional mitotic counting and Stage 2 AI-assisted mitotic counting. The upper panels show Stage 1 and Stage 2 ICC estimates with 95% confidence intervals, while the lower panels show the corresponding between-stage differences (ΔICC = Stage 2 - Stage 1) with 95% confidence intervals. Statistical significance of the difference between Stage 1 and Stage 2 is indicated by asterisks: $p < 0.05$ = *, $p < 0.01$ = **, $p < 0.001$ = ***, and $p < 0.0001$ = ****.*

## S1.3 Bonferroni-adjusted analysis on interobserver agreement

The primary analysis assessed the effect of AI assistance separately across eight tumour groups using the 5% significance level. As an additional analysis, a Bonferroni adjustment was applied to account for the eight tumour-specific comparisons and provide a more stringent assessment of statistical significance. The nominal significance level of 5% was divided by eight, giving an adjusted significance level of 0.625% and a corresponding confidence level of 99.375%.

After Bonferroni adjustment, the lower bound of the 99.375% confidence interval remained above zero for breast resection, melanoma, thyroid, leiomyoma and sarcoma. These five tumour groups therefore met the Bonferroni-adjusted criterion for superiority.

For breast biopsy, gastrointestinal neuroendocrine tumours and meningioma, the adjusted confidence intervals included zero, with lower bounds of -0.002, -0.012 and -0.068, respectively. These groups did not meet the stricter Bonferroni-adjusted criterion, although the estimated differences in ICC remained positive at 0.159, 0.371 and 0.323. For breast biopsy, the relatively high Stage 1 ICC of 0.753 may also have limited the scope for further improvement.

These findings should not be interpreted as evidence of no benefit in the three groups that did not meet the adjusted criterion. Bonferroni correction applies a more stringent threshold across multiple comparisons, increasing the level of evidence required for statistical significance. All three groups retained positive differences in ICC and showed improved inter-pathologist agreement at the conventional 95% confidence level.

## S1.4 Interobserver agreement on mitotic scores

The thresholds used to convert raw mitotic counts into scores for the applicable tumour types are summarised below and were aligned with the relevant clinical guidelines. For thyroid tumours, a single cut-off was applied across all samples for this analysis, although in routine practice the relevant mitotic threshold may vary according to morphology.

***Table S1:*** *Tumour-specific mitotic count thresholds used to convert raw mitotic counts into ordinal clinical scores.*

| Tumour type | Score 1 | Score 2 | Score 3 | Counting area |
|---|---|---|---|---|
| Breast carcinoma | ≤ 7 | ≥ 8 and ≤ 14 | ≥ 15 | 2 mm² |
| GI NET | ≤ 1 | ≥ 2 and ≤ 20 | ≥ 21 | 2 mm² |
| Meningioma | < 2.5 | ≥ 2.5 and < 12.5 | ≥ 12.5 | 1 mm² |
| Sarcoma | ≤ 9 | ≥ 10 and ≤ 19 | ≥ 20 | 2 mm² |
| Thyroid | ≤ 4 | > 4 | | 2 mm² |
| Leiomyoma | ≤ 1 | > 1 | | 1 mm² |

Numerical results for Fleiss' kappa are provided below in Table S2.

***Table S2:*** *Fleiss' kappa for Stage 1 and Stage 2 by tumour type, with between-stage differences (Δκ) and 95% confidence intervals.*

| Tumour type | Slides | Fleiss' $\kappa_{\text{Stage 1}}$ | Fleiss' $\kappa_{\text{Stage 2}}$ | Δ Fleiss' κ |
|---|---|---|---|---|
| Breast resection | 47 | 0.363 [0.204, 0.541] | 0.745 [0.621, 0.871] | 0.382 [0.188, 0.561] |
| Breast biopsy | 46 | 0.323 [0.143, 0.528] | 0.691 [0.540, 0.824] | 0.367 [0.209, 0.550] |
| Thyroid | 47 | 0.084 [-0.085, 0.327] | 0.590 [0.395, 0.763] | 0.506 [0.269, 0.732] |
| GI NET | 41 | 0.088 [-0.053, 0.298] | 0.496 [0.273, 0.728] | 0.408 [0.156, 0.626] |
| Meningioma | 47 | 0.282 [0.033, 0.627] | 0.663 [0.407, 0.860] | 0.381 [-0.043, 0.691] |
| Leiomyoma | 45 | 0.300 [0.110, 0.524] | 0.423 [0.231, 0.627] | 0.123 [-0.156, 0.377] |
| Sarcoma | 52 | 0.407 [0.271, 0.561] | 0.892 [0.778, 0.959] | 0.485 [0.309, 0.639] |

## S1.5 Pairwise pathologist correlation

Correlation was higher in Stage 2 than in Stage 1 for every group. Across the combined dataset, the pooled correlation increased from 0.764 during routine counting to 0.975 during AI-assisted counting. The Stage 2 correlation indicates a very strong relationship between pathologists' mitotic counts during AI-assisted assessment. The 95% confidence interval for the difference was entirely above zero, indicating higher pairwise reader consistency during Stage 2.

Pairwise correlation increased with AI assistance across all tumour groups and centres. The largest increase was observed for thyroid, while high Stage 2 correlations were also seen across the other tumour groups. The 95% confidence interval for the between-stage difference remained entirely above zero in every tumour group and centre, supporting the ICC findings of greater interobserver consistency with AI assistance.

Pairwise correlation also increased at each centre. The largest increase was observed at MD Anderson Madrid, where correlation increased from 0.737 in Stage 1 to 0.975 in Stage 2. At Vilnius, correlation increased from 0.762 to 0.976, while at North Tees it increased from 0.848 to 0.982. The smaller increase observed at North Tees reflected the relatively high correlation already present during Stage 1. Nevertheless, Stage 2 correlations exceeded 0.97 at all three centres, indicating a consistently very strong relationship between pathologists' mitotic counts during AI-assisted counting.

Numerical results are provided in Table S3, with centre-level results shown in Figure S5.

***Table S3:*** *Pooled pairwise Pearson correlations for Stage 1 and Stage 2, overall and by centre and tumour type, with between-stage differences (Δr) and 95% confidence intervals.*

| **Group** | **Slides** | **r** $_{\text{Stage 1}}$ | **r** $_{\text{Stage 2}}$ | **Δ r** |
|---|---|---|---|---|
| Combined | 385 | 0.764 [0.698, 0.829] | 0.975 [0.959, 0.985] | 0.211 [0.159, 0.284] |
| Vilnius | 225 | 0.762 [0.674, 0.844] | 0.976 [0.945, 0.988] | 0.214 [0.153, 0.313] |
| North Tees | 93 | 0.848 [0.736, 0.916] | 0.982 [0.969, 0.990] | 0.134 [0.062, 0.221] |
| MD Anderson Madrid | 67 | 0.737 [0.556, 0.854] | 0.975 [0.885, 0.989] | 0.238 [0.131, 0.367] |
| Breast resection | 46 | 0.758 [0.437, 0.886] | 0.977 [0.894, 0.991] | 0.218 [0.073, 0.430] |
| Breast biopsy | 46 | 0.755 [0.521, 0.885] | 0.919 [0.854, 0.970] | 0.165 [0.056, 0.340] |
| Melanoma | 53 | 0.783 [0.657, 0.875] | 0.912 [0.843, 0.962] | 0.129 [0.033, 0.247] |
| Thyroid | 57 | 0.105 [-0.062, 0.342] | 0.764 [0.682, 0.883] | 0.659 [0.473, 0.887] |
| GI NET | 41 | 0.801 [0.400, 0.941] | 0.980 [0.689, 0.996] | 0.180 [0.058, 0.477] |
| Meningioma | 45 | 0.573 [0.251, 0.793] | 0.950 [0.843, 0.980] | 0.377 [0.182, 0.631] |
| Leiomyoma | 45 | 0.744 [0.248, 0.902] | 0.968 [0.846, 0.990] | 0.223 [0.062, 0.598] |
| Sarcoma | 52 | 0.839 [0.749, 0.915] | 0.984 [0.969, 0.997] | 0.145 [0.089, 0.283] |

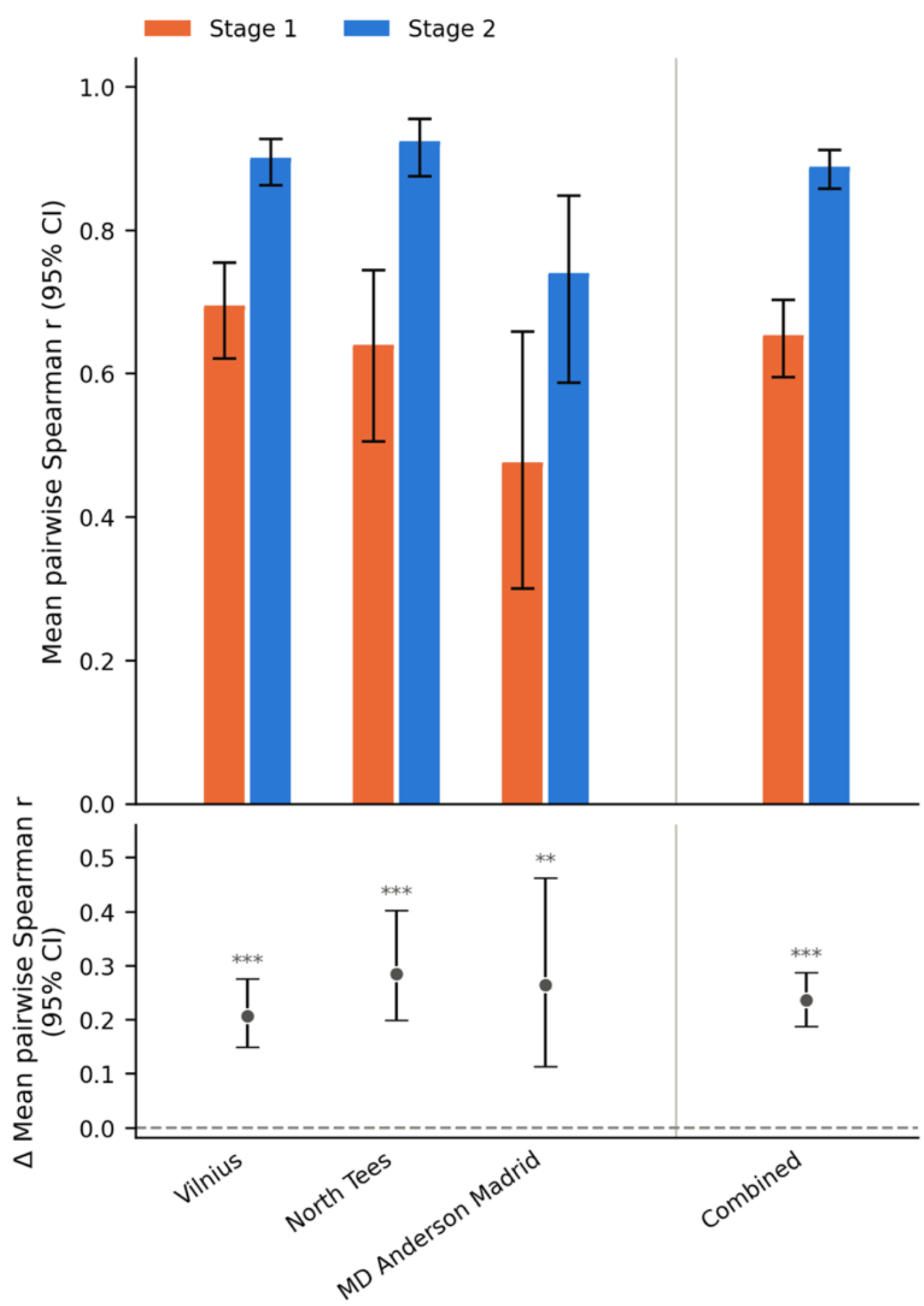

***Figure S5****: Pairwise Pearson correlation by participating centre. The upper panel shows the pooled Pearson correlation coefficients for Stage 1 and Stage 2 for each participating centre, with whiskers representing 95% confidence intervals. The lower panel shows the corresponding between-stage difference (Δr = Stage 2 - Stage 1) with 95% confidence intervals. Statistical significance of the difference between Stage 1 and Stage 2 is indicated by asterisks: p < 0.05 = *, p < 0.01 = **, p < 0.001 = ***, and p < 0.0001 = ****.*

Pairwise Pearson correlation increased in Stage 2 across all sex, age and scanners. The 95% confidence interval for the between-stage difference remained entirely above zero in every group, indicating that the increase was observed consistently across the demographic and technical conditions assessed rather than being driven by a particular patient group or scanner.

We summarise the pairwise Pearson correlation results by sex, age group and scanner in Table S4 and Figure S6. Correlation increased with AI assistance across all subgroups assessed. Stage 2 correlation was high in both female and male patients, at 0.970 and 0.982, respectively. Across age groups, Stage 2 values ranged from 0.973 to 0.985, compared with 0.747 to 0.872 in Stage 1. A similar pattern was observed across scanners, with Stage 2 correlations of at least 0.975 for all three scanner types. The confidence intervals for the differences were above zero across all sex, age and scanner groups.

***Table S4:*** *Pooled pairwise Pearson correlations for Stage 1 and Stage 2 by sex, age group, and scanner, with between-stage differences (Δr) and 95% confidence intervals.*

| Group | Slides | $r_{\text{Stage 1}}$ | $r_{\text{Stage 2}}$ | Δ r |
|---|---|---|---|---|
| Female | 295 | 0.754 [0.655, 0.824] | 0.970 [0.935, 0.982] | 0.217 [0.151, 0.296] |
| Male | 90 | 0.821 [0.686, 0.894] | 0.982 [0.957, 0.993] | 0.162 [0.104, 0.318] |
| <40 | 52 | 0.872 [0.726, 0.950] | 0.973 [0.914, 0.992] | 0.100 [0.045, 0.262] |
| 40–69 | 231 | 0.747 [0.655, 0.834] | 0.976 [0.948, 0.987] | 0.228 [0.162, 0.329] |
| 70+ | 102 | 0.799 [0.641, 0.895] | 0.985 [0.952, 0.994] | 0.186 [0.090, 0.318] |
| Aperio AT2 | 225 | 0.762 [0.669, 0.846] | 0.976 [0.944, 0.987] | 0.214 [0.152, 0.314] |
| Hamamatsu NanoZoomer S360 | 140 | 0.793 [0.702, 0.862] | 0.975 [0.952, 0.986] | 0.182 [0.117, 0.257] |
| 3DHistech P1000 | 20 | 0.824 [0.443, 0.954] | 0.975 [0.921, 0.994] | 0.152 [0.029, 0.438] |

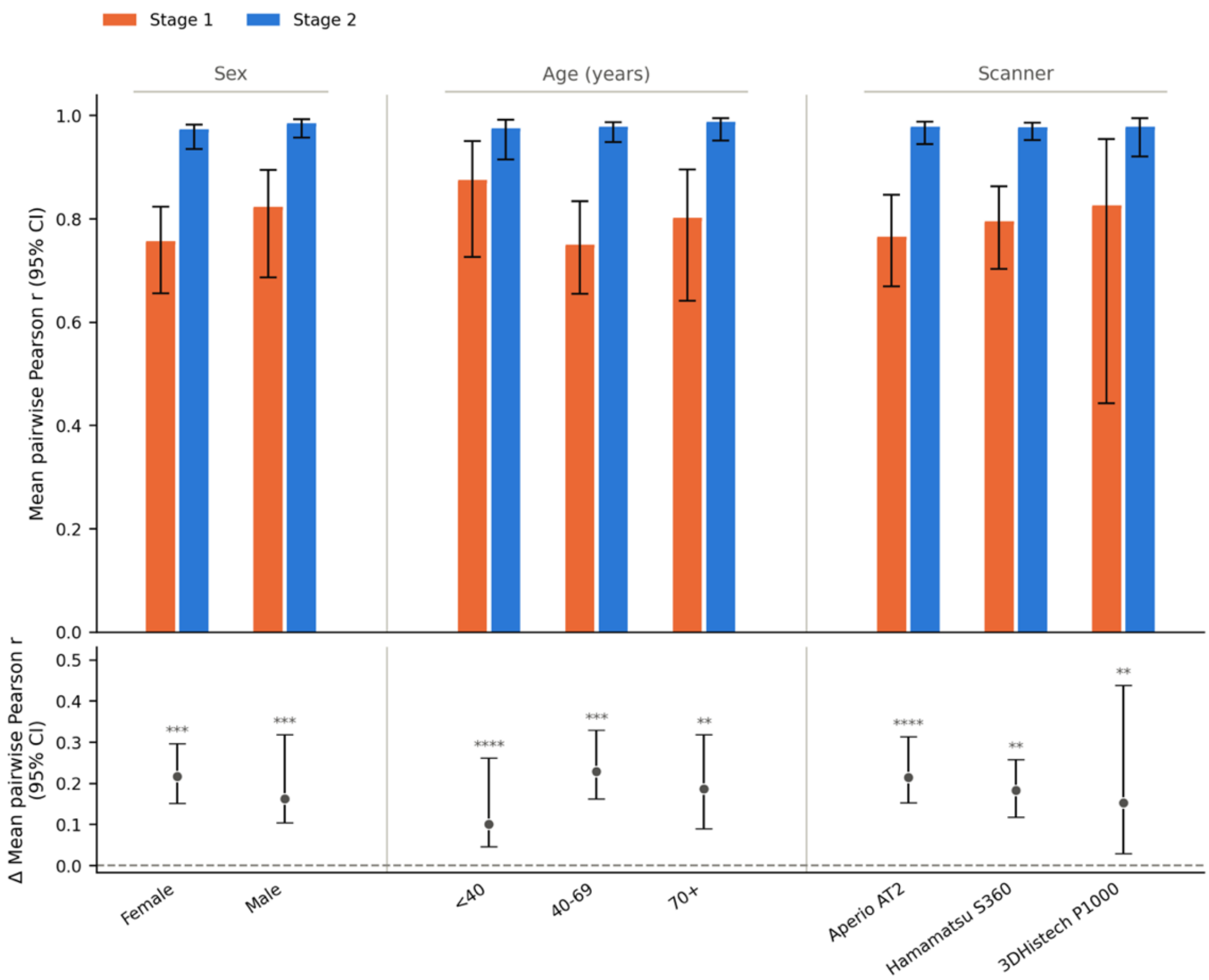

***Figure S6:*** *Pairwise Pearson correlation stratified by sex, age group, and scanner type. The upper panel shows the pooled Pearson correlation coefficients for Stage 1 and Stage 2 within each subgroup, with whiskers representing 95% confidence intervals. The lower panel shows the corresponding between-stage difference (Δr = Stage 2 - Stage 1) with 95% confidence intervals. Statistical significance of the difference between Stage 1 and Stage 2 is indicated by asterisks: p < 0.05 = *, p < 0.01 = **, p < 0.001 = ***, and p < 0.0001 = ****.*

## S1.6 Mitotic count assessment time

Results for mitotic count assessment time in Stage 1 and Stage 2, together with the estimated time saved, are provided in Table S5 for the overall dataset and by participating centre, tumour group, sex, age group and scanner. Across all 385 slides, the average pathologist-level median assessment time decreased from 286.4 seconds in Stage 1 to 127.8 seconds in Stage 2, with an estimated average time saving of 151.8 seconds, or approximately 2.5 minutes per assessment. Time was lower in Stage 2 across all analysis groups. The largest tumour-group time savings were observed for sarcoma and meningioma, at 214.6 seconds and 208.8 seconds, respectively. Similar reductions were also observed across sex, age and scanner groups. These subgroup results are shown visually in Figure S7.

***Table S5:*** *Stage 1 and Stage 2 mitotic count assessment times and paired time savings. Stage-specific values represent the median assessment time across slides within each group. Time saved was calculated from the paired Stage 1 and Stage 2 assessment times for each slide and is reported as the median of these paired differences, with 95% confidence intervals. It should therefore not be interpreted as the arithmetic difference between the reported Stage 1 and Stage 2 medians.*

| Group | Slides | Median Stage 1 (s) | Median Stage 2 (s) | Time saved (s) |
|---|---|---|---|---|
| Overall | 385 | 286.4 [276.9, 297.1] | 127.8 [121.3, 135.2] | 151.8 [144.1, 161.8] |
| Vilnius | 225 | 315.9 [297.5, 343.7] | 136.3 [124.2, 151.2] | 169.5 [156.5, 192.2] |
| North Tees | 93 | 289.8 [278.3, 305.9] | 146.6 [136.4, 157.1] | 143.6 [128.3, 158.8] |
| MD Anderson Madrid | 63 | 193.3 [179.3, 210.2] | 128.0 [113.0, 137.0] | 67.7 [56.7, 78.0] |
| Breast resection | 46 | 268.4 [251.6, 305.4] | 132.3 [118.1, 146.4] | 146.9 [122.3, 172.2] |
| Breast biopsy | 46 | 257.7 [238.2, 292.5] | 144.0 [121.7, 159.3] | 118.1 [94.8, 149.9] |
| Melanoma | 53 | 233.8 [215.9, 255.4] | 92.7 [84.7, 105.2] | 137.4 [115.5, 160.8] |
| Thyroid | 57 | 289.1 [268.4, 311.3] | 121.0 [108.3, 141.2] | 167.5 [146.7, 186.0] |
| GI NET | 41 | 335.4 [298.5, 393.3] | 136.9 [122.1, 169.1] | 190.9 [142.3, 252.2] |
| Meningioma | 45 | 375.8 [343.6, 413.6] | 145.2 [122.6, 196.4] | 208.8 [181.7, 251.6] |
| Leiomyoma | 45 | 308.8 [276.5, 339.8] | 123.9 [106.4, 148.1] | 180.4 [142.1, 205.7] |
| Sarcoma | 52 | 355.0 [311.5, 395.2] | 138.9 [122.1, 162.9] | 214.6 [174.4, 242.4] |
| Female | 295 | 227.0 [212.0, 240.0] | 98.0 [90.0, 105.0] | 111.0 [104.0, 120.6] |
| Male | 90 | 238.0 [200.0, 272.5] | 105.0 [87.0, 120.0] | 117.0 [107.0, 132.0] |
| <40 | 52 | 240.0 [200.0, 277.5] | 83.5 [63.0, 100.5] | 126.0 [106.0, 146.5] |
| 40-69 | 231 | 225.0 [210.0, 240.0] | 103.0 [95.0, 114.0] | 111.0 [100.0, 120.0] |
| 70+ | 102 | 227.0 [201.0, 257.0] | 100.0 [86.0, 115.5] | 113.0 [98.0, 127.0] |
| Aperio AT2 | 225 | 230.0 [209.2, 242.0] | 78.0 [72.0, 84.0] | 127.5 [118.0, 140.0] |
| Hamamatsu NanoZoomer S360 | 140 | 215.5 [201.0, 238.5] | 142.0 [131.0, 151.5] | 91.5 [81.0, 103.0] |
| 3DHistech P1000 | 20 | 300.0 [251.0, 316.0] | 150.5 [136.0, 165.5] | 137.0 [115.0, 173.5] |

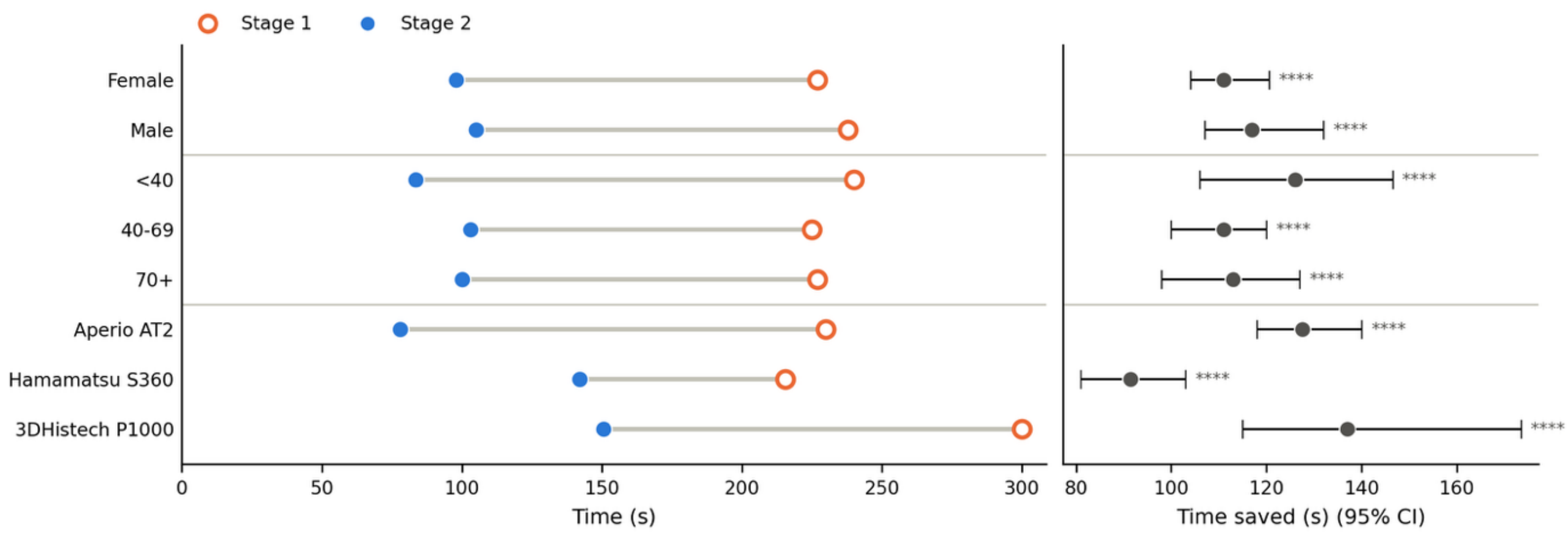

***Figure S7:*** *Median assessment time in Stage 1 and Stage 2 by sex, age group, and scanner type. Results represent the mean of the pathologist-level median assessment times within each subgroup. Time saved is calculated as Stage 1 minus Stage 2. Statistical significance of the between-stage difference is indicated by asterisks: $p < 0.05$ = *, $p < 0.01$ = **, $p < 0.001$ = ***, and $p < 0.0001$ = ****.*

Table S6 presents the corresponding results for the 13 pathologists included in the primary analysis. This provides a reader-level view of the change in assessment time and shows how the amount of time saved varied between individual pathologists.

***Table S6:*** *Stage 1 and Stage 2 mitotic count assessment times and paired time savings. Stage-specific values represent the median assessment time across slides within each group. Time saved was calculated from the paired Stage 1 and Stage 2 assessment times for each slide and is reported as the median of these paired differences, with 95% confidence intervals. It should therefore not be interpreted as the arithmetic difference between the reported Stage 1 and Stage 2 medians.*

| Pathologist | Slides | Median Stage 1 (s) | Median Stage 2 (s) | Time saved (s) |
|---|---|---|---|---|
| A | 88 | 150.0 [130.0, 160.0] | 120.0 [97.5, 145.0] | 30.0 [1.5, 40.5] |
| B | 235 | 150.0 [140.0, 164.0] | 75.0 [64.0, 83.0] | 85.0 [74.0, 98.0] |
| C | 184 | 108.5 [99.0, 122.0] | 40.0 [38.0, 43.0] | 68.0 [59.5, 75.5] |
| D | 235 | 476.0 [430.0, 516.0] | 78.0 [71.0, 86.0] | 374.0 [342.0, 417.0] |
| E | 95 | 689.0 [654.0, 753.0] | 305.0 [283.0, 328.0] | 380.0 [364.0, 431.0] |
| F | 44 | 270.0 [247.5, 290.0] | 184.0 [166.0, 202.5] | 77.5 [48.0, 119.0] |
| G | 44 | 336.2 [299.0, 381.0] | 82.5 [71.0, 99.0] | 249.0 [205.1, 298.0] |
| H | 44 | 396.5 [353.0, 481.0] | 135.5 [114.0, 185.0] | 216.5 [168.5, 307.0] |
| I | 131 | 410.0 [358.0, 446.0] | 212.0 [194.0, 236.0] | 213.0 [173.0, 236.0] |
| J | 133 | 298.0 [281.0, 300.0] | 177.0 [167.0, 184.0] | 110.0 [100.0, 122.0] |
| K | 77 | 124.0 [100.0, 137.0] | 44.0 [38.0, 55.0] | 67.0 [55.0, 92.0] |
| L | 63 | 130.0 [110.0, 156.0] | 55.0 [42.0, 62.0] | 74.0 [54.0, 102.0] |
| M | 63 | 185.0 [156.0, 207.0] | 154.0 [119.0, 171.0] | 30.0 [13.0, 45.0] |

## S1.7 AI-assisted hotspots

Across all assessments, pathologists selected one of the AI-proposed hotspots in 87.2% of cases and defined a manual hotspot in 12.8%. Use of an AI-proposed hotspot was high across all tumour groups, ranging from 81.2% for thyroid tumours to 92.2% for meningioma. This indicates that the proposed regions were considered suitable for mitotic counting in most assessments, while the option to define an alternative region remained important. Results are provided in Table S7.

Results for hotspot selection by sex, age group and scanner are provided in Table S8 and Figure S8.

***Table S7:*** *Proportion of assessments using a MitPro-proposed or manually defined hotspot by tumour group.*

| Group | MitPro hotspot (%) | Manual hotspot (%) |
|---|---|---|
| Breast resection | 91.2 | 8.8 |
| Breast biopsy | 84.8 | 15.2 |
| Melanoma | 82.5 | 17.5 |
| Thyroid | 81.2 | 18.8 |
| GI NET | 85.3 | 14.7 |
| Meningioma | 92.2 | 7.8 |
| Leiomyoma | 89.9 | 10.1 |
| Sarcoma | 91.7 | 8.3 |
| **Combined** | **87.2** | **12.8** |

***Table S8:*** *Proportion of assessments using a MitPro-proposed or manually defined hotspot by sex, age group and scanner.*

| Group | MitPro hotspot (%) | Manual hotspot (%) |
|---|---|---|
| Female | 85.9 | 14.1 |
| Male | 91.4 | 8.6 |
| <40 | 86.7 | 13.3 |
| 40–69 | 85.4 | 14.6 |
| 70+ | 91.5 | 8.5 |
| Aperio AT2 | 87.1 | 12.9 |
| Hamamatsu NanoZoomer S360 | 88.0 | 12.0 |
| 3DHistech P1000 | 82.9 | 17.1 |

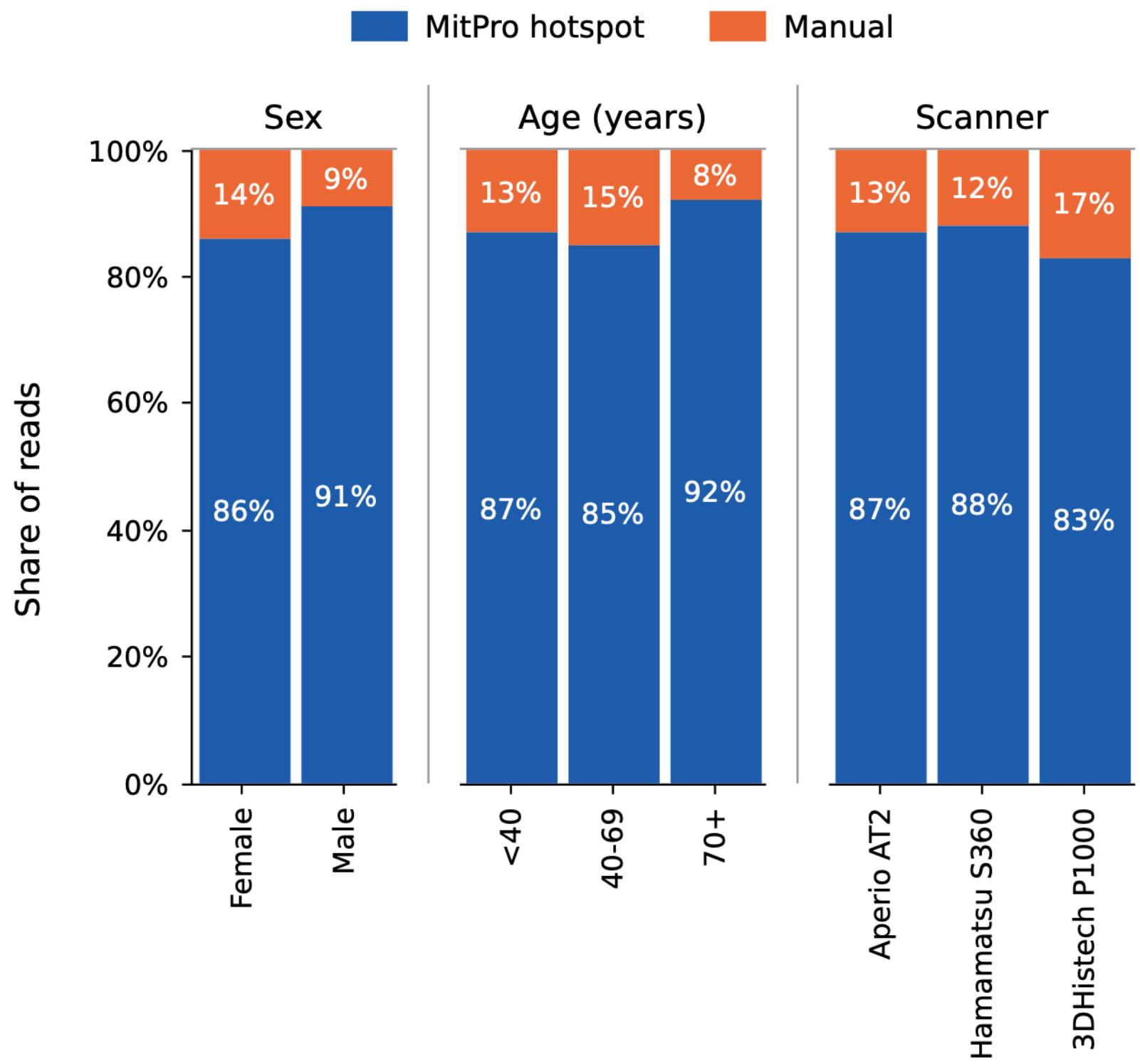

***Figure S8:*** *Hotspot selection by sex, age group, and scanner type. Bars show the proportion of assessments in which a MitPro-proposed hotspot was selected, or an alternative hotspot was manually defined.*

## S1.8 Impact on mitotic count and score

Below, we show the distributions of mitotic counts in Stage 1 and Stage 2 for each tumour group. Overall, there was a slight shift towards higher mitotic counts during AI-assisted assessment, although the amount of change varied between tumour groups. The larger shifts were generally observed in tumour groups with higher mitotic activity in our dataset, where more individual mitotic figures were encountered and therefore more opportunities existed for differences in hotspot selection or missed figures during routine counting. Sarcoma was a notable example and showed the largest shift between stages. For all eight groups, the Stage 2 mean remained within one Stage 1 standard deviation of the corresponding Stage 1 mean, indicating that the observed shifts remained within the underlying variability of the Stage 1 distributions.

Mitotic counts were also converted into tumour-specific scores using the thresholds defined in Table S1. We then assessed how the assigned score changed between Stage 1 and Stage 2. Results are provided in Table S9, with the corresponding distributions shown in Figure S9.

Across the combined dataset, most mitotic scores remained unchanged between stages. Where scores changed, increases were more common than decreases. This pattern is consistent with AI assistance directing pathologists towards regions with higher mitotic activity and helping to identify mitotic figures that may otherwise have been missed, although the study was not designed to determine the specific cause of individual score changes.

***Table S9:*** *Proportion of mitotic scores that were lower, unchanged, or higher in Stage 2 compared with Stage 1, by tumour group.*

| Tumour group | Lower (%) | Same (%) | Higher (%) |
|---|---|---|---|
| Breast resection | 7.8 | 49.6 | 42.6 |
| Breast biopsy | 10.1 | 71.0 | 18.8 |
| Thyroid | 7.8 | 69.5 | 22.7 |
| GI NET | 4.1 | 54.5 | 41.5 |
| Meningioma | 3.5 | 84.4 | 12.1 |
| Leiomyoma | 5.2 | 59.3 | 35.6 |
| Sarcoma | 3.2 | 64.7 | 32.1 |
| **Combined** | **5.9** | **64.9** | **29.1** |

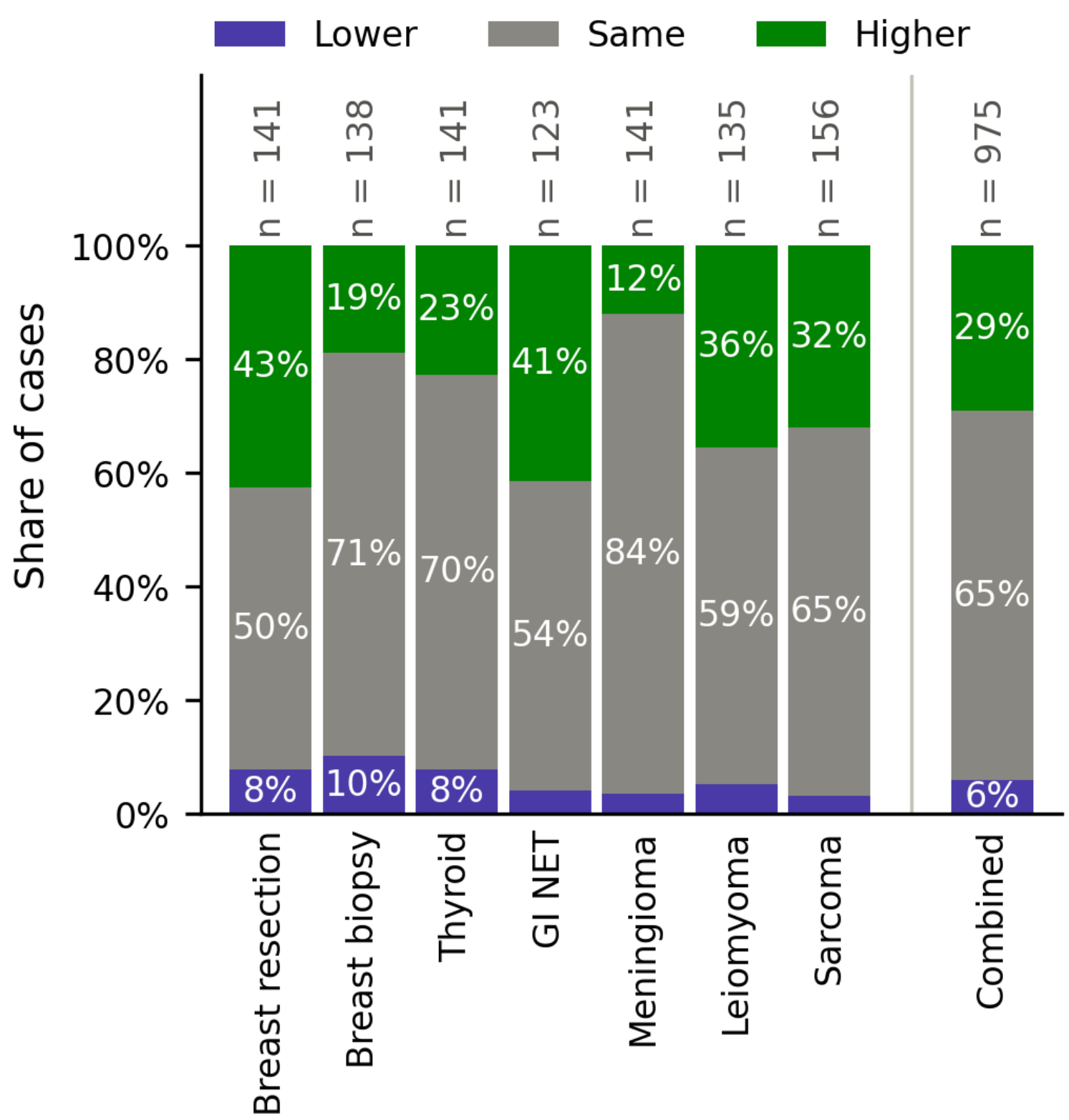

***Figure S9:*** *Distribution of mitotic score changes between Stage 1 and Stage 2 by tumour group and overall. Stacked bars show the proportion of assessments with lower, unchanged, or higher mitotic scores in Stage 2 relative to Stage 1.*

Mitotic score changes were also assessed by sex, age group and scanner to examine whether the pattern observed in the combined dataset was consistent across these categories. Results are shown in Table S10 and Figure S10.

***Table S10:*** *Proportion of mitotic scores that were lower, unchanged, or higher in Stage 2 compared with Stage 1, by sex, age group and scanner.*

| Sex | Lower (%) | Same (%) | Higher (%) |
|---|---|---|---|
| Female | 6.3 | 64.0 | 29.7 |
| Male | 4.5 | 68.7 | 26.8 |
| <40 | 1.6 | 67.4 | 31.0 |
| 40–69 | 6.9 | 64.1 | 29.0 |
| 70+ | 5.8 | 65.8 | 28.4 |
| Aperio AT2 | 5.3 | 63.8 | 30.9 |
| Hamamatsu NanoZoomer S360 | 6.6 | 65.5 | 27.9 |
| 3DHistech P1000 | 8.8 | 71.9 | 19.3 |

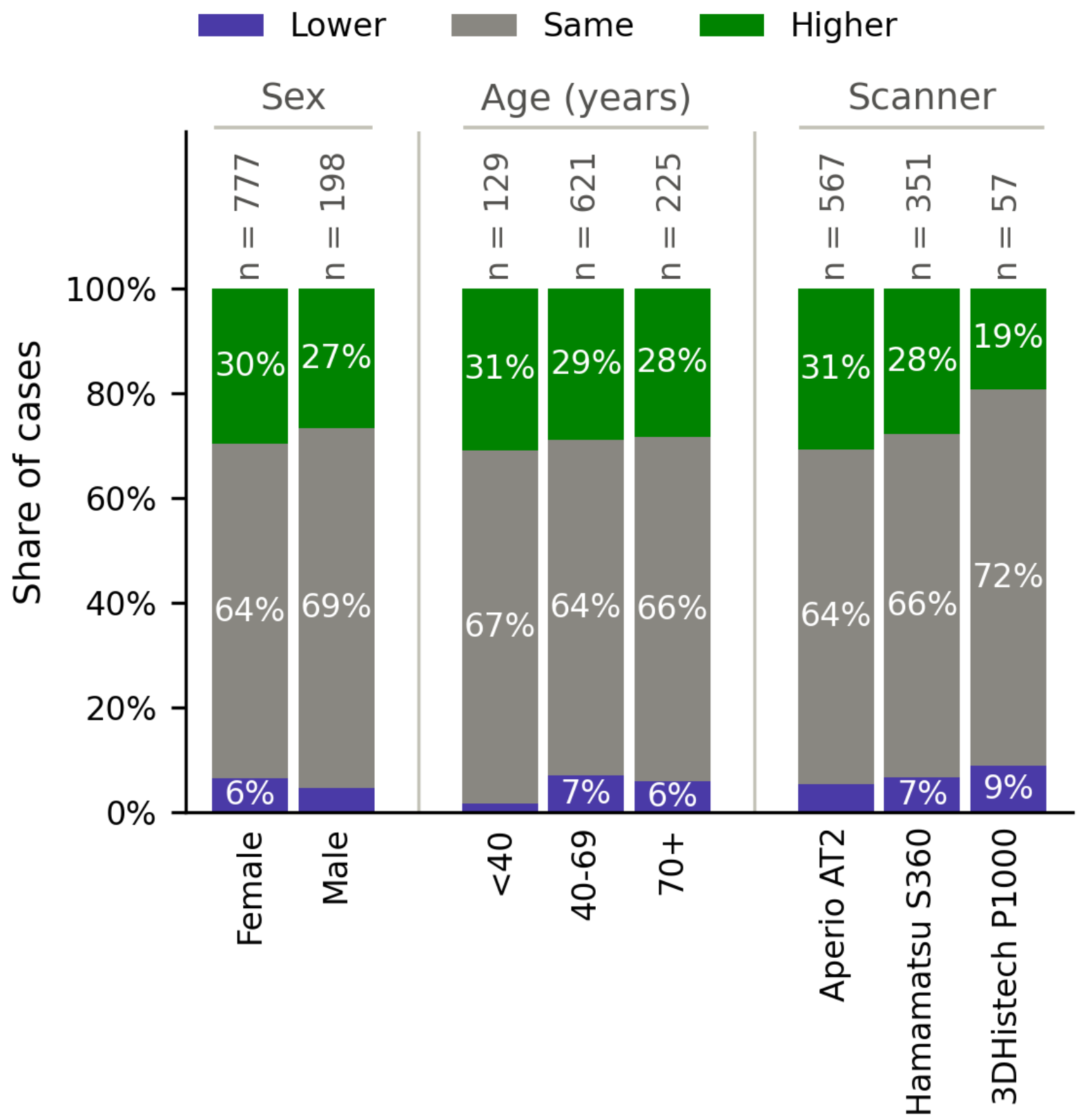

***Figure S10:*** *Distribution of mitotic score changes between Stage 1 and Stage 2 by sex, age group, and scanner type. Stacked bars show the proportion of assessments with lower, unchanged, or higher mitotic scores in Stage 2 relative to Stage 1.*

To further contextualise the changes in mitotic score between stages, we compared Stage 1 inter-pathologist agreement with score stability between Stage 1 and Stage 2. The results are shown below, with Panel A presenting the proportion of Stage 1 pairwise comparisons with the same score and Panel B presenting the proportion of cases in which the score remained unchanged between Stage 1 and Stage 2.

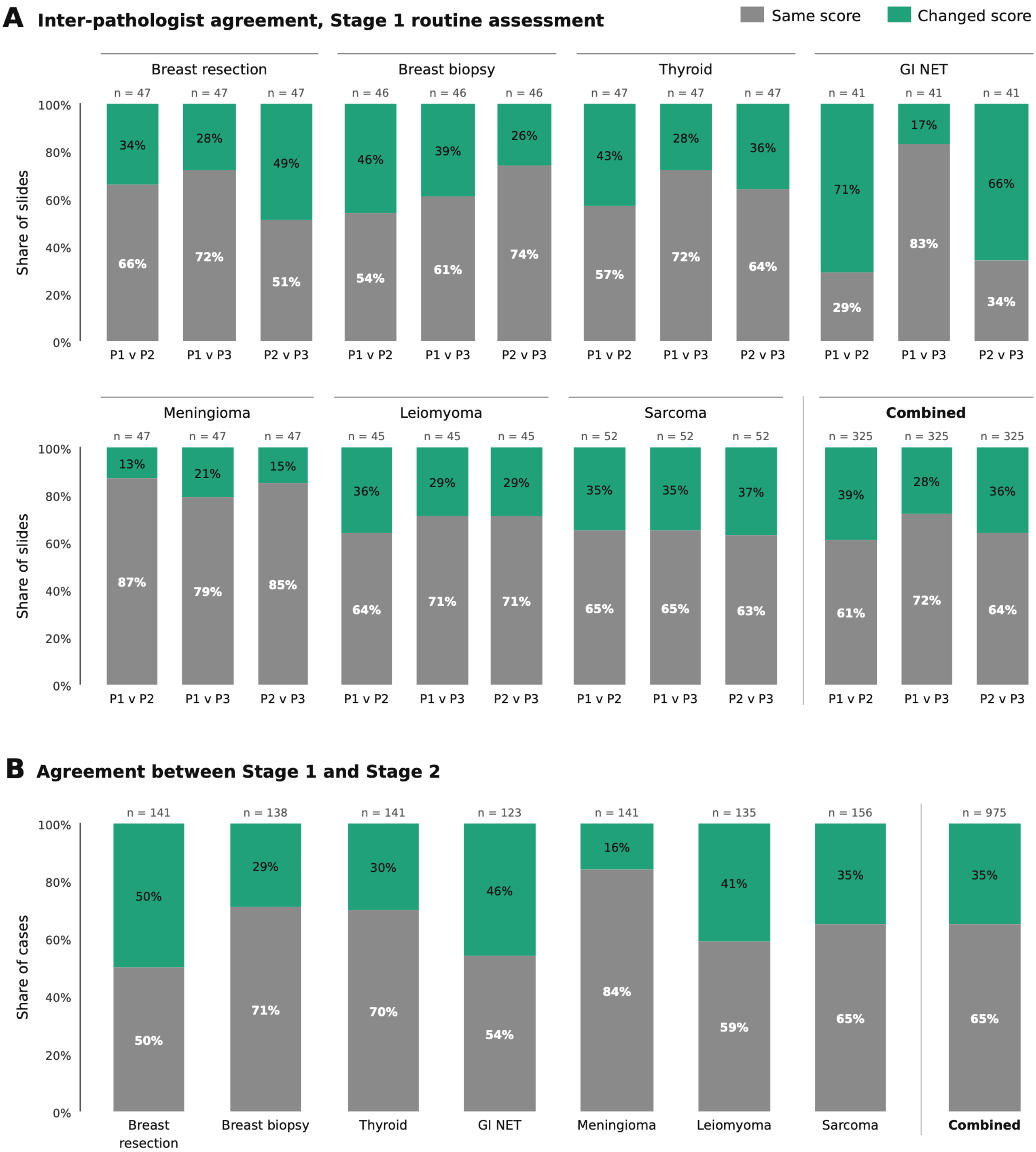

***Figure S11:*** *Percentage of slides and cases assigned the same mitotic score. A) Percentage of slides for which each pair of pathologists (P1-P3) assigned the same mitotic score during Stage 1 routine assessment, shown by tumour type. B) Percentage of cases for which the assigned mitotic score was the same between Stage 1 and Stage 2, shown by tumour type. Melanoma is excluded from both panels because a mitotic score is not assigned for this tumour type.*

# S2 Supporting Analysis

## S2.1 Performance across image management systems

### S2.1.1 Overview

MitPro was evaluated in a second image management system to assess whether the improvement observed with AI assistance was reproduced in a different supported digital pathology environment.

The analysis included 95 unique slides from the National Centre of Pathology in Vilnius and North Tees and Hartlepool NHS Foundation Trust. The same slides were assessed in both HALO AP and Sectra, with all tumour groups from the main study represented. Results were pooled across tumour groups because the objective was to assess performance within each image management system rather than compare tumour-specific outcomes.

Both systems met the minimum integration requirements for MitPro, including display of proposed hotspot regions, detected mitotic figures, mitotic count, hotspot area and mitotic density. The systems differed in aspects of the review workflow. In HALO AP, pathologists could add or remove detected mitotic figures within the selected hotspot, with the displayed count updating automatically. In Sectra, the AI-generated count remained static, so annotation tools were used to mark false-positive and missed mitotic figures for inclusion in the final verified count. These differences affected how the AI output was reviewed and corrected, rather than the underlying MitPro detections.

Each slide was reviewed by five pathologists within each system. The reader groups were not necessarily the same between HALO AP and Sectra. The results were therefore interpreted as separate within-system comparisons of Stage 1 and Stage 2, rather than as a direct comparison between the two image management systems.

### S2.1.2 Interobserver consistency and correlation

Inter-pathologist agreement improved with AI assistance in both systems. In HALO AP, ICC increased from 0.672 in Stage 1 to 0.965 in Stage 2, a difference of 0.294. In Sectra, ICC increased from 0.608 to 0.962, a difference of 0.354. The confidence intervals for the between-stage differences were above zero in both systems.

Pairwise Pearson correlation showed the same pattern. Pooled correlation increased from 0.803 to 0.982 in HALO AP and from 0.745 to 0.978 in Sectra, indicating greater consistency between individual pathologists during AI-assisted assessment.

These findings show that improvements in pathologist agreement and correlation were reproduced across both image management systems. As the reader groups differed between systems, the results should be interpreted as within-system comparisons rather than evidence of a difference in performance between HALO AP and Sectra.

Numerical results are presented in Table S11 and Table S12, with the corresponding visual results shown in Figure S12.

***Table S11:*** *Interobserver agreement in Stage 1 and Stage 2 by image management system, with between-stage differences (ΔICC) and 95% confidence intervals.*

| Group | $ICC_{Stage\ 1}$ | $ICC_{Stage\ 2}$ | Δ ICC |
|---|---|---|---|
| HALO AP | 0.672 [0.555, 0.784] | 0.965 [0.950, 0.975] | 0.294 [0.184, 0.412] |
| Sectra | 0.608 [0.543, 0.667] | 0.962 [0.946, 0.979] | 0.354 [0.295, 0.437] |

***Table S12:*** *Pooled pairwise Pearson correlations for Stage 1 and Stage 2 by image management system, with between-stage differences (Δr).*

| Group | $r_{Stage\ 1}$ | $r_{Stage\ 2}$ | Δ r |
|---|---|---|---|
| HALO AP | 0.803 [0.667, 0.885] | 0.982 [0.953, 0.991] | 0.178 |
| Sectra | 0.745 [0.612, 0.833] | 0.978 [0.959, 0.988] | 0.234 |

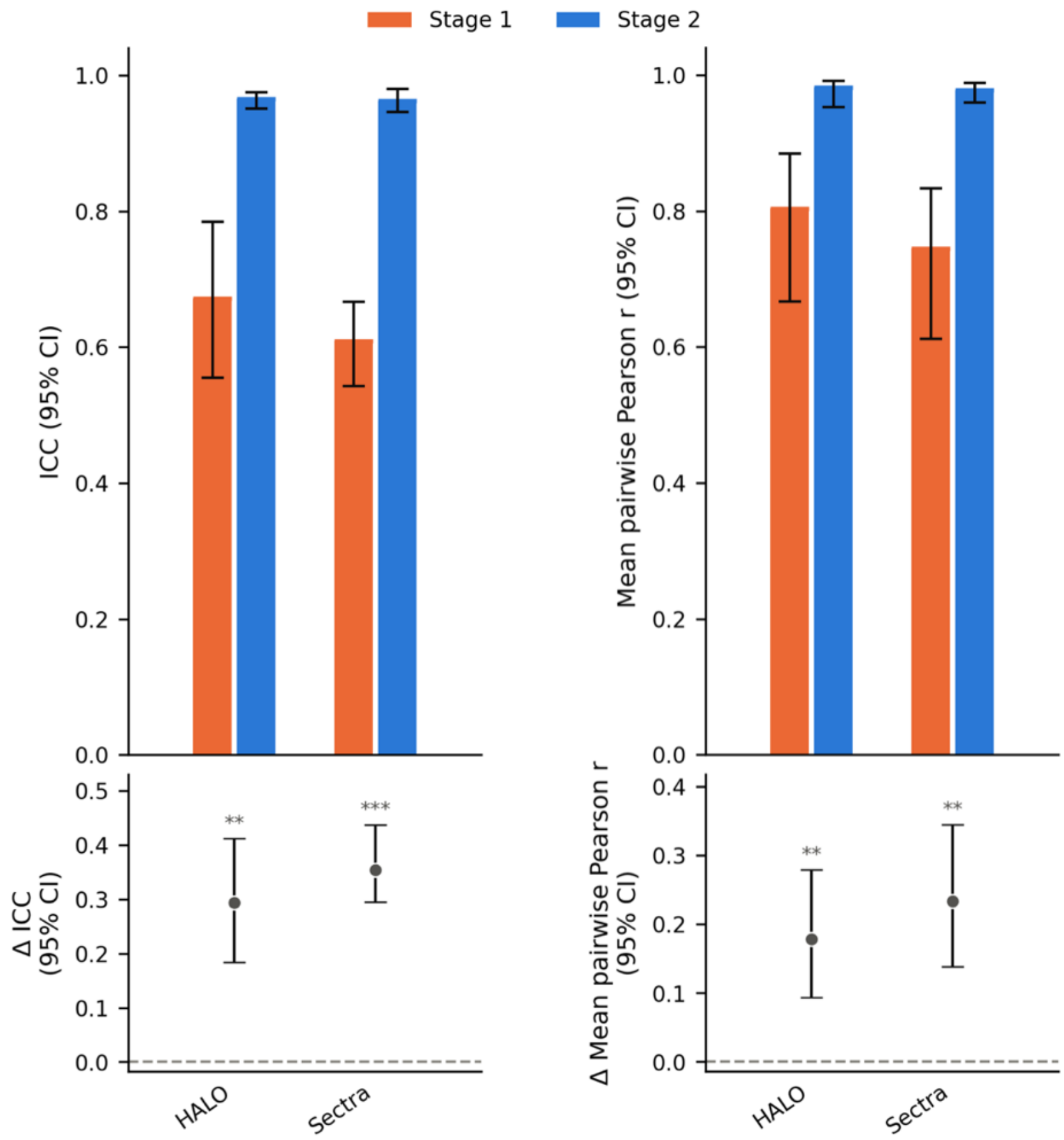

***Figure S12:*** *Interobserver agreement and correlation by image management system. The left plots show inter-pathologist agreement (ICC) for Stage 1 and Stage 2 for HALO AP and Sectra, while the right plots show the corresponding pooled pairwise Pearson correlation coefficients. For each metric, the upper panel shows the Stage 1 and Stage 2 estimates with 95% confidence intervals, and the lower panel shows the corresponding between-stage difference (ΔICC = Stage 2 − Stage 1 and Δr = Stage 2 - Stage 1, respectively) with 95% confidence intervals. Statistical significance of the difference between Stage 1 and Stage 2 is indicated by asterisks: p < 0.05 = *, p < 0.01 = **, p < 0.001 = ***, and p < 0.0001 = ****.*

### S2.1.3 Assessment time

Assessment time was lower with AI-assisted counting in both image management systems. Numerical results are provided in Table S13, with the corresponding visual results shown in Figure S13.

In HALO AP, assessment time decreased from 337.1 seconds in Stage 1 to 144.1 seconds in Stage 2, giving an estimated saving of 184.7 seconds, or approximately 3.1 minutes per assessment. In Sectra, assessment time decreased from 412.4 seconds to 185.4 seconds, giving an estimated saving of 215.1 seconds, or approximately 3.6 minutes per assessment. The confidence intervals for time saved were above zero in both systems.

Despite the difference in the review workflow between the systems, a substantial reduction in assessment time was observed in both systems.

These findings show that the efficiency benefit of MitPro was maintained across two different image management system workflows.

***Table S13:*** *Stage 1 and Stage 2 assessment times by image management system, with estimated time saved and 95% confidence intervals.*

| Group | Time $_{Stage\ 1}$ | Time $_{Stage\ 2}$ | Time Saved |
|---|---|---|---|
| HALO AP | 337.1 [320.9, 357.4] | 144.1 [134.3, 155.2] | 184.7 [168.8, 201.9] |
| Sectra | 412.4 [384.4, 441.6] | 185.4 [173.1, 201.5] | 215.1 [185.7, 247.6] |

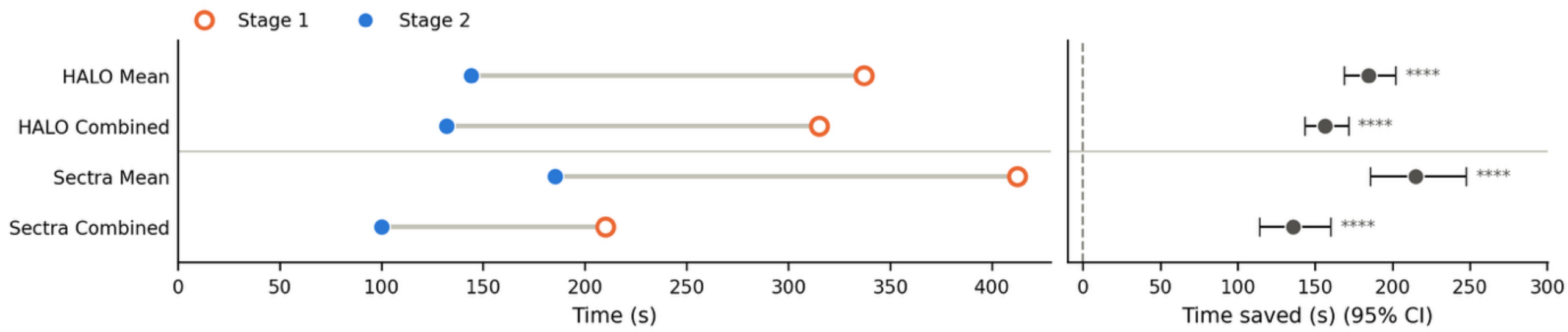

***Figure S13:*** *Assessment time by image management system. The dumbbell plot shows Stage 1 and Stage 2 assessment times for HALO AP and Sectra. The corresponding time saved (Stage 1 − Stage 2) is shown to the right, with whiskers representing 95% confidence intervals. Statistical significance of the between-stage difference is indicated by asterisks: $p < 0.05$ = *, $p < 0.01$ = **, $p < 0.001$ = ***, and $p < 0.0001$ = ****.*

## S2.2 Performance on additional tumour types

### S2.2.1 Overview

A supporting analysis was performed in tumour types outside the intended use of MitPro to assess whether the improvements associated with AI-assisted mitotic counting were also observed across different tumour morphologies.

The analysis included 76 slides, comprising 38 glioblastoma cases and 38 non-small cell lung cancer (NSCLC) cases from the National Centre of Pathology in Vilnius and North Tees and Hartlepool NHS Foundation Trust. Each slide represented a unique patient, and the NSCLC cohort included both lung adenocarcinoma and lung squamous cell carcinoma.

As mitotic counting is not widely used as a formal clinical assessment in these tumour types, no applicable clinical cut-offs were available for categorical mitotic scores. The analysis therefore focused on raw mitotic counts. These results provide supporting evidence of broader model generalisability, but do not support an extension of the intended use of MitPro.

### S2.2.2 Interobserver consistency and correlation

Across all 76 slides, ICC increased from 0.602 in Stage 1 to 0.987 in Stage 2. Similar improvements were observed in both centres and tumour types, with ICC increasing from 0.503 to 0.993 for glioblastoma and from 0.725 to 0.981 for NSCLC. The 95% confidence interval for the between-stage difference remained above zero in every analysis.

Pairwise correlation also increased, from 0.738 to 0.994 overall. Correlation increased from 0.777 to 0.998 for glioblastoma and from 0.735 to 0.992 for NSCLC, with the same pattern observed at both centres.

Results are shown below in Table S14, Table S15 and Figure S14.

***Table S14:*** *Interobserver agreement in Stage 1 and Stage 2 for glioblastoma and NSCLC, overall and by participating centre, with between-stage differences (ΔICC) and 95% confidence intervals.*

| Group | Slides | ICC $_{Stage\ 1}$ | ICC $_{Stage\ 2}$ | Δ ICC |
|---|---|---|---|---|
| Combined | 76 | 0.602 [0.401, 0.745] | 0.987 [0.977, 0.992] | 0.385 [0.243, 0.589] |
| Vilnius | 60 | 0.584 [0.374, 0.771] | 0.988 [0.977, 0.994] | 0.404 [0.218, 0.616] |
| North Tees | 16 | 0.655 [0.440, 0.775] | 0.977 [0.923, 0.992] | 0.321 [0.205, 0.516] |
| Glioblastoma | 38 | 0.503 [0.322, 0.716] | 0.993 [0.984, 0.996] | 0.489 [0.275, 0.671] |
| NSCLC | 38 | 0.725 [0.605, 0.846] | 0.981 [0.963, 0.992] | 0.256 [0.145, 0.371] |

***Table S15:*** *Pooled pairwise Pearson correlations for Stage 1 and Stage 2 for glioblastoma and NSCLC, overall and by participating centre, with between-stage differences (Δr) and 95% confidence intervals.*

| Group | Slides | r Stage 1 | r Stage 2 | Δ r |
|---|---|---|---|---|
| Combined | 76 | 0.738 [0.661, 0.849] | 0.994 [0.990, 0.997] | 0.257 [0.178, 0.368] |
| Vilnius | 60 | 0.761 [0.677, 0.864] | 0.995 [0.989, 0.997] | 0.234 [0.150, 0.337] |
| North Tees | 16 | 0.671 [0.495, 0.921] | 0.993 [0.987, 0.999] | 0.322 [0.161, 0.658] |
| Glioblastoma | 38 | 0.777 [0.713, 0.902] | 0.998 [0.995, 0.999] | 0.221 [0.097, 0.283] |
| NSCLC | 38 | 0.735 [0.548, 0.877] | 0.992 [0.984, 0.997] | 0.256 [0.125, 0.463] |

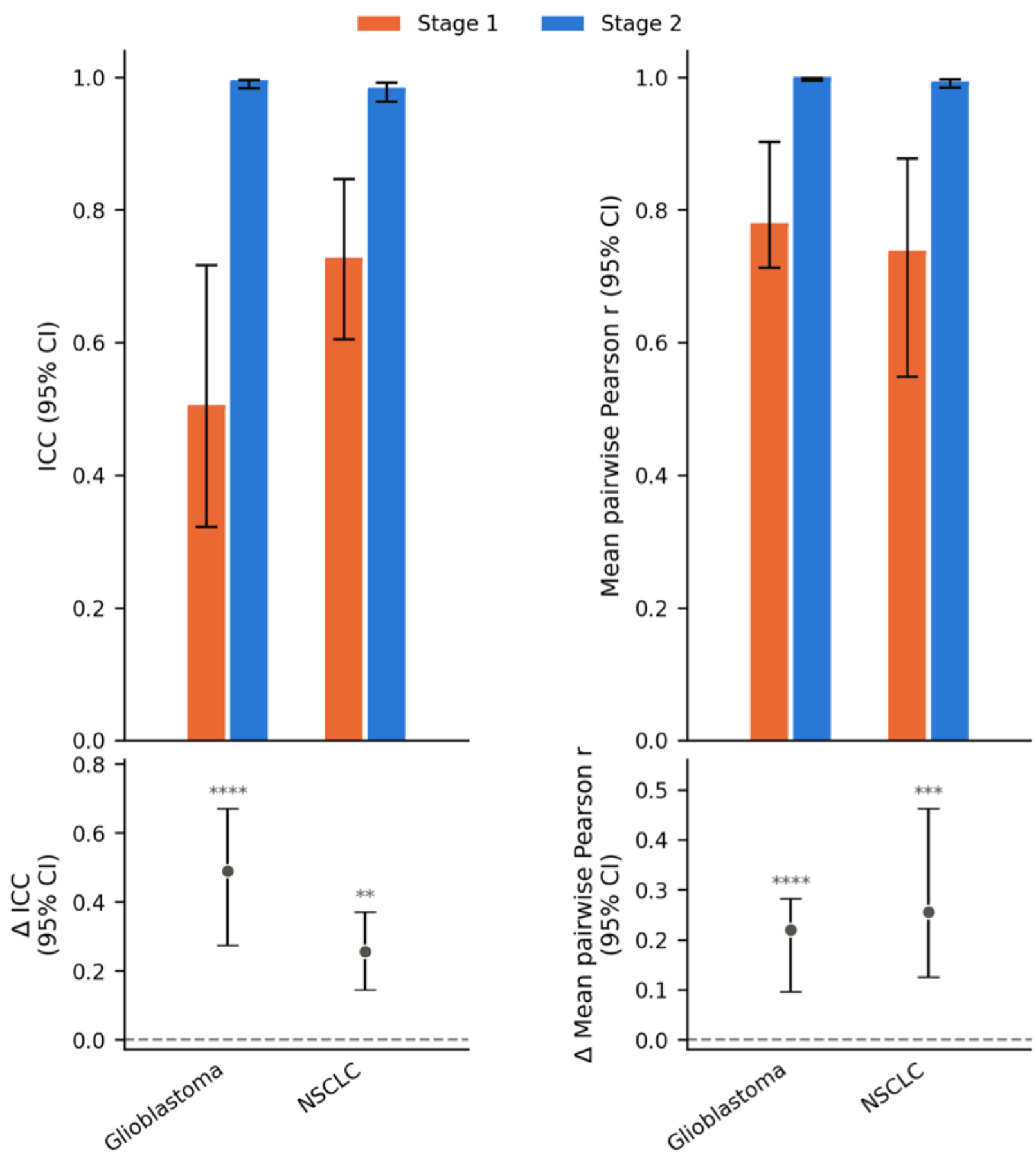

***Figure S14:*** *Interobserver agreement and correlation by tumour type. The left plots show inter-pathologist agreement (ICC) for Stage 1 and Stage 2 in glioblastoma and NSCLC, while the right plots show the corresponding pooled pairwise Pearson correlation coefficients. For each metric, the upper panel shows the Stage 1 and Stage 2 estimates with 95% confidence intervals, and the lower panel shows the corresponding between-stage difference (ΔICC = Stage 2 − Stage 1 and Δr = Stage 2 - Stage 1, respectively) with 95% confidence intervals. Statistical significance of the difference between Stage 1 and Stage 2 is indicated by asterisks: p < 0.05 = *, p < 0.01 = **, p < 0.001 = ***, and p < 0.0001 = ****.*

### S2.2.3 Assessment time

Assessment time was lower during Stage 2 for both tumour types and at both participating centres, as shown in Table S16.

For glioblastoma, assessment time decreased from 363.6 seconds in Stage 1 to 132.6 seconds in Stage 2. The estimated time saved was 221.1 seconds, corresponding to approximately 3.7 minutes per assessment. For NSCLC, assessment time decreased from 419.4 seconds to 151.4 seconds, with an estimated saving of 265.4 seconds, or approximately 4.4 minutes per assessment.

The same pattern was observed when the data were considered by participating centre. In Vilnius, assessment time decreased from 357.5 seconds to 129.9 seconds, with a time saving of 238.8 seconds. At North Tees, assessment time decreased from 412.1 seconds to 153.7 seconds, corresponding to a saving of 252.0 seconds.

In all analyses, the 95% confidence interval for time saved remained entirely above zero, supporting a consistent reduction in assessment time with AI-assisted counting for both tumour types, despite being outside the intended use of MitPro.

Results are shown in Table S16 and Figure S15.

***Table S16:*** *Stage 1 and Stage 2 assessment times for glioblastoma and NSCLC and by participating centre, with estimated time saved and 95% confidence intervals.*

| Group | Slides | Time $_{Stage\ 1}$ | Time $_{Stage\ 2}$ | Time Saved |
|---|---|---|---|---|
| Vilnius | 60 | 357.5 [334.4, 388.8] | 129.9 [99.2, 144.1] | 238.8 [206.8, 256.4] |
| North Tees | 16 | 412.1 [374.8, 475.4] | 153.7 [134.5, 188.3] | 252.0 [205.8, 305.8] |
| Glioblastoma | 38 | 363.6 [332.2, 407.2] | 132.6 [108.4, 151.2] | 221.1 [197.1, 267.1] |
| NSCLC | 38 | 419.4 [366.9, 468.3] | 151.4 [128.6, 183.8] | 265.4 [219.8, 292.6] |

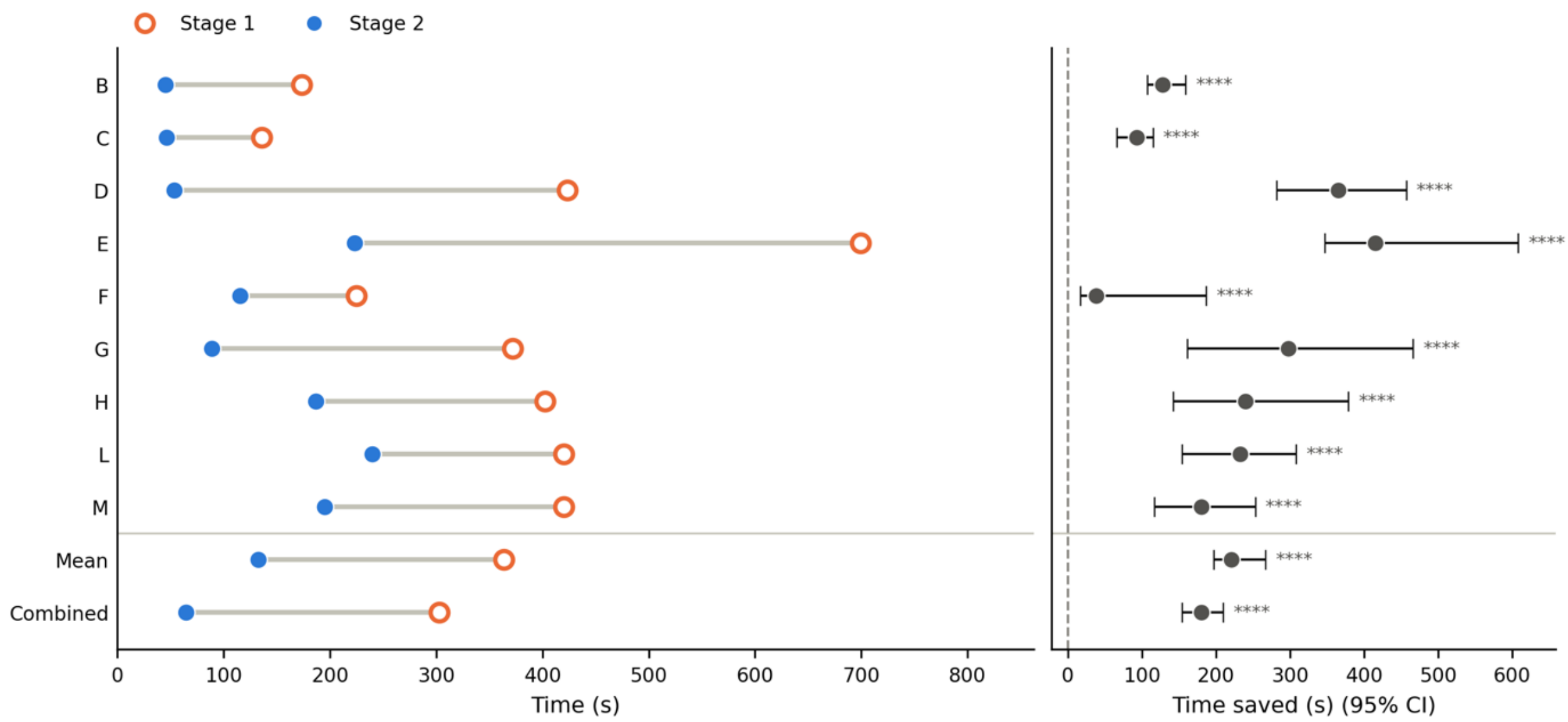

***Figure S15:*** *Assessment time by pathologist for glioblastoma. The dumbbell plot shows Stage 1 and Stage 2 assessment times for each pathologist. The corresponding time saved (Stage 1 − Stage 2) is shown to the right, with whiskers representing 95% confidence intervals. Statistical significance of the between-stage difference is indicated by asterisks: p < 0.05 = *, p < 0.01 = **, p < 0.001 = ***, and p < 0.0001 = ****.*

# S3 Mitotic figure detection performance

## S3.1 Dataset

The analytical validation dataset comprised 215 whole-slide images (WSIs) across seven tumour types: melanoma, thyroid, gastrointestinal neuroendocrine tumour, meningioma, leiomyoma, sarcoma and breast cancer. The dataset included slides from three clinical centres and scanners. Detailed distributions by tumour type, subtype, centre, sex, age group and scanner are provided in Table S17 - Table S20.

Ground truth annotation was performed on predefined 1 mm² regions selected within the WSIs. These regions were preselected by pathologists to capture variation in morphology and mitotic activity across the supported tumour types. The same region size was used throughout the analytical validation to provide a standardised area for annotation and subsequent performance assessment. In total, 463 regions were selected from the 215 slides.

All annotation was performed within HALO AP. Each selected region was reviewed independently by multiple pathologists, who were blinded to the annotations made by the other readers. Pathologists marked individual mitotic figures within the predefined 1 mm² regions without access to the annotations of the other participants. This produced 11,001 individual pathologist annotations across the dataset.

Following independent annotation, annotations were combined where multiple pathologists identified the same mitotic figure. The spatial locations of the corresponding annotations were averaged to define a single consensus position. This resulted in 1,827 consensus mitotic figure annotations, which formed the ground truth for the analytical performance assessment.

The number of selected regions and consensus annotations varied across tumour groups, reflecting differences in the amount of tissue sampled and the mitotic activity present within the selected regions. The dataset also included a range of histological subtypes within several tumour groups, particularly melanoma, thyroid, meningioma, sarcoma and breast cancer.

Not every selected region contained mitotic figures. The dataset intentionally included regions with little or no mitotic figures as well as non-tumour tissue, such as stroma and adipose tissue, and areas containing common histological or imaging artefacts. These regions are relevant to analytical validation because MitPro is applied to WSIs containing a wide range of tissue appearances, many of which do not contain mitotic figures. Their inclusion therefore allowed performance to be assessed in both mitotically active and inactive regions, including the potential for false-positive detections.

***Table S17:*** *Summary of the number of annotations collected for each tumour type.*

| Tissue Group | Slides | Regions | Total Annotations | Final Annotations |
|---|---|---|---|---|
| **All** | 215 | 463 | 11,001 | 1,827 |
| Melanoma | 37 | 89 | 2,275 | 444 |
| Thyroid | 33 | 66 | 696 | 119 |
| GI NET | 16 | 25 | 661 | 77 |
| Meningioma | 31 | 69 | 998 | 200 |
| Leiomyoma | 30 | 57 | 633 | 115 |
| Sarcoma | 35 | 99 | 3,760 | 688 |
| Breast | 33 | 58 | 1,978 | 184 |

***Table S18:*** *Summary of the number of annotations collected for each cancer subtype.*

| Tissue Group | Subtype | Slides | Regions | Total Annotations | Final Annotations |
|---|---|---|---|---|---|
| Melanoma | Malignant melanoma, not specified | 11 | 16 | 738 | 105 |
| | Superficial spreading melanoma | 9 | 25 | 557 | 160 |
| | Nodular melanoma | 5 | 12 | 454 | 64 |
| | Acral lentiginous melanoma | 4 | 17 | 173 | 40 |
| | Dysplastic melanocytic naevus | 4 | 7 | 36 | 7 |
| | Lentigo maligna melanoma | 4 | 12 | 317 | 68 |
| Thyroid | Thyroid Papillary | 19 | 32 | 429 | 71 |
| | Thyroid Follicular | 5 | 15 | 58 | 15 |
| | Thyroid Medullary | 4 | 7 | 79 | 9 |
| | Thyroid Oncocytic | 3 | 6 | 111 | 19 |
| | Thyroid Follicular Adenoma | 2 | 6 | 19 | 5 |
| GI NET | GI NET | 16 | 25 | 661 | 77 |
| Meningioma | Meningothelial meningioma | 16 | 25 | 274 | 49 |
| | Atypical meningioma | 6 | 18 | 246 | 43 |
| | Transitional meningioma | 5 | 15 | 172 | 18 |
| | Fibrous meningioma | 3 | 6 | 60 | 5 |
| | Anaplastic meningioma | 1 | 5 | 246 | 85 |
| Leiomyoma | Leiomyoma | 30 | 57 | 633 | 115 |
| Sarcoma | Fibrosarcoma and fibroblastic/myofibroblastic tumours | 5 | 5 | 249 | 8 |
| | Leiomyosarcoma | 5 | 15 | 691 | 205 |
| | Ewing sarcoma family tumours | 4 | 22 | 727 | 138 |
| | Liposarcoma | 4 | 17 | 189 | 44 |
| | Malignant peripheral nerve sheath tumour | 4 | 13 | 560 | 129 |
| | Other rare sarcoma subtypes | 4 | 6 | 78 | 12 |
| | Synovial sarcoma | 3 | 12 | 131 | 24 |
| | Clear cell sarcoma | 2 | 3 | 187 | 17 |
| | Epithelioid sarcoma | 2 | 4 | 367 | 57 |
| | Rhabdomyosarcoma | 2 | 2 | 581 | 54 |
| Breast | Infiltrating duct carcinoma | 25 | 36 | 1702 | 141 |
| | Lobular carcinoma | 8 | 22 | 276 | 43 |

***Table S19:*** *Summary of the number of annotations collected for each centre.*

| Tissue Group | Slides | Regions | Total Annotations | Final Annotations |
|---|---|---|---|---|
| Vilnius | 123 | 223 | 6,559 | 723 |
| North Tees | 53 | 107 | 2,399 | 488 |
| MD Anderson Madrid | 39 | 133 | 2,043 | 616 |

***Table S20:*** *Summary of the number of annotations collected across sex, age group and scanners.*

| Tissue Group | Slides | Regions | Total Annotations | Final Annotations |
|---|---|---|---|---|
| Male | 55 | 128 | 4,120 | 790 |
| Female | 160 | 335 | 6,881 | 1,037 |
| <40 | 25 | 46 | 1,199 | 156 |
| 40-69 | 135 | 296 | 6,630 | 1,155 |
| 70+ | 55 | 121 | 3,172 | 516 |
| Aperio AT2 | 123 | 223 | 6,559 | 723 |
| Hamamatsu NanoZoomer S360 | 80 | 217 | 4,004 | 1,023 |
| 3DHistech P1000 | 12 | 23 | 438 | 81 |

## S3.2 Performance across tumour types

MitPro was evaluated across the complete analytical validation dataset by comparing detection results with the consensus pathologist annotations. Across all tumour types and centres combined, the model achieved a precision of 0.728, recall of 0.894 and $F_1$ score of 0.803. Results are provided in Table S21, with a visual summary shown in Figure S16.

Recall was higher than precision, indicating that the model more often produced additional candidate detections than missed mitotic figures present in the consensus ground truth.

Performance was also assessed separately across the seven tumour types. $F_1$ scores ranged from 0.704 for leiomyoma to 0.864 for meningioma. Variation between tumour groups was driven mainly by differences in precision and recall, with leiomyoma showing the lowest $F_1$ score because of lower precision while recall remained high.

***Table S21:*** *Summary of MitPro analytical performance across the complete validation dataset, participating clinical centres and main tumour groups. Precision, recall and F1 score are reported with 95% confidence intervals.*

| Subgroup | TP | FP | FN | Precision | Recall | $F_1$ Score |
|---|---|---|---|---|---|---|
| Combined | 1,634 | 611 | 193 | 0.728 [0.687, 0.763] | 0.894 [0.871, 0.915] | 0.803 [0.776, 0.827] |
| Melanoma | 412 | 158 | 32 | 0.723 [0.648, 0.798] | 0.928 [0.871, 0.963] | 0.813 [0.770, 0.864] |
| Thyroid | 103 | 22 | 16 | 0.824 [0.679, 0.918] | 0.866 [0.716, 0.929] | 0.844 [0.761, 0.904] |
| GI NET | 62 | 16 | 15 | 0.795 [0.700, 0.874] | 0.805 [0.661, 0.923] | 0.800 [0.694, 0.876] |
| Meningioma | 191 | 51 | 9 | 0.789 [0.624, 0.860] | 0.955 [0.929, 0.980] | 0.864 [0.756, 0.904] |
| Leiomyoma | 100 | 69 | 15 | 0.592 [0.477, 0.678] | 0.870 [0.788, 0.922] | 0.704 [0.611, 0.767] |
| Sarcoma | 620 | 242 | 68 | 0.719 [0.627, 0.776] | 0.901 [0.869, 0.932] | 0.800 [0.742, 0.837] |
| Breast | 146 | 53 | 38 | 0.734 [0.659, 0.805] | 0.793 [0.712, 0.875] | 0.762 [0.707, 0.823] |

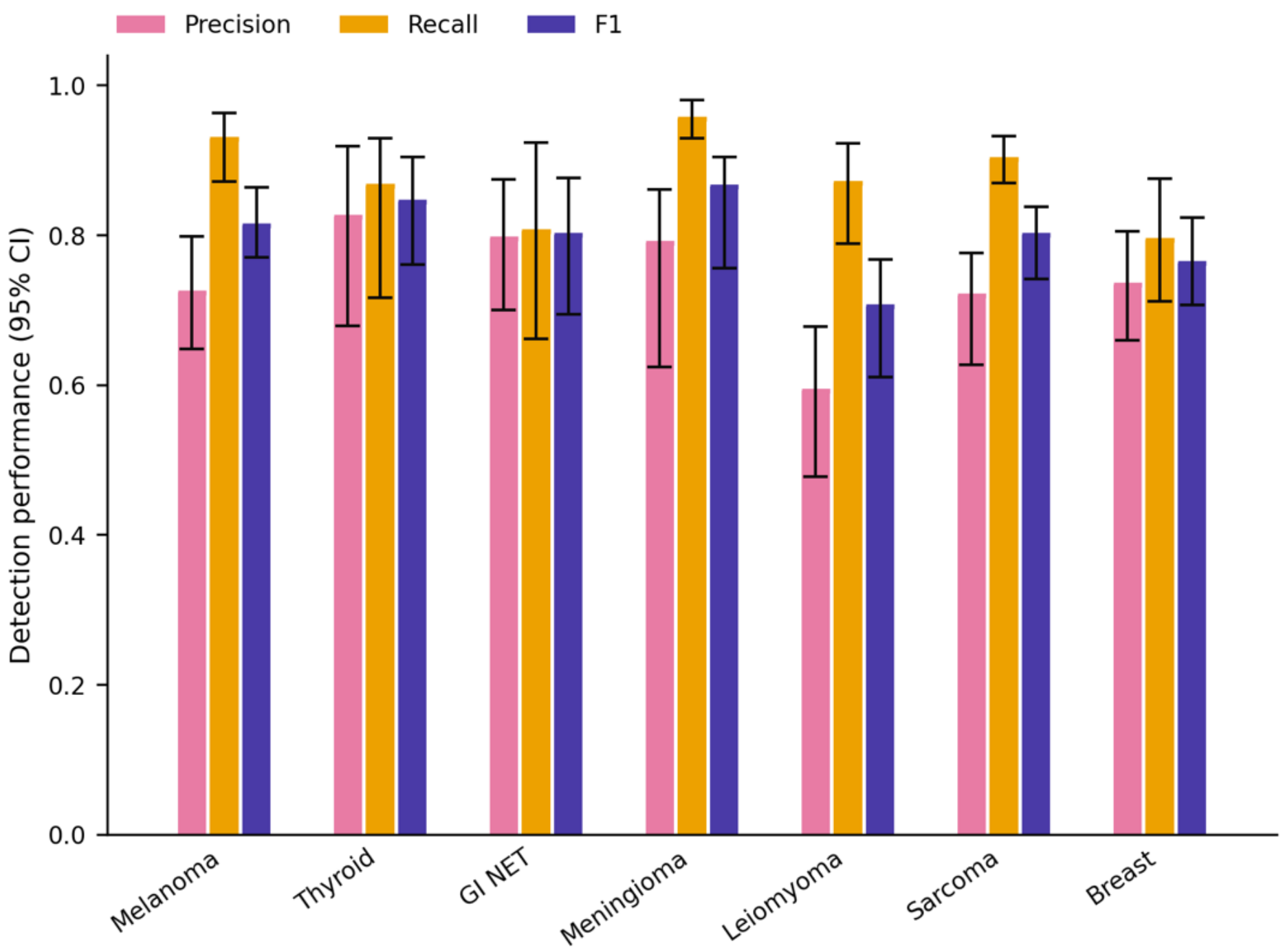

***Figure S16:*** *Precision, recall and $F_1$ score for MitPro across the main tumour groups in the analytical validation dataset.*

## S3.3 Performance across subtypes

Analytical performance was also assessed across the histological subtypes represented within the main tumour groups. A visual summary of precision, recall and $F_1$ score by subtype is shown in Figure S17. Some variation between subtypes was observed, reflecting differences in morphology, mitotic activity and the amount of data available within each subgroup.

Breast cancer included infiltrating duct and lobular carcinoma, with biopsy and resection specimens combined for this analytical assessment. Lobular carcinoma showed higher observed performance than infiltrating duct carcinoma, with an $F_1$ score of 0.884 compared with 0.722, although the lobular subgroup included fewer slides. Performance across melanoma, thyroid and meningioma subtypes was generally strong. Most melanoma subtypes had $F_1$ scores between 0.77 and 0.88, while thyroid and meningioma subtypes were also generally high, with several smaller groups showing very high observed values. These estimates should be interpreted in the context of the limited number of slides and detections in some subgroups.

Recall was usually higher than precision across the subtype analysis, indicating that additional candidate detections were more common than missed mitotic figures. This pattern was particularly evident in several melanoma, meningioma and sarcoma subtypes.

Sarcoma showed the greatest variation across subtypes. Strong performance was observed for malignant peripheral nerve sheath tumour, clear cell sarcoma, leiomyosarcoma and several other groups, with $F_1$ scores around 0.81 to 0.87. Lower values were observed for fibrosarcoma and fibroblastic or myofibroblastic tumours and for the grouped category of other rare sarcomas, with $F_1$ scores of 0.533 and 0.516, respectively. These lower estimates were based on small numbers of slides and detections. GI NET and leiomyoma were each represented as a single subtype and therefore reflected the corresponding tumour-group results.

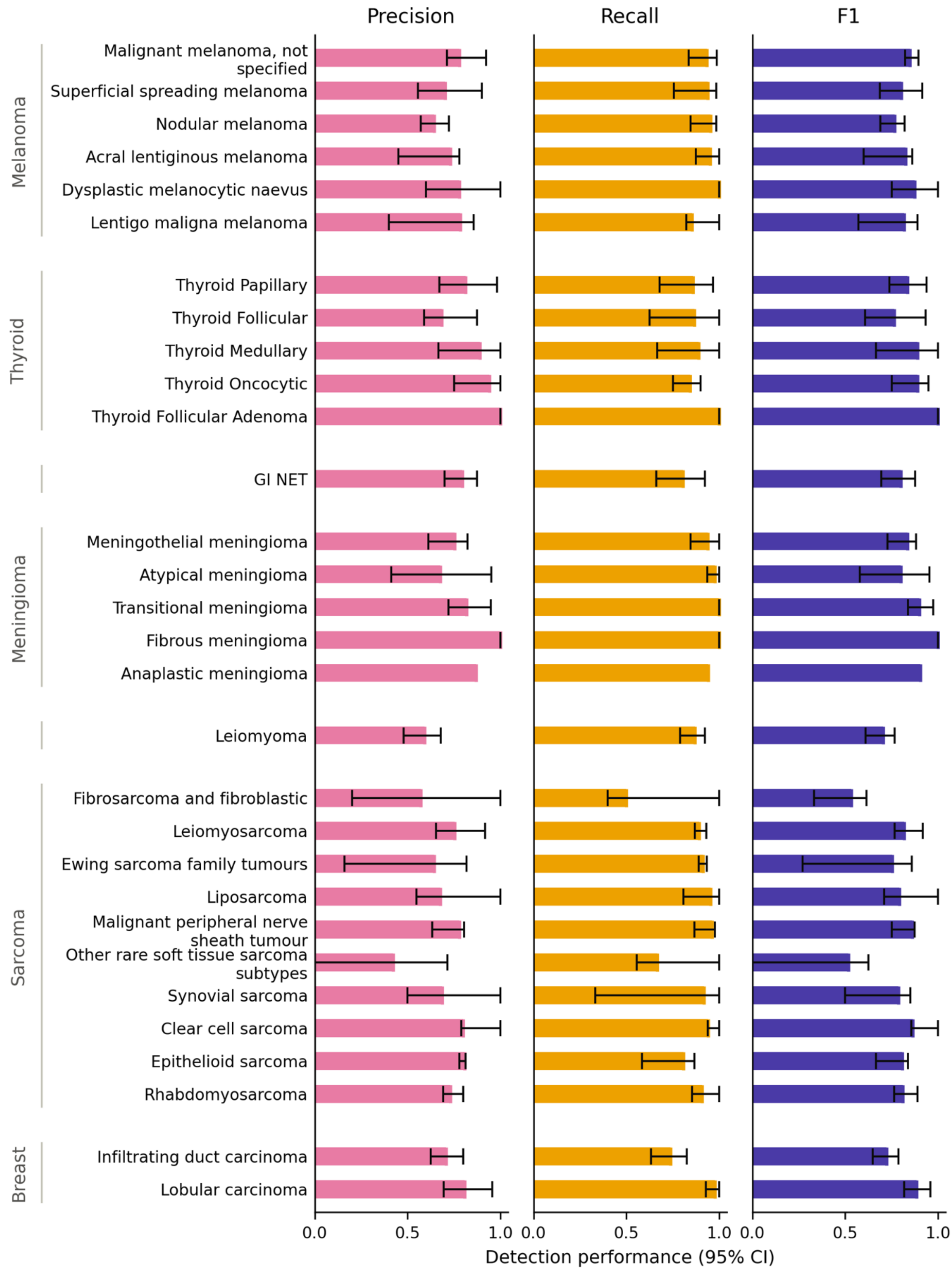

***Figure S17:*** *Precision, recall and $F_1$ score across the cancer subtypes represented in the analytical validation dataset. Error bars show 95% confidence intervals where available. Confidence intervals are not shown for subtypes represented by a single slide because bootstrap resampling was performed at the slide level.*

## S3.4 Performance across centre, sex, age and scanner

Analytical performance was also assessed across the three participating centres and by sex, age group and scanner to examine whether any notable variation was present across the conditions represented in the validation dataset. Results are provided in Tables 24 and 25, with corresponding visual summaries shown in Figure S18 and Figure S19.

Performance was generally similar across the three centres, with $F_1$ scores ranging from 0.773 to 0.849. The highest observed $F_1$ score was at MD Anderson Madrid, while Vilnius and North Tees showed similar results. Recall was higher than precision at all three centres, consistent with the pattern observed in the complete analytical validation dataset.

Performance was also similar between female and male patients, with $F_1$ scores of 0.786 and 0.825, respectively. Across age groups, $F_1$ scores ranged from 0.758 in patients under 40 years to 0.822 in patients aged 70 years or older. The under 40 group was the smallest age category and showed lower precision, while recall remained high.

Across scanners, $F_1$ scores ranged from 0.783 for the Aperio AT2 to 0.820 for the 3DHistech P1000, with the Hamamatsu NanoZoomer S360 at 0.816. Precision and recall were more closely balanced for the 3DHistech P1000, although this result was based on fewer slides. Scanner type was also linked to contributing centre, so differences between scanners may reflect other characteristics of the corresponding datasets and should not be interpreted as a direct comparison of scanner performance.

Performance therefore remained similar across centre, sex, age and scanner, with recall generally high and greater variation observed in precision. This supports the consistency of analytical performance across the demographic, clinical and technical conditions represented in the validation dataset.

***Table S22:*** *Summary of MitPro analytical performance across the three participating clinical centres. Precision, recall and $F_1$ score are reported with 95% confidence intervals.*

| Subgroup | TP | FP | FN | Precision | Recall | $F_1$ Score |
|---|---|---|---|---|---|---|
| Vilnius | 637 | 268 | 86 | 0.704 [0.622, 0.760] | 0.881 [0.841, 0.912] | 0.783 [0.729, 0.818] |
| North Tees | 412 | 166 | 76 | 0.713 [0.671, 0.757] | 0.844 [0.794, 0.884] | 0.773 [0.741, 0.804] |
| MD Anderson Madrid | 585 | 177 | 31 | 0.768 [0.689, 0.830] | 0.950 [0.905, 0.972] | 0.849 [0.803, 0.889] |

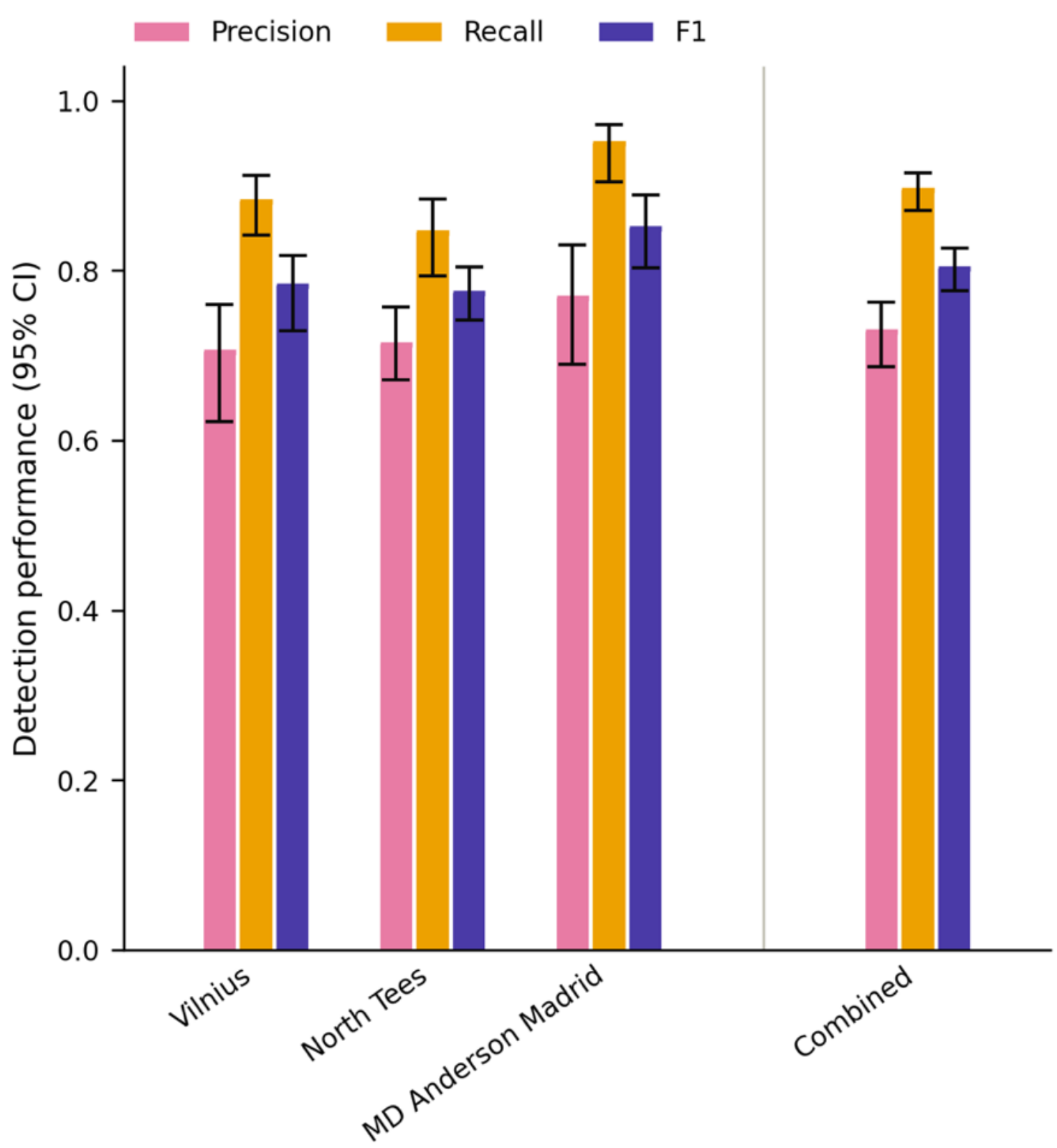

***Figure S18:*** *Precision, recall and $F_1$ score for MitPro across the three participating clinical centres and the combined analytical validation dataset. Error bars show 95% confidence intervals*

***Table S23:*** *Summary of MitPro analytical performance across sex, age group and scanner type. Precision, recall and $F_1$ score are reported with 95% confidence intervals.*

| Subgroup | TP | FP | FN | Precision | Recall | $F_1$ Score |
|---|---|---|---|---|---|---|
| Female | 917 | 379 | 120 | 0.708 [0.664, 0.746] | 0.884 [0.854, 0.908] | 0.786 [0.757, 0.812] |
| Male | 717 | 232 | 73 | 0.756 [0.670, 0.812] | 0.908 [0.866, 0.939] | 0.825 [0.772, 0.861] |
| <40 | 136 | 67 | 20 | 0.670 [0.403, 0.835] | 0.872 [0.786, 0.962] | 0.758 [0.554, 0.870] |
| 40-69 | 1,023 | 379 | 132 | 0.730 [0.688, 0.768] | 0.886 [0.855, 0.914] | 0.800 [0.774, 0.829] |
| 70+ | 475 | 165 | 41 | 0.742 [0.664, 0.809] | 0.921 [0.875, 0.947] | 0.822 [0.774, 0.862] |
| Aperio AT2 | 637 | 268 | 86 | 0.704 [0.622, 0.760] | 0.881 [0.841, 0.912] | 0.783 [0.729, 0.818] |
| Hamamatsu NanoZoomer S360 | 931 | 329 | 92 | 0.739 [0.691, 0.784] | 0.910 [0.878, 0.936] | 0.816 [0.784, 0.848] |
| 3DHistech P1000 | 66 | 14 | 15 | 0.825 [0.754, 0.894] | 0.815 [0.652, 0.925] | 0.820 [0.725, 0.875] |

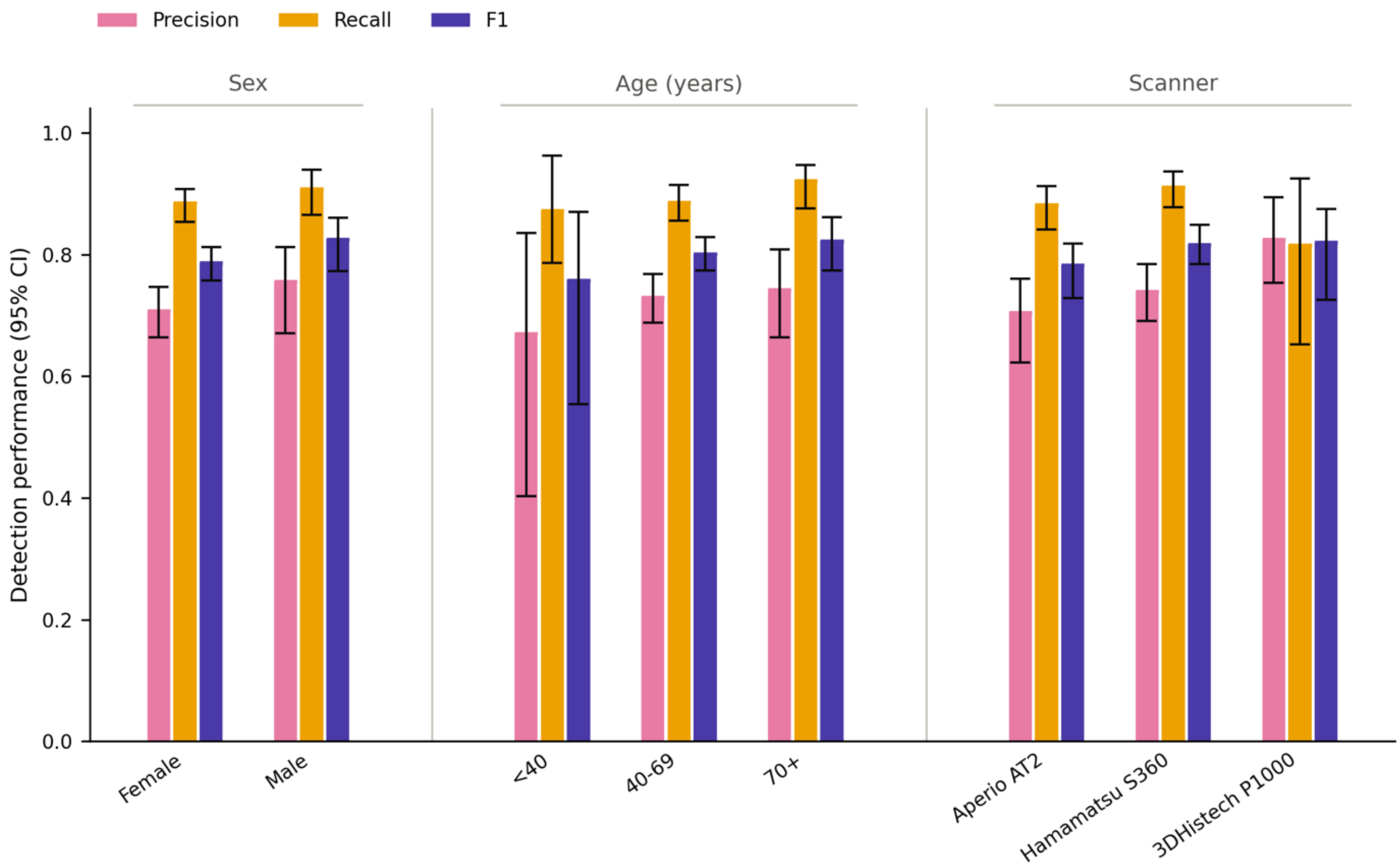

***Figure S19:*** *Precision, recall and $F_1$ score for MitPro across sex, age group and scanner type in the analytical validation dataset. Error bars show 95% confidence intervals.*

## S3.5 Generalisation to additional cancer types

To provide additional evidence of model generalisation, MitPro was also evaluated on glioblastoma and non-small cell lung cancer, two tumour types outside the intended use of the tool.

Across the combined dataset, MitPro achieved an $F_1$ score of 0.794, with similar performance for glioblastoma and NSCLC at 0.784 and 0.809, respectively. Results were also broadly similar across the two contributing centres, by sex and across the scanners represented. Some variation was observed by age, particularly in the under 40 group, which had fewer cases and wider confidence intervals.

***Table S24:*** *Summary of MitPro performance on the additional glioblastoma and NSCLC datasets used to assess generalisation and robustness beyond the intended-purpose tumour groups. Results are presented for the*

*combined dataset, contributing clinical centres, tumour type, sex, age group and scanner type. Precision, recall and $F_1$ score are reported with 95% confidence intervals.*

| Subgroup | TP | FP | FN | Precision | Recall | $F_1$ Score |
|---|---|---|---|---|---|---|
| Combined | 772 | 270 | 131 | 0.741 [0.629, 0.810] | 0.855 [0.793, 0.905] | 0.794 [0.736, 0.824] |
| Vilnius | 432 | 183 | 48 | 0.702 [0.600, 0.854] | 0.900 [0.824, 0.941] | 0.789 [0.727, 0.854] |
| North Tees | 340 | 87 | 83 | 0.796 [0.754, 0.861] | 0.804 [0.715, 0.875] | 0.800 [0.763, 0.831] |
| Glioblastoma | 467 | 191 | 66 | 0.710 [0.616, 0.826] | 0.876 [0.814, 0.928] | 0.784 [0.735, 0.840] |
| NSCLC | 305 | 79 | 65 | 0.794 [0.730, 0.897] | 0.824 [0.681, 0.909] | 0.809 [0.749, 0.847] |
| Female | 311 | 86 | 60 | 0.783 [0.747, 0.841] | 0.838 [0.741, 0.913] | 0.810 [0.770, 0.849] |
| Male | 461 | 184 | 71 | 0.715 [0.610, 0.867] | 0.867 [0.776, 0.923] | 0.783 [0.725, 0.840] |
| <40 | 124 | 93 | 10 | 0.571 [0.514, 0.906] | 0.925 [0.000, 0.979] | 0.707 [0.000, 0.853] |
| 40-69 | 431 | 112 | 93 | 0.794 [0.746, 0.848] | 0.823 [0.742, 0.874] | 0.808 [0.769, 0.843] |
| 70+ | 217 | 65 | 28 | 0.770 [0.715, 0.897] | 0.886 [0.762, 0.954] | 0.824 [0.806, 0.869] |
| Aperio AT2 | 432 | 183 | 48 | 0.702 [0.600, 0.854] | 0.900 [0.824, 0.941] | 0.789 [0.727, 0.854] |
| 3DHistech P1000 | 340 | 87 | 83 | 0.796 [0.754, 0.861] | 0.804 [0.715, 0.875] | 0.800 [0.763, 0.831] |

## S3.6 Robustness analysis

Additional robustness testing was performed to assess whether MitPro maintained mitotic figure detection performance under technical variation that may arise during WSI acquisition and processing. The combined validation dataset was evaluated following controlled changes in image sharpness, noise, compression and staining appearance.

Stain variation was assessed using four reference images from independent external centres with different staining appearances and scanner systems. The validation images were normalised to each reference using Macenko stain normalisation, producing alternative versions of the same images with different colour and staining characteristics while preserving the underlying tissue morphology. The four reference stain appearances are shown in Figure S20.

Figure S21 illustrates the range of image variation introduced during the robustness testing. The top row shows examples following stain normalisation to each of the four reference targets. The remaining rows show increasing levels of Gaussian blur, JPEG compression and Gaussian noise.

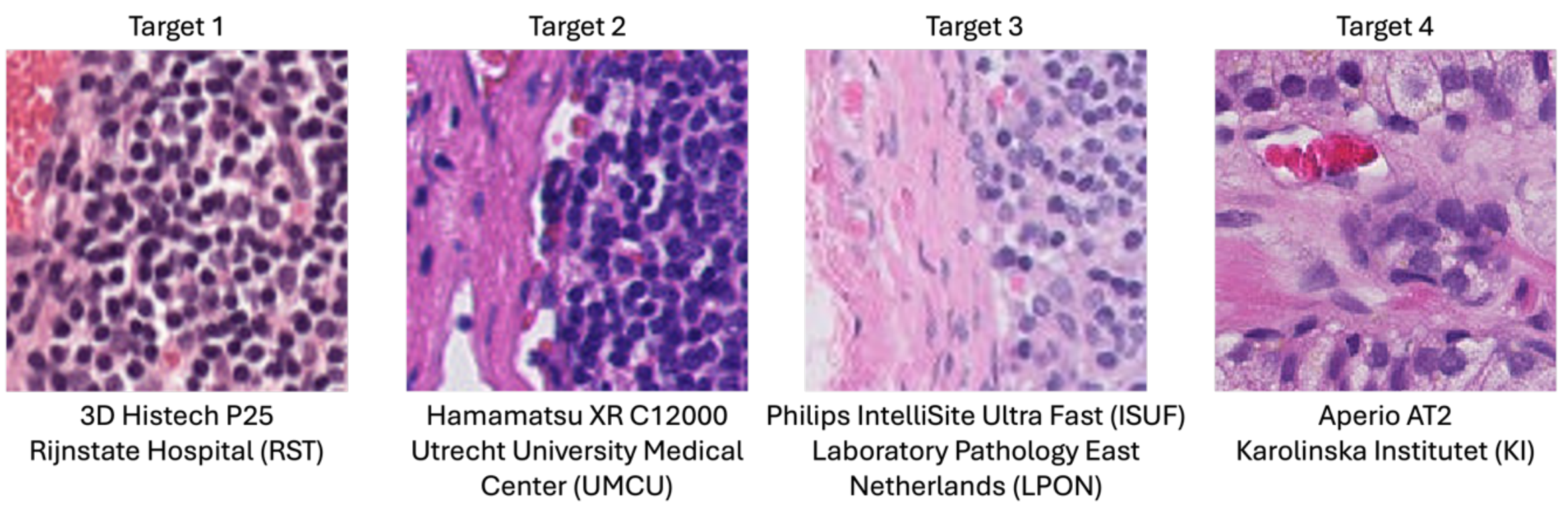

***Figure S20:*** *Target images used for stain normalisation. The combined validation set was normalised to make staining appearance consistent with the target images from these four centres, allowing evaluation of model performance under stain variation.*

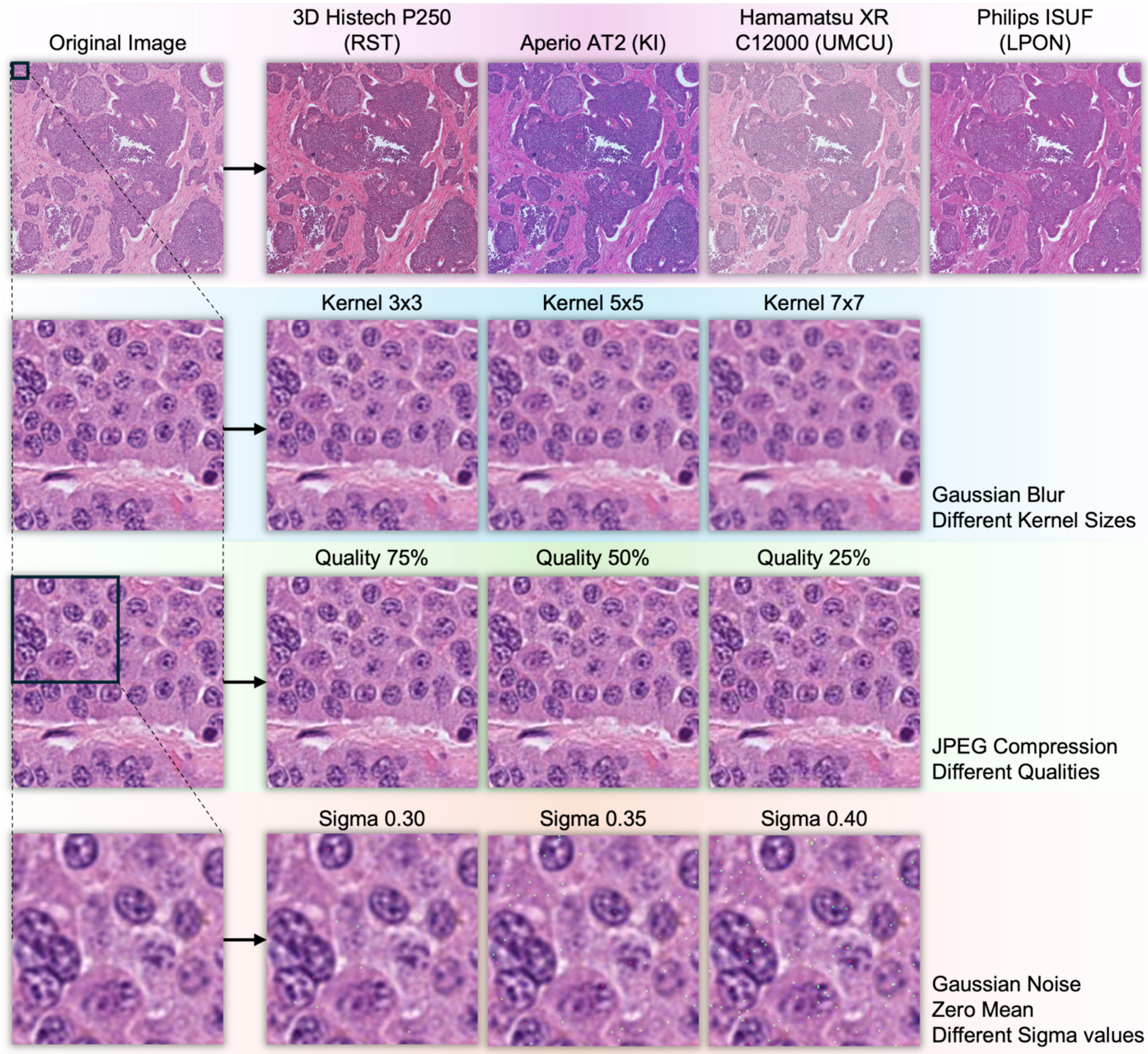

***Figure S21**: Appearance of images under input variation. The top row shows stain normalisation to each of the target images in **Figure S20**. The second to fourth rows show the input images with varying levels of blur, compression, and noise applied.*

Performance remained relatively stable across the evaluated levels of blur and noise. For Gaussian blur, $F_1$ decreased only slightly from 0.746 for the original images to 0.738 and 0.739 at blur kernel SDs of 3 and 5 pixels, respectively. Across the Gaussian noise conditions, $F_1$ remained between 0.736 and 0.745, compared with 0.746 for the original images. Numerical results are provided in Table S25 and Table S26.

JPEG compression also had a limited effect at moderate compression levels. $F_1$ was 0.738 at JPEG quality 75 and 0.736 at quality 50, compared with 0.746 for the original images. At the strongest compression level, JPEG quality 25, $F_1$ decreased to 0.721, driven mainly by a reduction in recall from 0.800 to 0.726. Results are provided in Table S27.

Following stain normalisation, $F_1$ ranged from 0.679 to 0.724 across the four target appearances, compared with 0.746 for the original images. The largest change was observed for Target 1, where recall decreased to 0.657 while precision remained similar to the original images. The other three stain targets produced $F_1$ scores between 0.722 and 0.724. Numerical results are provided in Table S28.

Figure S22 summarises precision, recall and $F_1$ score across all robustness experiments relative to the original images. Across the evaluated conditions, performance was generally maintained, with the largest changes observed under the strongest JPEG compression and for one of the stain-normalised target appearances.

***Table S25:*** *Detection performance across increasing Gaussian blur levels. Blur strength is reported as the blur kernel standard deviation (SD) in pixels. Precision, recall, and F1 score are reported with 95% confidence intervals in square brackets. TP, true positives; FP, false positives; FN, false negatives.*

| Blur kernel SD (px) | TP | FP | FN | Precision | Recall | $F_1$ Score |
|---|---|---|---|---|---|---|
| Original | 2185 | 945 | 545 | 0.698 [0.641, 0.730] | 0.800 [0.773, 0.828] | 0.746 [0.719, 0.766] |
| 3 | 2172 | 981 | 558 | 0.689 [0.629, 0.723] | 0.796 [0.769, 0.823] | 0.738 [0.710, 0.759] |
| 5 | 2199 | 1021 | 531 | 0.683 [0.631, 0.716] | 0.805 [0.780, 0.832] | 0.739 [0.712, 0.760] |

***Table S26:*** *Detection performance across increasing levels of Gaussian noise. Noise strength is reported as the noise standard deviation (SD). Precision, recall, and F1 score are reported with 95% confidence intervals in square brackets. TP, true positives; FP, false positives; FN, false negatives.*

| Noise SD | TP | FP | FN | Precision | Recall | $F_1$ Score |
|---|---|---|---|---|---|---|
| Original | 2185 | 945 | 545 | 0.698 [0.643, 0.730] | 0.800 [0.773, 0.827] | 0.746 [0.719, 0.767] |
| 0.2 | 2183 | 945 | 547 | 0.698 [0.641, 0.730] | 0.800 [0.772, 0.826] | 0.745 [0.719, 0.766] |
| 0.3 | 2170 | 937 | 560 | 0.698 [0.642, 0.730] | 0.795 [0.767, 0.822] | 0.744 [0.717, 0.764] |
| 0.4 | 2093 | 868 | 637 | 0.707 [0.653, 0.737] | 0.767 [0.739, 0.794] | 0.736 [0.709, 0.756] |

***Table S27:*** *Detection performance across increasing levels of JPEG compression. Compression level is reported using JPEG quality, with lower values indicating stronger compression. Precision, recall, and F1 score are reported with 95% confidence intervals in square brackets. TP, true positives; FP, false positives; FN, false negatives.*

| JPEG quality | TP | FP | FN | Precision | Recall | $F_1$ Score |
|---|---|---|---|---|---|---|
| Original | 2185 | 945 | 545 | 0.698 [0.642, 0.730] | 0.800 [0.773, 0.828] | 0.746 [0.719, 0.766] |
| 75 | 2205 | 1041 | 525 | 0.679 [0.629, 0.711] | 0.808 [0.783, 0.831] | 0.738 [0.709, 0.759] |
| 50 | 2177 | 1009 | 553 | 0.683 [0.631, 0.715] | 0.797 [0.770, 0.823] | 0.736 [0.706, 0.758] |
| 25 | 1981 | 787 | 749 | 0.716 [0.679, 0.745] | 0.726 [0.696, 0.754] | 0.721 [0.697, 0.742] |

***Table S28:*** *Detection performance following stain variation using four target stain appearances. Precision, recall, and F1 score are reported with 95% confidence intervals in square brackets. TP, true positives; FP, false positives; FN, false negatives.*

| Stain target | TP | FP | FN | Precision | Recall | $F_1$ Score |
|---|---|---|---|---|---|---|
| Original | 2185 | 945 | 545 | 0.698 [0.642, 0.730] | 0.800 [0.773, 0.828] | 0.746 [0.719, 0.766] |
| Target 1 | 1793 | 757 | 937 | 0.703 [0.646, 0.738] | 0.657 [0.617, 0.698] | 0.679 [0.656, 0.703] |
| Target 2 | 1981 | 778 | 749 | 0.718 [0.675, 0.750] | 0.726 [0.691, 0.760] | 0.722 [0.698, 0.747] |
| Target 3 | 1910 | 654 | 820 | 0.745 [0.698, 0.775] | 0.700 [0.655, 0.734] | 0.722 [0.695, 0.746] |
| Target 4 | 1954 | 715 | 776 | 0.732 [0.683, 0.764] | 0.716 [0.685, 0.746] | 0.724 [0.701, 0.747] |

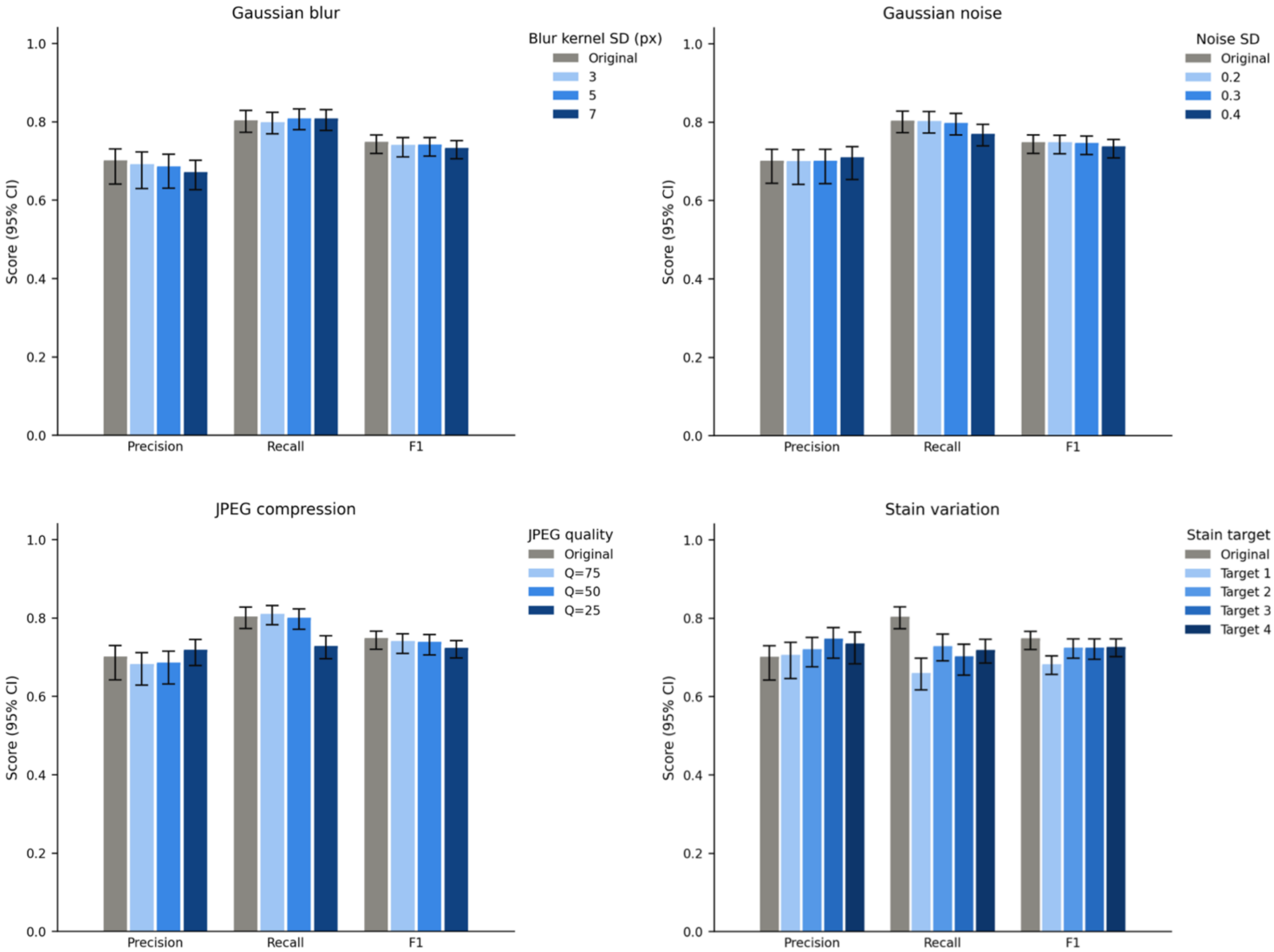

***Figure S22:*** *Detection performance of the combined validation dataset under simulated sources of technical variability. Panels show precision, recall and F1 score for Gaussian blur (top left; blur kernel SD in pixels), additive Gaussian noise (top right; noise SD), JPEG compression (bottom left; JPEG quality, lower values indicating stronger compression) and stain variation (bottom right; Macenko normalisation to four target images from independent centres), each compared with the unperturbed original. Error bars represent 95% confidence intervals.*